\documentclass[10pt,conference,compsoc]{IEEEtran}
\usepackage{diagbox}
\usepackage{comment}
\usepackage{siunitx}
\IEEEoverridecommandlockouts
\usepackage{textcomp}
\usepackage[table]{xcolor}
\usepackage{booktabs}
\usepackage[flushleft]{threeparttable}
\usepackage{multicol}
\usepackage{amssymb, amsthm}
\usepackage{multirow}
\usepackage{graphicx}
\usepackage[utf8]{inputenc}
\usepackage[english]{babel}
\usepackage{rotating}
\usepackage{array}
\usepackage{cite}
\usepackage{hyperref}
\usepackage{xcolor}

\newcolumntype{L}[1]{>{\raggedright\let\newline\\\arraybackslash\hspace{0pt}}m{#1}}
\newcolumntype{C}[1]{>{\centering\let\newline\\\arraybackslash\hspace{0pt}}m{#1}}
\newcolumntype{R}[1]{>{\raggedleft\let\newline\\\arraybackslash\hspace{0pt}}m{#1}}
\usepackage{makecell}

\usepackage{amsmath}

\usepackage{pgfplots}

\pgfplotsset{compat=1.9}

\usepgfplotslibrary{statistics}
\usetikzlibrary{pgfplots.groupplots}
\usetikzlibrary{patterns}

\usetikzlibrary{decorations}

\definecolor{ARC}{rgb}{0.4, 0.6, 0.8}
\definecolor{DBL}{rgb}{0.64, 0.76, 0.68}
\definecolor{\mbox{UWCNN-3}}{rgb}{0.01, 0.28, 1.0}
\definecolor{Fusion}{rgb}{0.0, 0.75, 1.0}
\definecolor{\mbox{UWCNN-I}}{rgb}{1.0, 0.11, 0.0}
\definecolor{WaterNET}{rgb}{0.5, 0.09, 0.09}
\definecolor{FUnIE}{rgb}{0.84, 0.04, 0.33}
\definecolor{UWHL}{rgb}{0.0, 0.42, 0.24}
\definecolor{DEH_blue_pantone}{RGB}{109,196,223}

\definecolor{UCIQE}{rgb}{1.0, 0.44, 0.37}
\definecolor{UIQM}{rgb}{0.56, 0.74, 0.56}
\definecolor{CCF}{rgb}{0.54, 0.81, 0.94}
\definecolor{CCF_over}{RGB}{110,169,242}

\definecolor{typeI}{rgb}{0.740,0.942,0.981}
\definecolor{typeIA}{rgb}{0.735,0.936,0.974}
\definecolor{typeIB}{rgb}{0.730,0.930,0.967}
\definecolor{typeII}{rgb}{0.710,0.905,0.940}
\definecolor{typeIII}{rgb}{0.680,0.865,0.885}
\definecolor{type1}{rgb}{0.680,0.865,0.840}
\definecolor{type3}{rgb}{0.650,0.810,0.750}
\definecolor{type5}{rgb}{0.620,0.720,0.600}
\definecolor{type7}{rgb}{0.580,0.630,0.420}
\definecolor{type9}{rgb}{0.520,0.530,0.210}

\definecolor{3mtypeI}{rgb}{0.222,0.742,0.909}
\definecolor{3mtypeIA}{rgb}{0.215,0.718,0.877}
\definecolor{3mtypeIB}{rgb}{0.207,0.696,0.846}
\definecolor{3mtypeII}{rgb}{0.180,0.607,0.734}
\definecolor{3mtypeIII}{rgb}{0.145,0.484,0.543}
\definecolor{3mtype1}{rgb}{0.145,0.484,0.418}
\definecolor{3mtype3}{rgb}{0.116,0.349,0.237}
\definecolor{3mtype5}{rgb}{0.092,0.193,0.078}
\definecolor{3mtype7}{rgb}{0.066,0.099,0.013}
\definecolor{3mtype9}{rgb}{0.038,0.042,0.000}

\ifCLASSINFOpdf
\else
\fi
\usepackage{graphicx}

\usepackage[normalem]{ulem}
\usepackage{etoolbox}

\newcounter{rowcntr}[table]
\renewcommand{\therowcntr}{\arabic{rowcntr}}
\newcolumntype{N}{>{\refstepcounter{rowcntr}\therowcntr}c}

\newcounter{rowcntrN}[table]
\renewcommand{\therowcntrN}{\arabic{rowcntrN}}
\newcolumntype{Q}{>{\refstepcounter{rowcntrN}\therowcntrN}c}
\AtBeginEnvironment{tabular}{\setcounter{rowcntr}{0}}
\AtBeginEnvironment{tabular}{\setcounter{rowcntrN}{0}}

\usetikzlibrary{decorations.pathmorphing}
\tikzset{discont/.style={decoration={zigzag,segment length=12pt, amplitude=4pt},decorate}}
\def\discontarrow(#1)(#2)(#3)(#4);{
  \draw[discont] (#2) -- (#3);
  \draw[->] (#1) -- (#2) (#3) -- (#4);
}

\begin{document}
\title{Underwater image filtering: methods, datasets and evaluation}



\author{Chau Yi Li, Riccardo Mazzon, Andrea Cavallaro}

\author{
\IEEEauthorblockN{Chau Yi Li}
\IEEEauthorblockA{chauyi.li@qmul.ac.uk}
\and
\IEEEauthorblockN{Riccardo Mazzon}
\IEEEauthorblockA{r.mazzon@qmul.ac.uk}
\and
\IEEEauthorblockN{Andrea Cavallaro}
\IEEEauthorblockA{a.cavallaro@qmul.ac.uk}
\thanks{All authors are at Centre for Intelligent Sensing, Queen Mary University of London.}
}

\maketitle

\begin{abstract}
Underwater images are degraded by the selective attenuation of light that distorts colours and reduces contrast. The degradation extent depends on the water type, the distance between an object and the camera, and the depth  under the water surface  the object is at. Underwater image filtering aims to restore or to enhance the appearance of objects captured in an underwater image. Restoration methods compensate for the actual degradation, whereas enhancement methods improve either the perceived image quality or the performance of computer vision algorithms. The growing interest in underwater image filtering methods--including learning-based approaches used for both restoration and enhancement--and the associated challenges call for a comprehensive review of the state of the art. In this paper, we review the design principles of filtering methods and revisit the oceanology background that is fundamental to identify the degradation causes. We discuss image formation models and the results of restoration methods in various water types. Furthermore, we present task-dependent enhancement methods and categorise datasets for training neural networks and for method evaluation.  
Finally, we discuss evaluation strategies, including subjective tests and quality assessment measures. We complement this survey with a platform (\url{https://puiqe.eecs.qmul.ac.uk/}), which hosts  state-of-the-art underwater filtering methods and  facilitates comparisons.
\end{abstract}

\section{Introduction}

Images taken in water are degraded by the selective attenuation of light, which causes colour casts and blurring~\cite{LightandWater, LightAbsorption, Jerlov}. Light attenuation reduces the illuminant on objects and the intensity of the reflected light reaching the camera~\cite{SPIE-uwmodel-McGlamery1980, Jaffe_model_1990}. The extent of the degradation  depends on the depth of an object under the water surface, its distance from the camera and the water  composition~\cite{Duntley:63, doi:10.1021/ed039p333, LightandWater}. While  artificial lighting can be used to compensate for the attenuated illuminant, this further degrades the image with a spherical overexposed area and an increased backscattering from suspended particles.

Underwater image filtering methods can be used to restore or enhance the content of an image. 
{\em Restoration} methods use a physics-based model~\cite{DBLP:conf/cvpr/SchechnerK04, RevisedModel_CVPR2018} to compensate for the effects of degradation in underwater  images~\cite{Schechner2005_polarised,Treibitz2009_descattering,Drews2013UDCP,Emberton2015,CB2010InitialResult,ARC,Blurriness_and_Light,BglightEstimationICIP2018_Submitted,MIL,Hou2020_DCPguided_TV,seathru2019CVPR,WCID,Emberton2018,BMVC_HL_2017, pmlr-v95-hu18a_restoration_CNN, Shin2016CNN,ICIP2017UWCNN}. Some methods need special devices such as polarisation filters~\cite{Schechner2005_polarised, Treibitz2009_descattering} or require the knowledge of the 
distance between the objects and the camera~\cite{seathru2019CVPR}. Furthermose, only a few methods compensate for the degradation through the depth~\cite{seathru2019CVPR, BMVC_HL_2017, WCID}. 
{\em Enhancement} methods modify colours and contrast either to improve the perceive quality of an image~\cite{Fu_retinex-based_ICIP2014,Tang_retinex_video_SOVP2019,Gao2019_Adaptive_Retinex,Sanchez-Ferreira_2019_BioInspired_ABC,Zhang2017_multisclae_retinex,Protasiuk2019_localcolourmapping,Ancuti2012,Ancuti2018,Iqbal2010UCC,JTF,Prabhakar2010_wavelet_threshold,Dai2019_uwlowlight,Khan2016_wavelet,VASAMSETTI2017_wavelet_variational} or to improve the performance of computer vision algorithms~\cite{Lakshmi2020_fishDetCountClass, mehrnejad2014towards, Rizzini2015,Foresti2000, Kallasi15, lee2012vision, Lopez2020_animalClassification_Cabled,skarlatos2012open,3DFromRobots_JohnsonRobertson_2010}.
In recent years, {\em learning-based} methods that use {neural networks} have gained increasing popularity~\cite{Li2018WaterGANUG,UIEDAL_Uplavikar_2019_CVPRW,Yu2019_UnderwaterGAN_PRIF,Hashisho_UNet_ISISPA_2019,Islam2019FastUI4ImprovedPerception,Li_2020_Cast-GAN,P2P_Sun_IET_2019,UGAN_Fabbri_2018,Anwar2018DeepUI_UWCNN,DenseGAN_Guo_2020,URCNN_ICIP2018_Hou, UWGAN_emerging_SPL_2018, MCycleGAN_Lu_Laser2019, li2019fusion}. The results of these methods heavily depend on the images used in training. However, images appropriate for the filtering purpose are difficult to generate, select or synthesise.

In this paper, we categorise, analyse and compare underwater image filtering methods for restoration and enhancement. We discuss the merits and limitations of these methods and of the evaluation measures used for their validation. We also revisit the fundamental oceanology knowledge to understand the causes of degradation in the water medium and present how degraded images can be modelled~(Sec.~\ref{sec:Preliminaries}). We discuss methods for restoration (Sec.~\ref{sec: methods_restoration}) and enhancement (Sec.~\ref{sec: methods_enhancement}), and then focus specifically on those based on neural networks (Sec.~\ref{sec: neural_network}).  
Moreover, we discuss commonly employed  datasets~(Sec.~\ref{sec: Datasets}) and evaluation approaches (Sec.~\ref{sec: evaluation}), and their limitations. Finally, we discuss open problems and future research directions (Sec.~\ref{sec: conclusion}). We complement our survey with an online Platform for Underwater Image Quality Evaluation (PUIQE)~\cite{PUIQE_website}, which enables testing the filtering methods and evaluation measures covered in this survey. The novelty of this survey is highlighted in Table~\ref{Tab: surveys}, which summarises and compares ours with previous surveys on this topic~\cite{Schettini2010,Lu2017,Wang2019,Yang2019,WaterNET_TIP2019, Raihan2019_review,Zhang2019,Anwar_Li_2020_Survey}. 


\section{Image formation in water}
\label{sec:Preliminaries}
In this section, we revisit the physics of degradations caused by the water medium and discuss underwater image formation models.

\subsection{What degrades underwater images?}

Light attenuation reduces by absorption the intensity of light travelling through a medium  and redirects by scattering its propagation~\cite{LightandWater}. This attenuation causes colour cast and blurring. 
Absorption reduces the intensity of ambient light that illuminates objects and affects the observed water colour that is related to the ambient light at depth~\cite{Jerlov}. The longer the distance the light travels, the more it is absorbed. Absorption also reduces the light reflected  by objects  along the object-camera distance~\cite{doi:10.1021/ed039p333}, which in this paper we refer to as the {\em range}.
Molecules and particles suspended in water change the direction of propagation of the light. This scattering increases the spatial width of the light beam  along the range, thus blurring  edges and details of the captured objects~\cite{Gould:99}. 

Water environments can be classified into  two main categories, namely  oceanic and coastal waters, according to the  Jerlov system~\cite{Jerlov}, which considers  light transmittance across one metre of water. In oceanic waters red is the most attenuated colour channel, whereas in coastal waters blue is the most attenuated colour channel. Oceanic and coastal waters are in turn divided into five subcategories, totalling 10 Jerlov water types (see Fig.~\ref{fig:: Jerlov_water_type}(a)).

\begin{table}[t!]
\caption{Surveys on underwater image filtering cover restoration (Restor.), enhancement (Enhancement), and learning-based (Learn.) methods. Enhancement methods aim to improve the perceived image quality or the task performance of computer vision algorithms (Task), which is first covered in our survey. While most surveys cover evaluation (Evaluation) in terms of the image quality assessment measures (IQA), subjective tests (Subj.) are first discussed in ours. 
}
\setlength\tabcolsep{1.8pt}
\centering 
\begin{tabular}{cc|cccc|c|ccL{1.1\columnwidth}ccccc}
\Xhline{3\arrayrulewidth}
\multicolumn{1}{c}{\textbf{Ref.}} &\textbf{Year}   &  
{\textbf{Restor.}} & \multicolumn{2}{c}{\textbf{Enhancement}} & \multicolumn{1}{c|}{{\textbf{Learn.}}} &  \textbf{Datasets}  &  \multicolumn{2}{c}{\textbf{Evaluation}} \\ 
 & &  & {Quality} & {Task} & & & {IQA}  & {Subj.} \\ 
\Xhline{3\arrayrulewidth}
\cite{Schettini2010}  &   2010   & \checkmark &  \checkmark &  &  &  &\checkmark & \\ 
\cite{Lu2017} & 2017& \checkmark & \checkmark  &  &  &  & \checkmark  & \\ 
\cite{Wang2019} & 2019& \checkmark &  \checkmark & &   &  & \checkmark &  \\
\cite{Yang2019} & 2019 & \checkmark & \checkmark &  &  & \checkmark & \checkmark &  \\ 
\cite{WaterNET_TIP2019} & 2019 &  \checkmark  & \checkmark &  & \checkmark & \checkmark & \checkmark & \\ 
\cite{Raihan2019_review} & 2019 & \checkmark &\checkmark  &  & \checkmark &  & &  \\ 
\cite{Zhang2019} & 2019  &  \checkmark & \checkmark &  & \checkmark &  & \checkmark  & \\ 
\cite{Anwar_Li_2020_Survey} & 2020 &  &  &  & \checkmark & \checkmark & \checkmark &  \\ \hline 
\multicolumn{2}{c|}{\textbf{Ours}}  &  \checkmark &   \checkmark & \checkmark & \checkmark & \checkmark & \checkmark  & \checkmark \\ 
\Xhline{3\arrayrulewidth}
\end{tabular}
\label{Tab: surveys}
\end{table}

The attenuation depends on the composition of water that may vary over time even at the same geo-location~\cite{Wernand2011_thesis}.  The optical properties can be divided into inherent and {apparent}.  
The {\em inherent} optical properties of water depend only on the water compositions~\cite{Preisendorfer_RTT_LightMeasurementInSea}. Examples of inherent properties include the beam absorption coefficient,~$a$, and the beam scattering coefficient,~$b$, which quantify the absorption and scattering of a collimated light beam, respectively. The beam attenuation coefficient,~$c$, is the sum of the beam absorption and scattering coefficients~\cite{Duntley:63}. The {\em apparent} optical properties of water depend instead on both the water composition and the ambient light. An example of apparent property is the diffuse attenuation coefficient of downwelling irradiance,~\mbox{$K_{d}$}, that quantifies the attenuation of ambient light~\cite{Jerlov}.  The intensity of ambient light illuminating an object at the vertical depth~$D$ can then be described as~\cite{Jerlov, Gordon_lambertLaw_valid}:
\begin{equation}\label{eq:ambLight}
E_D = E_0e^{-K_{\!d}D},
\end{equation}
where~$E_0$ is the ambient light intensity at the water surface. The deeper the object is under the water surface, the more attenuated the illuminant and hence the stronger the colour cast on that object. Furthermore, due to the exponential decrease in ambient light intensity (Eq.~\ref{eq:ambLight}), water types with large diffuse attenuation coefficients or scenes with large changes in the depth will have a non-uniform water colour. In addition, a portion of the ambient light is {\em backscattered} towards the camera and veils the object with the water colour,~$\frac{a}{c}E_D$~\cite{DBLP:conf/cvpr/SchechnerK04}. The farther from the camera an object, the stronger the veiling effect on its appearance. 

\begin{figure}[t!]
\setlength\tabcolsep{0.8pt}
\centering
\begin{tabular}{lr}

\begin{tikzpicture}
\node [anchor=east] (start) at (-0.08,3.9) {\footnotesize 0};
\node [anchor=east] (depth) at (-0.08,2) {\rotatebox{90}{\footnotesize depth (m)}};
\node [anchor=east] (end) at (-0.08,0.12) {\footnotesize 20};
 \draw [-stealth, line width=1pt, black] (-0.1,3.95) -- (-0.1,0.);
  \draw [decorate,decoration={brace,amplitude=4pt, mirror},xshift=0pt,yshift=2pt]
(0.05,-0.12) -- (1.86,-0.12) node [yshift=-8pt, black,midway] 
{\footnotesize Oceanic};
  \draw [decorate,decoration={brace,amplitude=4pt, mirror},xshift=0pt,yshift=2pt]
(1.9,-0.12) -- (3.71,-0.12) node [yshift=-8pt, black,midway] 
{\footnotesize Coastal};
\begin{scope}
    \node[anchor=south west,inner sep=0] (image) at (0,0) {\includegraphics[width=0.21\textwidth]{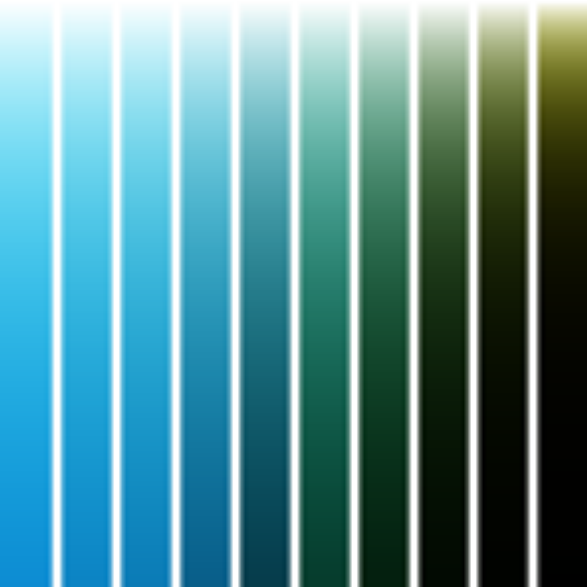}};
    \begin{scope}[x={(image.south east)},y={(image.north west)}]
    \end{scope}
\end{scope}
\end{tikzpicture}
& 
\begin{tikzpicture}
\node [anchor=west] (start) at (1.4,-0.32) {\footnotesize 0};
\node [anchor=west] (range) at (2.8,-0.32) {\footnotesize range (m)};
\node [anchor=west] (end) at (4.9,-0.32) {\footnotesize 20};
\draw [-stealth, line width=1pt, black] (1.52,-0.12) -- (5.33,-0.12);
\begin{scope}[xshift=1.5cm]
    \node[anchor=south west,inner sep=0] (image) at (0,0) {\includegraphics[width=0.212\textwidth]{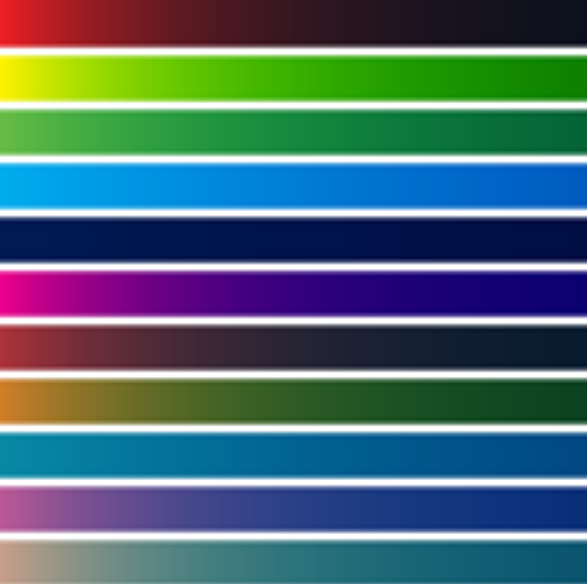}};
    \begin{scope}[x={(image.south east)},y={(image.north west)}]
    \end{scope}
\end{scope}
\end{tikzpicture}\\
\multicolumn{1}{c}{(a)} & \multicolumn{1}{c}{(b)} \\
\end{tabular}
\caption{Light attenuation in water. (a) Ambient light attenuated in the 10 Jerlov water types changes the water colour through the depth. Oceanic waters are of type I, IA, IB, II or III (left to right); coastal waters are of type 1, 3, 5, 7, or 9 (left to right)~\cite{Jerlov}: the larger the index, the more turbid the water. (b) Degradation along the range in type I water of 12 colours (leftmost of the rows at 0m). Note that the red colour appears black after 10m, and that yellow and green colours are indistinguishable after only a few metres. }
\label{fig:: Jerlov_water_type}
\end{figure}
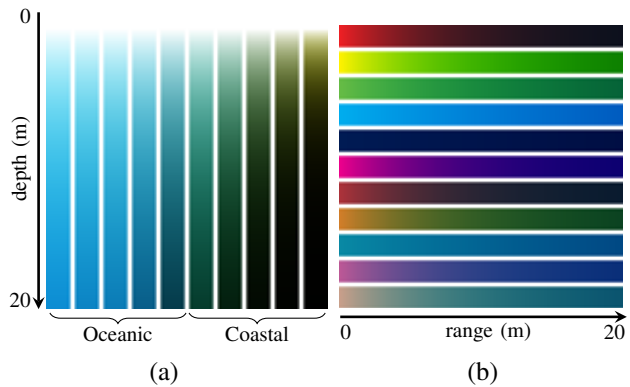

The intensity of the reflected light diminishes exponentially along its propagation in water~\cite{LightAbsorption}, modulated by the beam attenuation coefficient(s) and the distance travelled. The portion of the {transmitted} light intensity can be expressed as~\cite{doi:10.1021/ed039p333}:
\begin{equation}
\label{eq:: BeerLambert_Law} 
t =~\frac{\Phi_z}{\Phi_0} = e^{-cz},
\end{equation}
where~$\Phi_0$ is the initial light intensity and~$\Phi_z$ is the intensity after the light has travelled a distance of~$z$. As attenuation depends on the wavelength, each colour is degraded differently (see Fig.~\ref{fig:: Jerlov_water_type}(b)). 
Moreover, the transmission of lights of different wavelengths ($\lambda_1$ and~$\lambda_2$),  can be related as~\mbox{$t_{\lambda_2} = (t_{\lambda_1})^{{c_{\lambda_2}}/{c_{\lambda_1}}},$}
where~$c_{\lambda}$ is the attenuation coefficient of the subscripted wavelength~${\lambda}$. 

The spatial spread,~$w$, of the reflected light beam due to scattering along a range of~$z$ can be derived from the optical properties of the water and approximated as~\cite{scattering_Premoze04}:
\begin{equation}
w^2 \approx ~\frac{1}{2}\left(~\frac{2a}{3z} +~\frac{8}{b\langle\theta^2\rangle z^3}\right)^{-1},
\label{eq:: width_sigma}
\end{equation}
where~$\langle\theta^2\rangle$ is the average cosine angle that indicates the average scattering direction deviating from the original propagation. Typical oceanic waters have~$\langle\theta^2\rangle$ values between 0.80 and 0.95 that correspond to average angles between 
17.8\textdegree    and 36.6\textdegree. A larger scattering coefficient or a longer range travelled by the light results in a larger spread of  the beam. Therefore, at different ranges the degradation of the appearance of objects varies in their degree of colour cast and blurring.

As colour cast and blurring are caused by light attenuation that varies with the water type, depth and range, methods designed for a particular water environment or range may fail in other environments. For example, methods designed for shallow water or for scenes with objects at a short range usually under-perform in deeper waters and for objects farther from the camera, respectively.


\subsection{Image formation models}

This section reviews underwater image formation models that describe the degradation along the {\em range}, where the objects are under the illuminant attenuated along the {\em depth}. We discuss the earliest McGlamery-Jaffe model~\cite{SPIE-uwmodel-McGlamery1980, Jaffe_model_1990}, the Schechner-Karpel model~\cite{DBLP:conf/cvpr/SchechnerK04} -- which is the most used model in underwater image filtering -- and the Akkaynak-Treibitz model that revises the Schechner-Karpel model~\cite{RevisedModel_CVPR2018}.  The degraded image~$\textbf{I}$ comprises pixels that represent the objects in the scene or pure water mass. In the following, unless otherwise stated, all the arithmetic operations on images, including multiplications $\cdot$ and divisions  $\frac{()}{()}$, are element-wise. 

The McGlamery-Jaffe model~\cite{SPIE-uwmodel-McGlamery1980, Jaffe_model_1990} describes~$\textbf{I}$ as a linear combination of the \textit{attenuated} lights that are directly reflected from objects,~$\textbf{D}$, and the attenuated scattered reflected light from objects,~$\textbf{F}$, and transmitted backscattered light,~$\textbf{B}$, as 
\begin{equation}\label{eq:model}
\textbf{I} = \textbf{D} + \textbf{F} + \textbf{B}.
\end{equation}
The complete McGlamery-Jaffe model involves camera parameters and water properties, such as the volume scattering functions and the point spread function that governs scattering in~$\textbf{F}$. The water property information is particularly difficult to obtain as they are constantly changing even at the same geo-location~\cite{Wernand2011_thesis} and can only be measured with specialised devices~\cite{Jaffe_model_1990}, hence this model is mainly used in underwater navigation or monitoring where the devices are readily available~\cite{Jaffe_model_1990}. 

The Schechner-Karpel model considers how the \mbox{\textit{un-attenuated}} reflected light (effective scene radiance,~$\textbf{J}$) is degraded along the objects' distance from the camera, encoded in the range map~$\textbf{z}$~\cite{DBLP:conf/cvpr/SchechnerK04}. The model describes~$\textbf{I}$ as a combination of~$\textbf{J}$ and the backscattered ambient light (background light,~$\textbf{A}$) that represents the water colour, as
\begin{equation}\label{eq:model_X}
\textbf{I} =\textbf{T}\cdot~\textbf{J} + (\bf{1}-\textbf{T})~\cdot\textbf{A},
\end{equation}
where~\mbox{$\textbf{T}=e^{-c\textbf{z}}$} is the transmission map that describes the degradation extent of~$\textbf{J}$ with  value of each entry between 0 and 1 (Eq.~\ref{eq:: BeerLambert_Law}). The farther an object from the camera is, the smaller the value of~$\textbf{T}$ is. The term~\mbox{$\bf{1}-\textbf{T}$} describes the contribution of~$\textbf{A}$ in the degraded image. All entries in~$\bf{1}$, which has the same dimensions as~$\textbf{T}$, have values of 1. Moreover, the relationship between transmission maps of different colour channels can be obtained from~Eq.~\ref{eq:: BeerLambert_Law}, for example as
\begin{equation}
\label{eq:: transmission_relationship_wavelength}
 \textbf{T}_{r} = (\textbf{T}_{g})^{\frac{c_r}{c_g}},
\end{equation}
where~$c_r$ and~$c_g$ are the attenuation coefficients of red and green, respectively.

Akkaynak and Treibitz~\cite{RevisedModel_CVPR2018} noted {that} the transmissions for \textbf{J} and \textbf{A} are in fact different. In particular, the attenuation coefficients, $c^\textbf{J}$ and~$c^\textbf{A}$, which quantify the two transmissions, depend on different variables. While both coefficients vary with~$\textbf{E}$,~$c$, and the camera spectral function,~$c^\textbf{J}$ also varies with~$\textbf{z}$, whereas~$c^\textbf{A}$ also varies with the beam scattering coefficient~$b$. Thus the revised model in Eq.~\ref{eq:model} takes the following form:
\begin{equation}
\begin{alignedat}{4}
\label{eq:model_revised}
    \textbf{I} & = \textbf{T}^{\textbf{J}} \cdot \textbf{J} + (\textbf{1}-\textbf{T}^{\textbf{A}})\cdot\textbf{A},\\
    \text{where} \quad \textbf{T}^{\textbf{J}}&=e^{-c^\textbf{J}\textbf{z}}
     \quad  \text{and} \quad
    \textbf{T}^{\textbf{A}}=e^{-c^\textbf{A}\textbf{z}}.
    \end{alignedat}
\end{equation}

To compensate for the degradation along the range with Akkaynak-Treibitz model,~\textbf{J} is obtained as
\begin{equation}
\label{eq:modelJ}
\textbf{J} = 
\frac{\textbf{I} - ({\bf{1}} -\textbf{T}^{\textbf{A}})~\cdot\textbf{A}}{\textbf{T}^\textbf{J}}, 
\end{equation}
%
Similarly, for the Schechner-Karpel model (Eq.~\ref{eq:model}),~\textbf{J} is obtained with the above equation where \mbox{$\textbf{T}^\textbf{J}=\textbf{T}^\textbf{A}=\textbf{T}$}.
Note that underestimating the transmission overcompensates colours, as the smaller the values of the estimated~$\textbf{T}$ ($\textbf{T}^\textbf{J}$), the stronger the compensation is.  Furthermore, an inaccurately estimated~$\textbf{A}$ can distort the water colour. In particular,  challenging scenes have a non-uniform~$\textbf{A}$,  which is caused by the exponential decrease in ambient light intensity. 
%
%

Eq.~\ref{eq:modelJ}, which only compensates for the degradation along the range, can be extended to compensate for the attenuation through the depth if the illuminant~$\textbf{E}$ is known. 
{The scene radiance} under a white (un-attenuated) illuminant,~$\textbf{J}_\textbf{0}$,  can be obtained as~\cite{BUCHSBAUM1980_grayworld}:
\begin{equation}
\label{eq:modelJ0}
   \textbf{J}_\textbf{0} =  
   \frac{\textbf{J}}{\textbf{E}}.
\end{equation}

In the next section we discuss {the} restoration methods that use physics-based models to obtain~$\textbf{J}$ or~$\textbf{J}_\textbf{0}$. 

\section{Restoration}
\label{sec: methods_restoration}

In this section, we present restoration methods that use the physics-based models. These methods estimate~the background light and~transmission map(s) in Eq.~\ref{eq:modelJ} using extra information of the scene~\cite{Schechner2005_polarised, Treibitz2009_descattering,seathru2019CVPR} and {\em priors} that are learned using neural networks~\cite{Shin2016CNN, ICIP2017UWCNN, pmlr-v95-hu18a_restoration_CNN} or derived from  observations~\cite{Drews2013UDCP,Emberton2015,CB2010InitialResult,ARC,Blurriness_and_Light,BglightEstimationICIP2018_Submitted,MIL,Hou2020_DCPguided_TV,seathru2019CVPR,Emberton2018,WCID,BMVC_HL_2017,Shin2016CNN}. Some methods also estimate~the illuminant in Eq.~\ref{eq:modelJ0} using priors~\cite{BMVC_HL_2017, seathru2019CVPR} or from the estimated~background light~\cite{WCID}.

The transmission map, whose entries' values vary within the image, receives most effort from state-of-the-art among all the elements to be estimated. Other elements such as the background light are often assumed to be uniform within the image, although this is not universally true. Table~\ref{Table: restoration} summarises the restoration methods. 

\begin{table*}[t!]
\caption{Restoration methods that use a physics-based model to compensate for the colour degradation along the range (distance between an object and the camera) and the depth from the water surface. Additional considerations for the restoration are the non-uniform illuminant (Non-uni.),  which can be due to natural attenuation (Nat.) or artificial light (Art.), and a different transmission map for each colour channel (wavelength dependency, $\lambda$-dep.). The estimations can be simplified using extra scene information (Scene Info.). Priors can be derived from trained neural networks (NN), colour appearance in outdoor or underwater (Water) environments,  texture appearance, or colour statistics (Stat.). The underwater-specific image quality assessment measure (UW-IQA) used to validate the method is shown in the evaluation column, along with task evaluations, subjective tests and generic IQA measures. 
}
\scriptsize
\begin{minipage}[b]{1.0\linewidth}
\centering
\setlength\tabcolsep{0.7pt}
\begin{tabular}{ll|ccccccccccccc|cccc}
\Xhline{3\arrayrulewidth}
\multicolumn{2}{c|}{{\textbf{Reference}}} & \multicolumn{2}{c}{\textbf{Compensate}} &  \multicolumn{2}{c}{\textbf{ Non-uni.}} & \multicolumn{1}{c}{$\lambda$\textbf{-dep.}}  & \textbf{Scene Info.} & \multicolumn{5}{c}{\textbf{Prior}} & \multicolumn{2}{c|}{\textbf{Water type}} & \multicolumn{4}{c}{\textbf{Evaluation}} \\
\textbf{} & \textbf{} & Depth & Range &  Nat. & Art. & \textbf{} &  &  NN & {{Outdoor}} & {{Water}} & {{Texture}} & {{Stat.}} &{Oceanic} & {Coastal} & Task & Subjective & Generic & UW-IQA \\
\Xhline{3\arrayrulewidth}
\cite{Schechner2005_polarised} & Schechner and Karpel &  & \checkmark &  &  & \checkmark & \checkmark &  &  &  &  &  & \checkmark & \checkmark &  &  &  &  \\
\cite{Treibitz2009_descattering} & Treibitz and Schechner &  & \checkmark &  &  &  & \checkmark &  &  &  &  &  & \checkmark & \checkmark &  &  &  &  \\
\cite{pmlr-v95-hu18a_restoration_CNN} & Hu {et al.} &  & \checkmark &  &  &  &  & \checkmark & \multicolumn{4}{c}{} & \checkmark &  & \checkmark &  & \checkmark &  \\
\cite{Shin2016CNN} & Shin {et al.} &  & \checkmark &  &  &  &  & \checkmark & \multicolumn{4}{c}{} & \checkmark &  &  &  &  &  \\
\cite{ICIP2017UWCNN} & Wang {et al.} &  & \checkmark &  &  &  &  & \checkmark & \multicolumn{4}{c}{} &  &  &  &  & \checkmark & \cite{PCQI_Wang2015}\\
\cite{Drews2013UDCP} & Drews {et al.} &  & \checkmark &  &  &  &  &  &  & \checkmark &  &  & \checkmark &  &  &  &  &  \\
\cite{Emberton2015} & Emberton {et al.} &  & \checkmark &  &  &  &  &  &  & \checkmark &  &  & \checkmark &  &  & \checkmark & \checkmark &  \\
\cite{CB2010InitialResult} & Carlevaris-Bianco {et al.} &  & \checkmark &  &  &  &  &  &  & \checkmark &  &  & \checkmark &  &  &  &  &  \\
\cite{ARC} & Galdran {et al.} &  & \checkmark &  & \checkmark & \checkmark &  &  &  & \checkmark &  &  & \checkmark &  &  &  & \checkmark &  \\
\cite{Blurriness_and_Light} & Peng and Cosman &  & \checkmark &  & \checkmark & \checkmark &  &  &  & \checkmark & \checkmark &  & \checkmark &  &  &  & \checkmark & \cite{UCIQE},\cite{UIQM_Panetta2016} \\
\cite{BglightEstimationICIP2018_Submitted} & Li and Cavallaro &  & \checkmark &  \checkmark &  & \checkmark &  &  &  & \checkmark & \checkmark &  & \checkmark &  & \checkmark & \checkmark &  &  \\
\cite{MIL} & Li {et al.} &  & \checkmark &  &  & \checkmark &  &  &  & \checkmark & \checkmark & \checkmark & \checkmark & \checkmark &  &  & \checkmark & \cite{UCIQE} \\
\cite{Hou2020_DCPguided_TV} & Hou {et al.} &  & \checkmark &  &  &  &  &  &  & \checkmark & \checkmark &  & \checkmark &  &  & \checkmark & \checkmark & \cite{UCIQE} \\
\cite{WCID} & Chiang and Chen & \checkmark & \checkmark &  \checkmark  & \checkmark & \checkmark &  &  & \checkmark & \checkmark &  &  &  & \checkmark & \checkmark & \checkmark &  &  \\
\cite{seathru2019CVPR} & Akkaynak and Treibitz & \checkmark & \checkmark &  &  & \checkmark & \checkmark &  &  &  &  & \checkmark & \checkmark & \checkmark & \checkmark &  &  &  \\
\cite{Emberton2018} & Emberton {et al.} & \checkmark & \checkmark &  &  &  &  &  &  & \checkmark &  & \checkmark & \checkmark & \checkmark &  & \checkmark &  & \cite{UCIQE,PKU} \\
\cite{BMVC_HL_2017} & Berman {et al.} & \checkmark & \checkmark &  &  & \checkmark &  &  & \checkmark &  &  & \checkmark & \checkmark & \checkmark &  &  &  &  \\
 \Xhline{3\arrayrulewidth}
\end{tabular}
\label{Table: restoration}
\end{minipage}
\end{table*}

%

\subsection{Using scene information}

Information about the scene is used to simplify the estimation. This information can be obtained with devices, such as polarised filters, in image acquisition~\cite{Schechner2005_polarised, Treibitz2009_descattering}, or by algorithms that require specific set-up in the scene~\cite{seathru2019CVPR}.

Polarised filters can be used to capture the transmitted background light,~$\textbf{B}=({\bf{1}}-\textbf{T})\cdot\textbf{A}$. Since light is partially polarised,~$\textbf{B}$ can be described as the superposition (addition) of one unpolarised and one polarised components. Positioning the polarised filters at two orthogonal orientations captures the two linearly polarised components (at maximum and minimum intensities) of~\mbox{$\textbf{B}$}, which can then be added up to obtain the values of $\textbf{B}$. The transmission
map $\textbf{T}$ can then be derived from $\textbf{B}$ with $\textbf{A}$, which can
be manually selected from the image~\cite{Schechner2005_polarised} or automatically
estimated using the degree of polarisation~\cite{Treibitz2009_descattering}.

Information about the objects' range,~$\textbf{z}$, can simplify the estimation of transmission maps to that of beam attenuation coefficient (Eq.~\ref{eq:: BeerLambert_Law}). The exact~$\textbf{z}$ can be obtained from Structure from Motion algorithm, that only estimates the range in relative scales, by placing an object of known size in the scene~\cite{seathru2019CVPR}. An optimisation algorithm then estimates the range-dependent attenuation coefficients (for Akkaynak-Treibitz model~\cite{RevisedModel_CVPR2018}) using, as inputs, the known~$\textbf{z}$ and a local illuminant (estimated by the local space average  colour~\cite{Depthmapcolorconstancy})~\cite{seathru2019CVPR}. 
Moreover, pixels representing either objects faraway from the camera (large $\textbf{z}$) or objects with low reflected light intensities (small $\textbf{J}$) have intensities similar to the background light,~i.e.~$\textbf{I}\approx\textbf{A}$  as~$\textbf{T}\cdot\textbf{J}\approx\textbf{0}$. By identifying such pixels in~$\textbf{I}$ with low RGB intensities from each of several equidistant $\textbf{z}$ value clusters,~$\textbf{A}$ is then estimated using non-linear regression~\cite{seathru2019CVPR}.


Using a device or set-up restricts the methods to images captured by the specific device, and therefore cannot be applied to already captured images.

%

\subsection{Using priors}
When only the degraded image~$\textbf{I}$ is available, estimating~$\textbf{A}$ and~$\textbf{T}$ requires priors that describe how the image was degraded.
%
%
These priors may be learnt using neural networks~\cite{Shin2016CNN, ICIP2017UWCNN, pmlr-v95-hu18a_restoration_CNN}, or  refer to the appearance of colour~\cite{BMVC_HL_2017, BglightEstimationICIP2018_Submitted, Drews2013UDCP, Emberton2015, CB2010InitialResult, ARC, Emberton2018,WCID} or textures~\cite{BglightEstimationICIP2018_Submitted, Blurriness_and_Light} in the degraded underwater image. 

Neural networks employed in the estimation were trained on pairs of degraded images~$\textbf{I}$ and the corresponding~$\textbf{A}$ and~$\textbf{T}$~\cite{Shin2016CNN, ICIP2017UWCNN, pmlr-v95-hu18a_restoration_CNN}. These methods synthesise~$\textbf{I}$ from degradation-free images using Eq.~\ref{eq:model}. The ranges of the scene in images can estimated using a range estimation algorithm~\cite{Liu_DepthEstimation_CVPR2015} for images~\cite{pmlr-v95-hu18a_restoration_CNN} or already provided by the dataset~(e.g.~the Middlebury Stereo dataset~\cite{Middlebury_stereo})~\cite{pmlr-v95-hu18a_restoration_CNN}. The transmission map~$\textbf{T}$ can then be synthesised with the obtained~$\textbf{z}$ (Eq.~\ref{eq:: BeerLambert_Law}). Some methods also randomly generate~$\textbf{T}$~\cite{Shin2016CNN, ICIP2017UWCNN} which does not reflect the dependency on range, 
or randomly generate~$\textbf{A}$ which does not respect the physics laws of attenuation~\cite{Shin2016CNN, ICIP2017UWCNN, pmlr-v95-hu18a_restoration_CNN}. These methods hence train on unrealistic synthesised images, which hinders the power of neural networks. 




Most methods estimate a uniform~$\textbf{A}$ from the pixels with the smallest values in~$\textbf{T}$, which represent the furthest points in the scene~\cite{Drews2013UDCP, Emberton2015, CB2010InitialResult, ARC,Blurriness_and_Light, Emberton2018,WCID}. Misestimating~$\textbf{T}$ would then lead to misestimating~$\textbf{A}$. An alternative is to select~$\textbf{A}$ directly from pixels representing the water mass in~$\textbf{I}$. These pixels can be  identified by low intensity variance in a local window~\cite{BMVC_HL_2017} or those following the oceanic water colour (derived from empirical  data~\cite{Solonenko:15:inherit})~\cite{BglightEstimationICIP2018_Submitted}. Moreover, a non-uniform~$\textbf{A}$ can be estimated when both the change in depth in the scene and the attenuation coefficients are known~\cite{WCID}. However, this information is generally unavailable. 
Another approach directly estimates the change in~$\textbf{A}$ from pixels representing water in~$\textbf{I}$, without extra information about the scene, using linear regression~(Eq.~\ref{eq:ambLight})~\cite{BglightEstimationICIP2018_Submitted}.

%
%

\begin{figure*}[t!]
\setlength\tabcolsep{1pt}
\renewcommand{\arraystretch}{0.55}
\centering
\begin{tabular}{ccccccccccc}
 \multirow{2}{*}{\textbf{Prior}}
 & \multicolumn{2}{c}{\textbf{Outdoor}}  & \multicolumn{3}{c}{\textbf{Water-specific}} & \multicolumn{1}{c}{\textbf{Texture}} \\
\cmidrule(r){2-3}\cmidrule(r){4-6}
 & DCP* & Haze-line* & UDCP* & ARC* & CB & Laplacian py.\\
\includegraphics[width=0.138\textwidth]{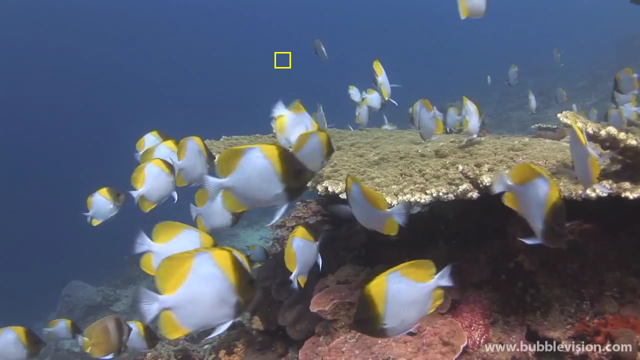}& \includegraphics[width=0.138\textwidth]{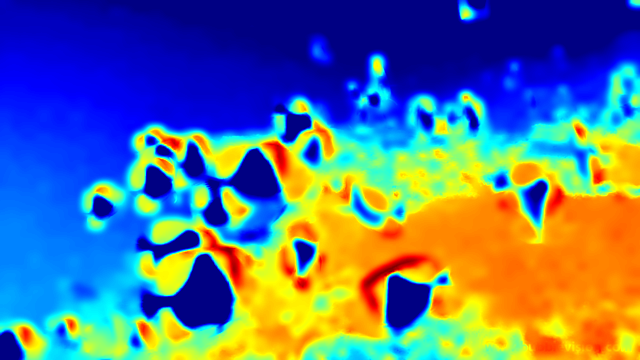} & \includegraphics[width=0.138\textwidth]{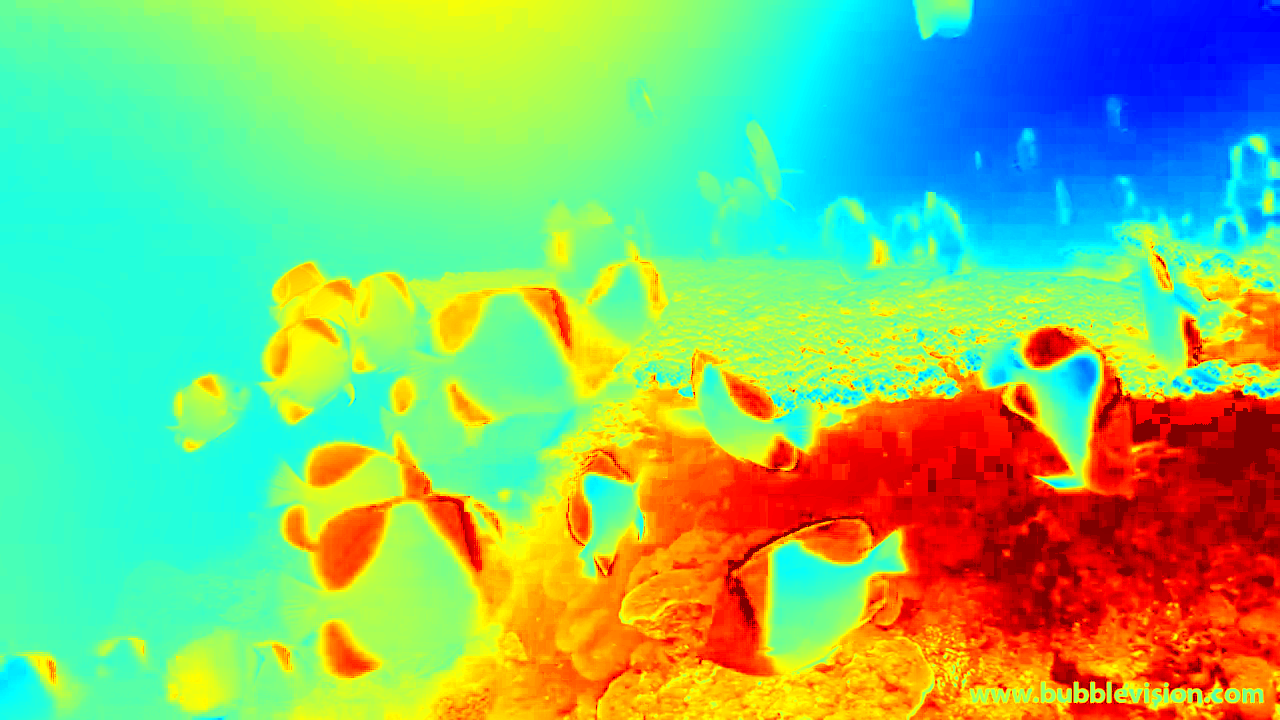}  & \includegraphics[width=0.138\textwidth]{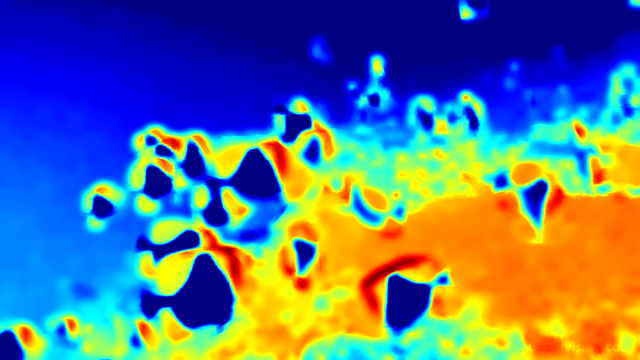} &  \includegraphics[width=0.138\textwidth]{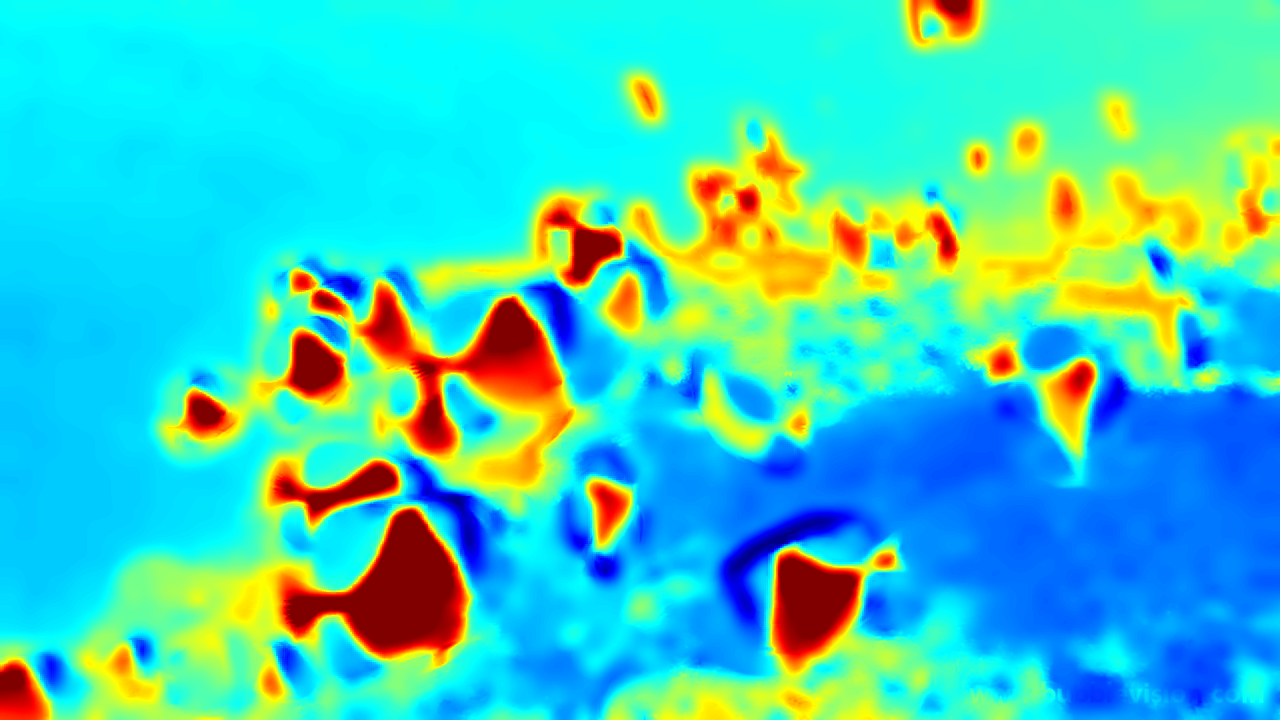} &
\includegraphics[width=0.138\textwidth]{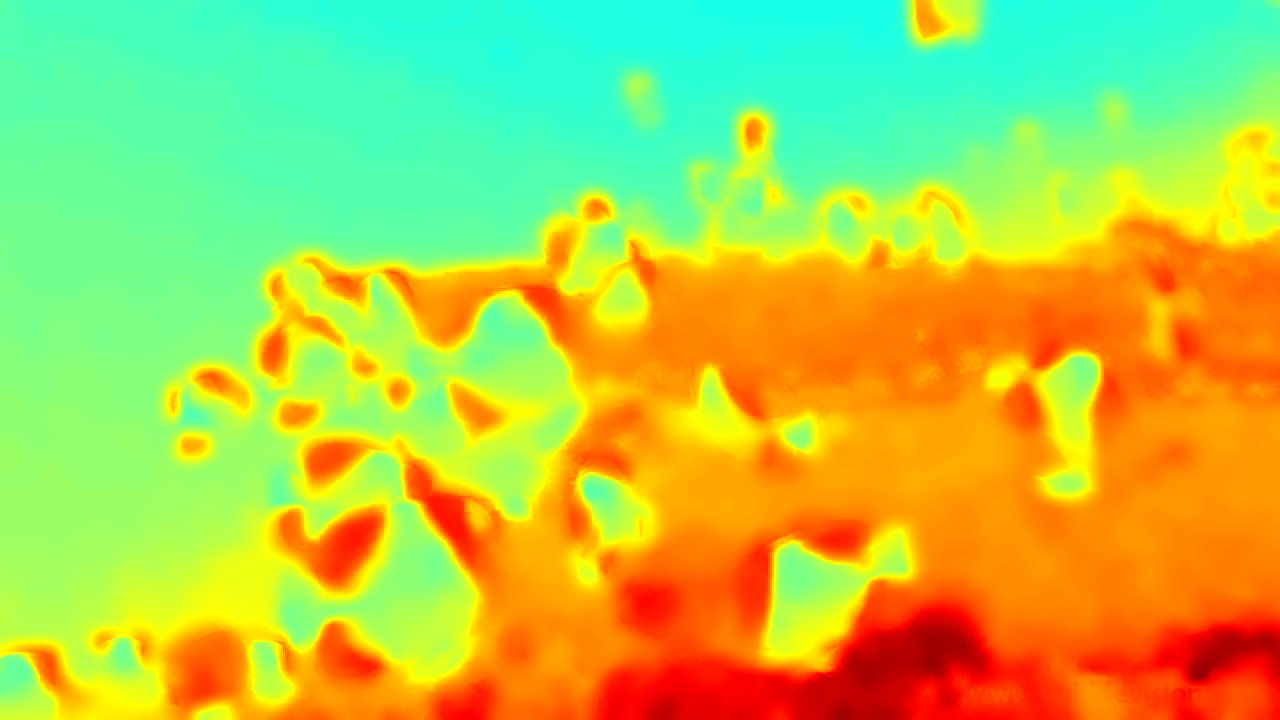} & \includegraphics[width=0.138\textwidth]{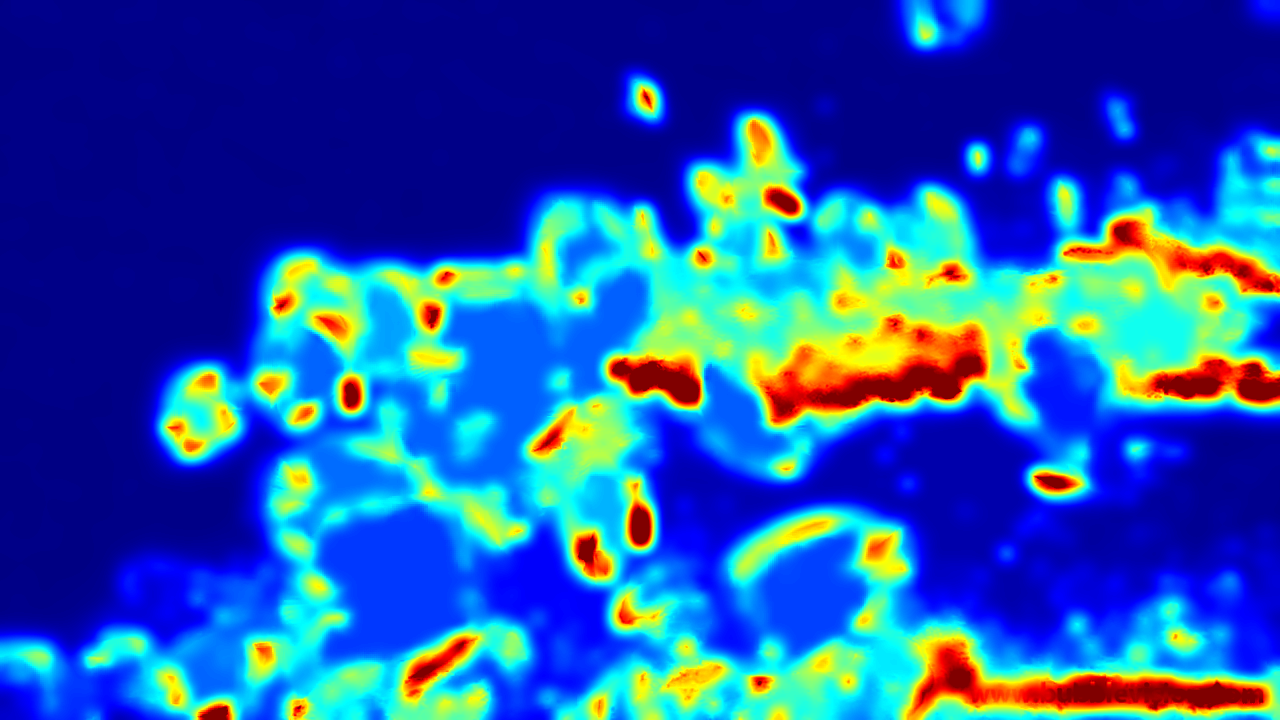} & 
\\
\includegraphics[width=0.138\textwidth]{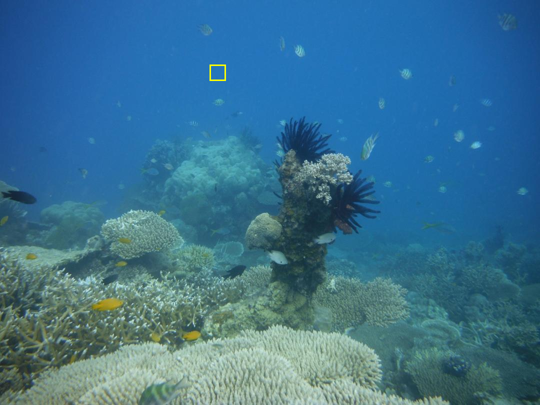}& \includegraphics[width=0.138\textwidth]{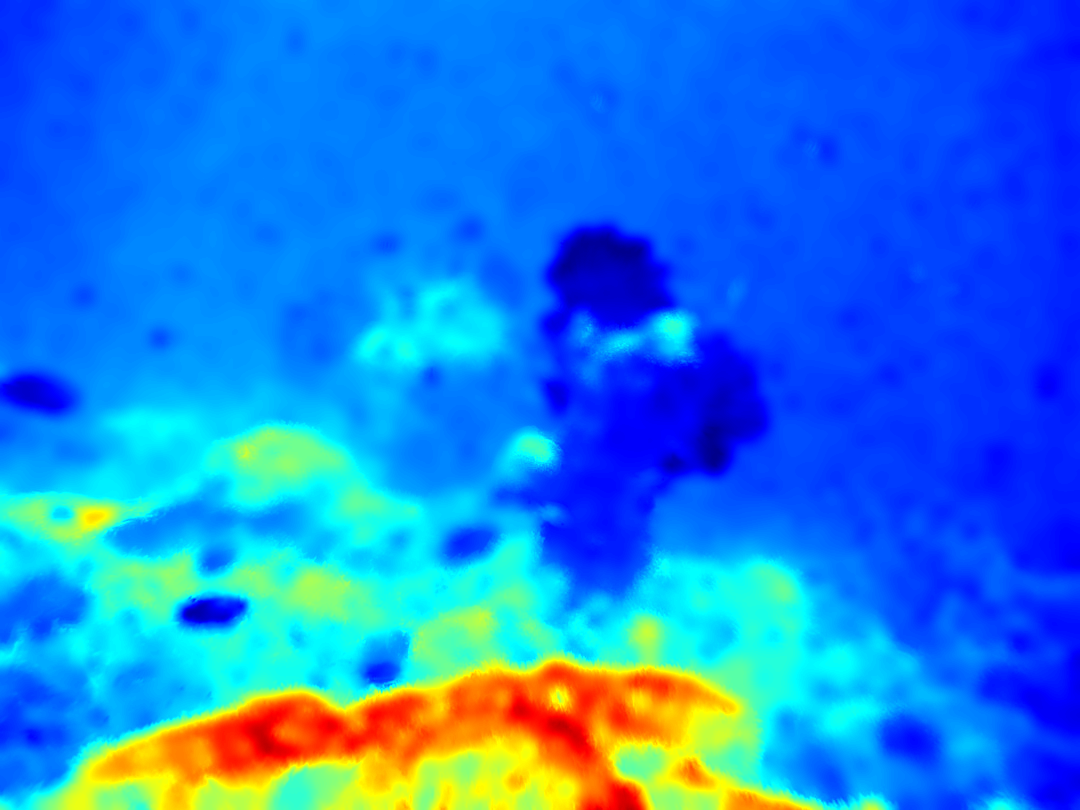} & \includegraphics[width=0.138\textwidth]{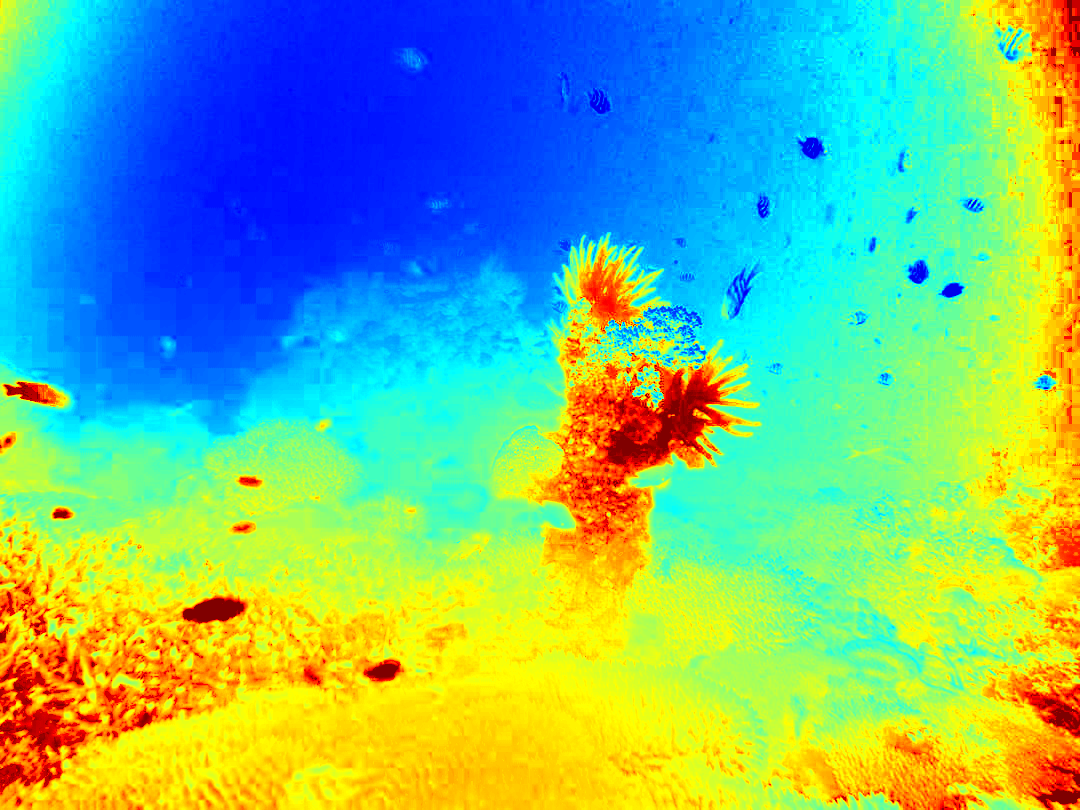} & \includegraphics[width=0.138\textwidth]{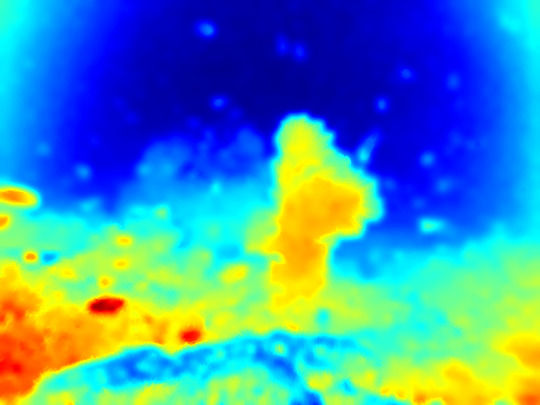} &  \includegraphics[width=0.138\textwidth]{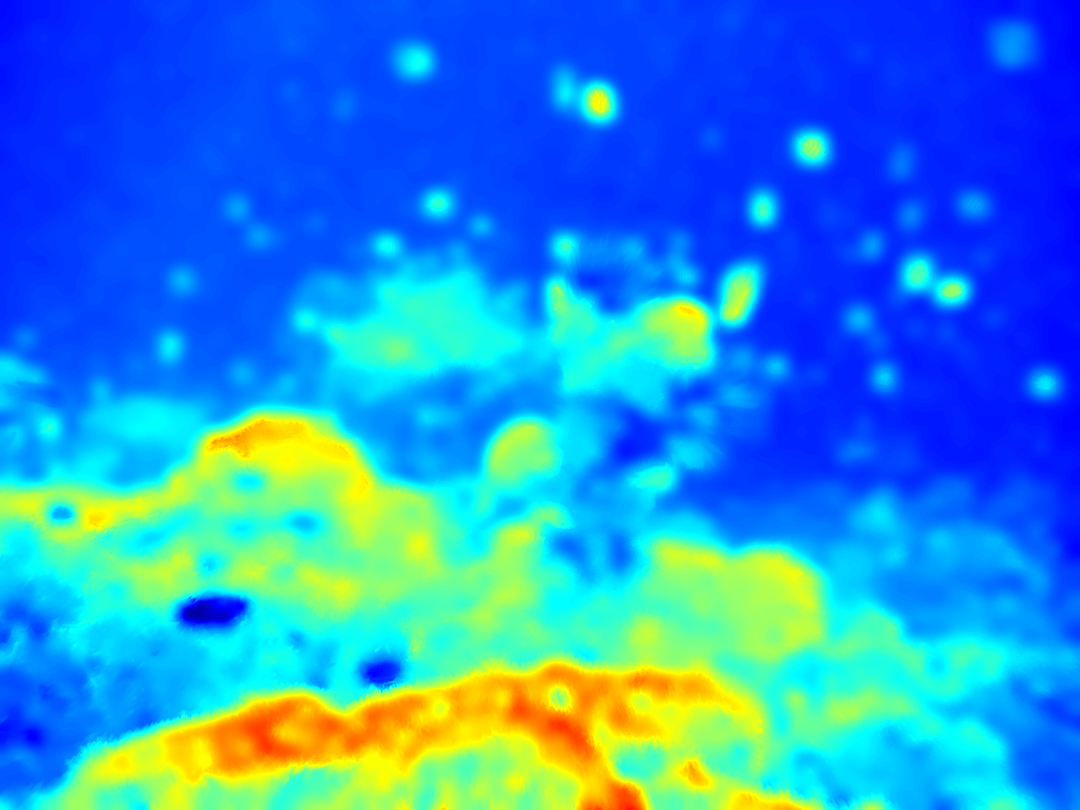} &
\includegraphics[width=0.138\textwidth]{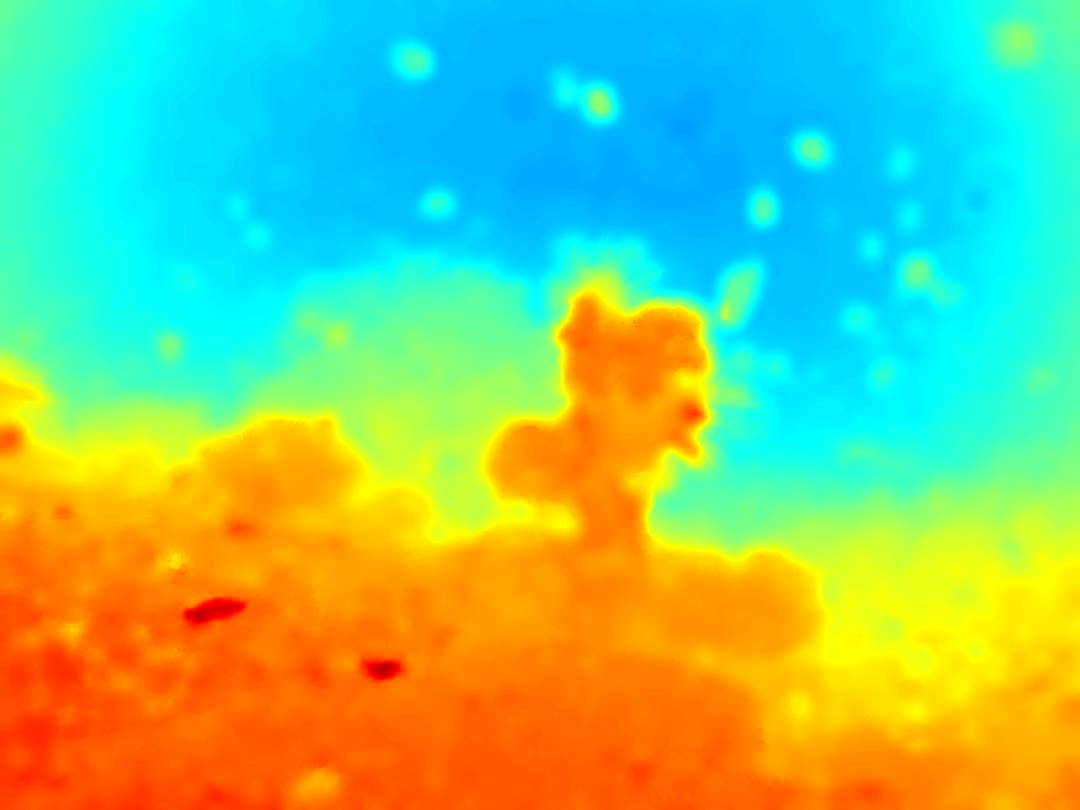} & \includegraphics[width=0.138\textwidth]{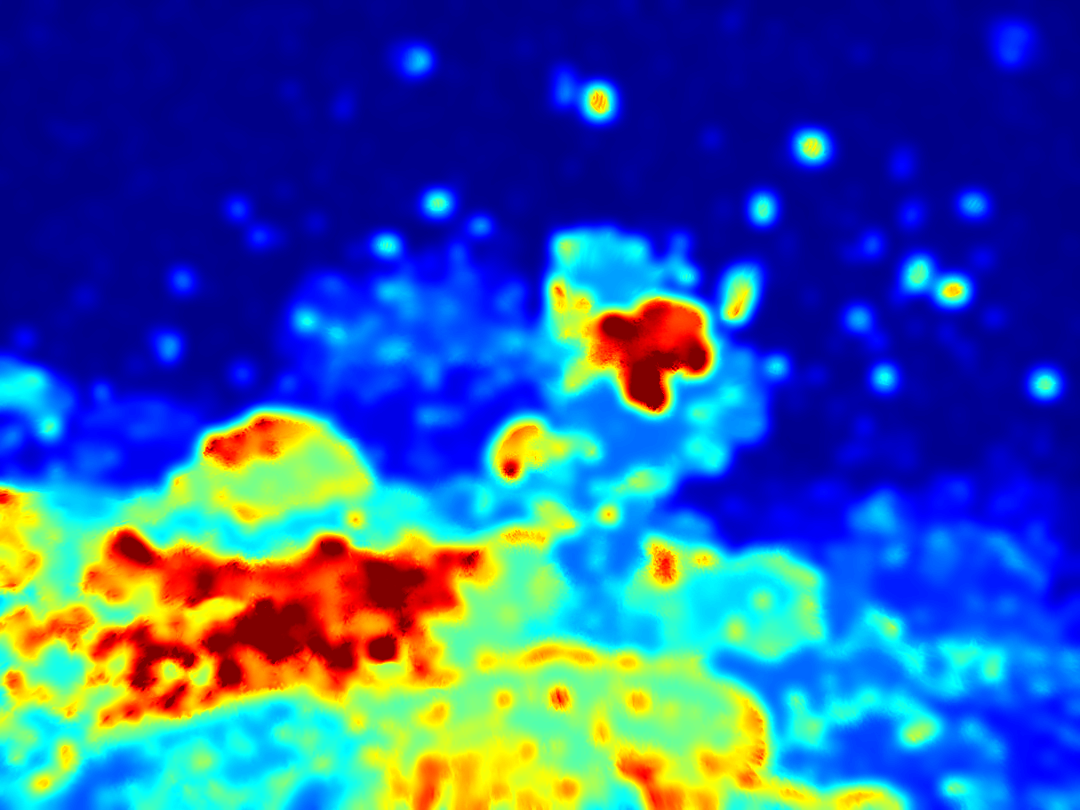} \\
\includegraphics[width=0.138\textwidth]{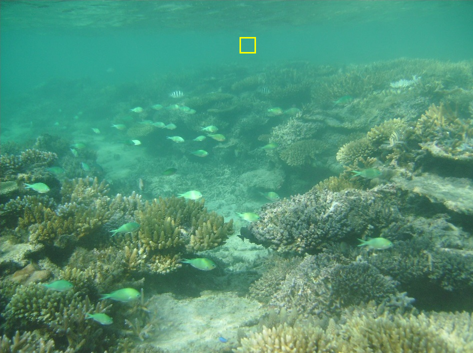}& \includegraphics[width=0.138\textwidth]{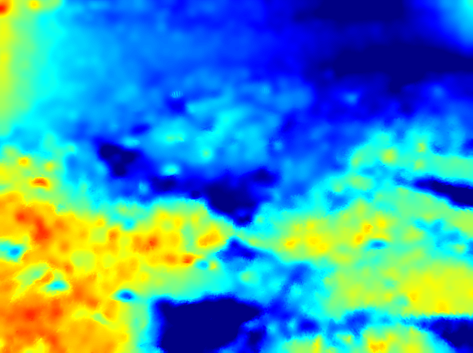} & \includegraphics[width=0.138\textwidth]{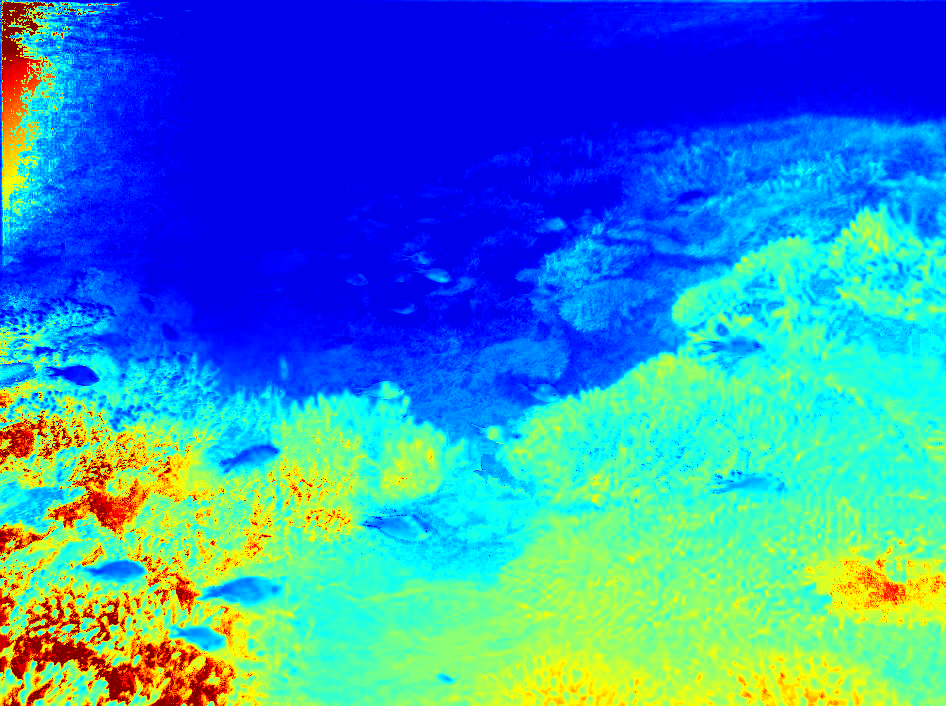} & \includegraphics[width=0.138\textwidth]{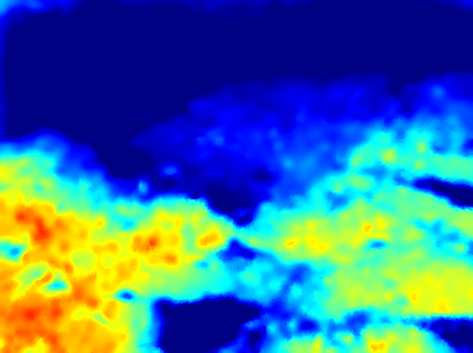} &  \includegraphics[width=0.138\textwidth]{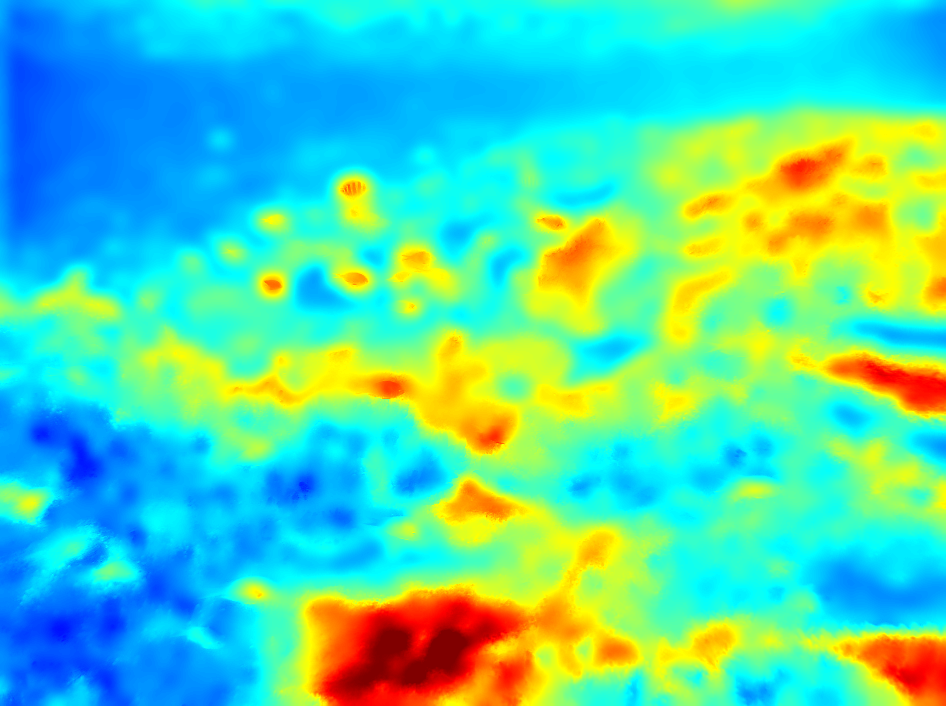} &
\includegraphics[width=0.138\textwidth]{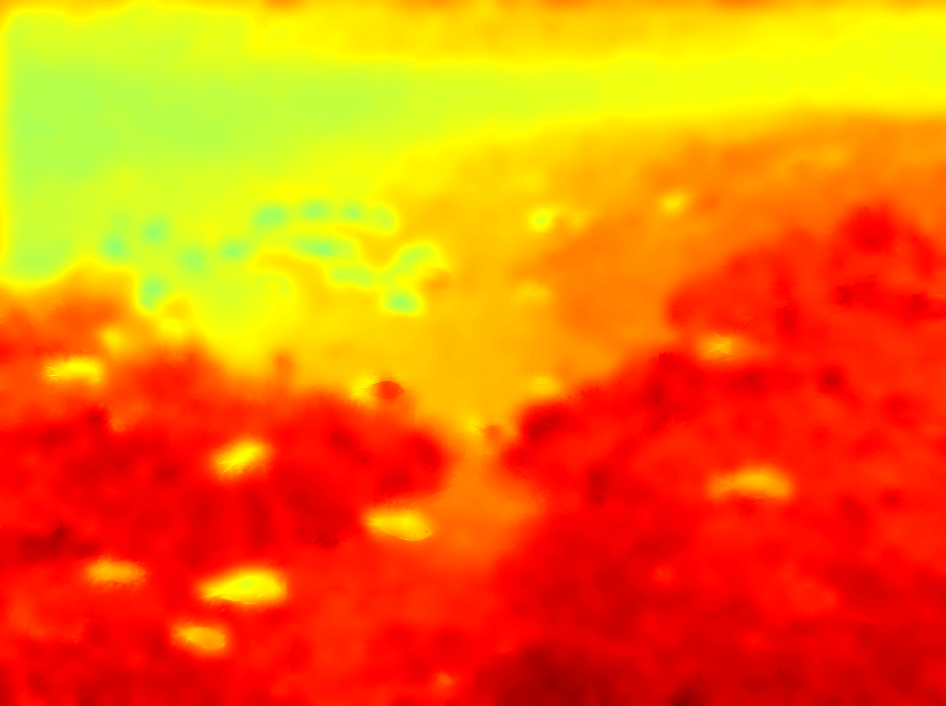} & \includegraphics[width=0.138\textwidth]{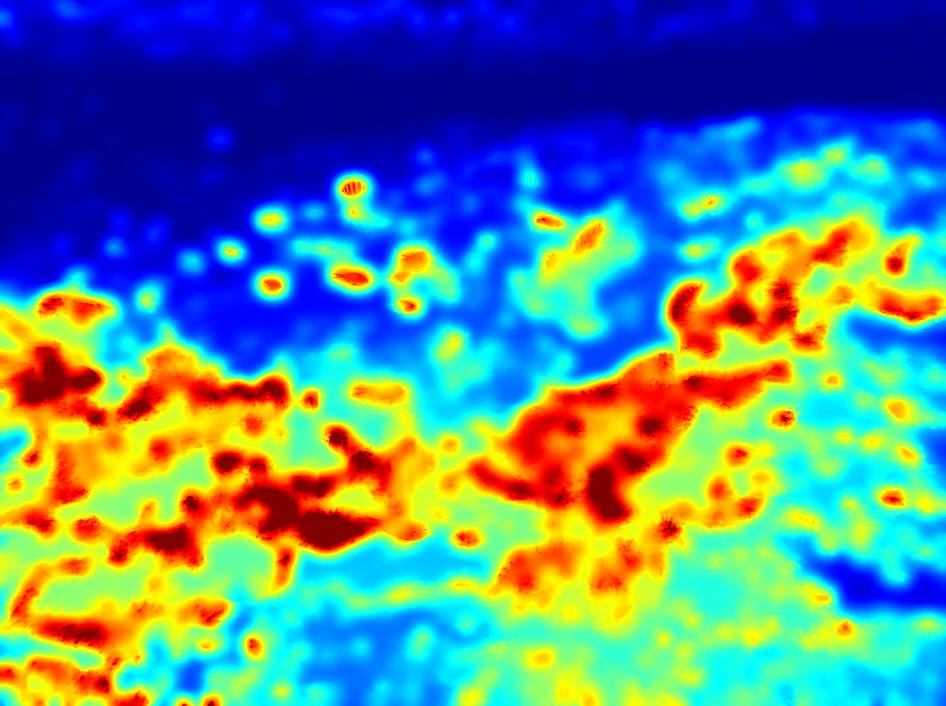}  \\
\includegraphics[width=0.138\textwidth]{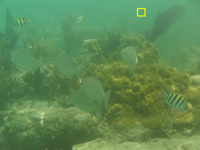}& \includegraphics[width=0.138\textwidth]{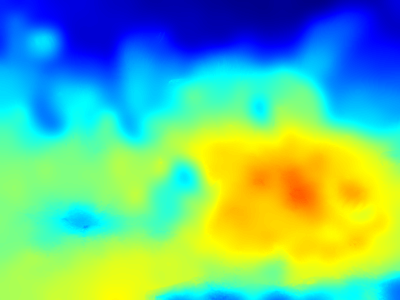} & \includegraphics[width=0.138\textwidth]{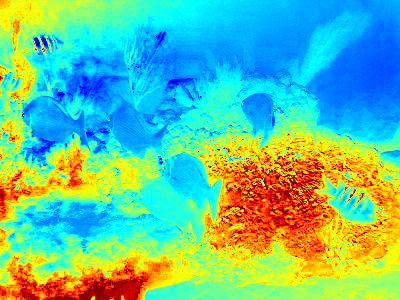} & \includegraphics[width=0.138\textwidth]{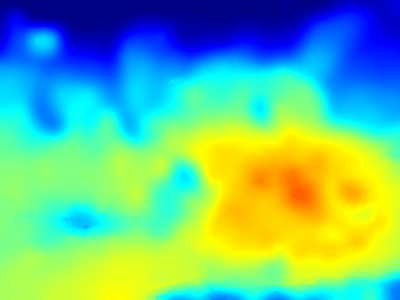}  &  \includegraphics[width=0.138\textwidth]{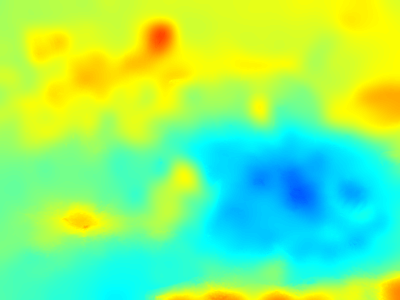}  &
\includegraphics[width=0.138\textwidth]{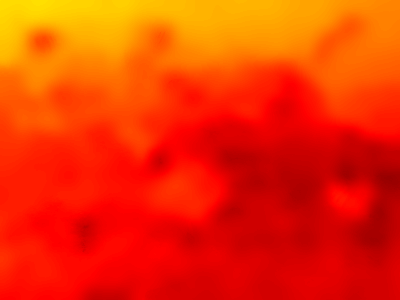} & \includegraphics[width=0.138\textwidth]{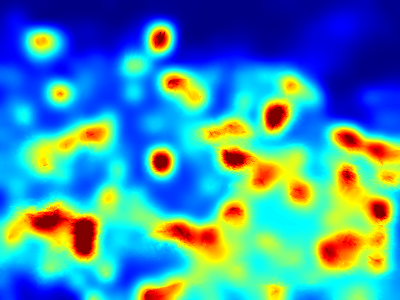} \\
\end{tabular}
\caption{Comparison of transmission maps $\textbf{T}$ estimated by outdoor, water-specific and texture priors for images captured in oceanic (rows 1-2)  and coastal (rows 3-4) waters.  Values vary in $[0,1]$ as follows: 0 \includegraphics[height=0.22cm]{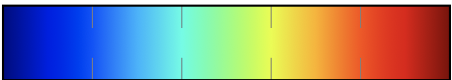} 1. The larger the value, the less compensated the colour as the the object is estimated to be closer to the camera (see Eq.~\ref{eq:modelJ}). Note that the three water-specific priors are designed for oceanic water. To facilitate the comparison, we selected the background light~\textbf{A} for all priors as the average intensity from a 15$\times$15 window that represents the water mass in the image (see the yellow rectangles in the first column). Priors marked with * requires a known background light~\textbf{A} to estimate~$\textbf{T}$; DCP: Dark Channel Prior~\cite{DCP}; Haze-line: Haze-line Prior~\cite{NonLocalImageDehazing}; UDCP: Underwater Dark Channel Prior~\cite{Drews2013UDCP},  CB: Carlevaris-Bianco et al.~\cite{CB2010InitialResult}; ARC: Automatic Red Channel~\cite{ARC}; Laplacian py.: Laplacian  pyramid~\cite{Blurriness_and_Light}.
}
\label{fig:: sample_transmission_map}
\end{figure*}

Many of the priors are employed to estimate~$\textbf{T}$ (Fig.~\ref{fig:: sample_transmission_map}). 
The {\em outdoor} colour priors proposed for dehazing are a popular choice\footnote{Hazy images are caused by the attenuation of light in air. As the wavelength dependence is negligible, the attenuation is non-selective.}~\cite{Emberton2018, BMVC_HL_2017, WCID}.
However, the main degradation source in hazy images is the veiling effect of haze~\cite{Narasimhan2002}, while that in water is the selective attenuation of light~\cite{DBLP:conf/cvpr/SchechnerK04}.
Therefore applying outdoor priors to underwater images generally misestimates the elements. 
The main {outdoor} priors exploiting the veiling effect are Dark Channel Prior (DCP)~\cite{DCP} and Haze-Line Prior~\cite{NonLocalImageDehazing}. DCP observes that the minimum intensity across the three colour channels (the Dark Channel) is usually close to zero in an haze-free image. In an hazy image, the Dark Channel's intensity is increasingly shifted towards that of the haze colour (represented by~$\textbf{A}$) with range. The smaller the Dark Channel value, the closer an object should be to the camera and hence the smaller the~$\textbf{T}$. However, when DCP is applied to oceanic water~\cite{WCID} to estimate~$\textbf{T}$ and~$\textbf{A}$, objects faraway from the camera are misestimated to be close, as most of the red light is attenuated. Alternatively, Haze-line Prior~\cite{NonLocalImageDehazing} makes use of the overall distribution of the RGB intensities, which can be well-represented by distinct RGB clusters~\cite{Orchard1991ColorQO}. The presence of haze shifts the RGB intensities towards the haze colour, thus forming a {haze-line} in the colour space. 
The smaller the difference between the pixel intensity's to the haze colour, the smaller the~$\textbf{T}$. However, Haze-line prior is not effective in water since the scene is under the attenuated illuminant and does not follow the same colour distribution as the in-air scene that are under an unattenuated light.

%
%
%
%

{\em Water-type specific} priors exploit the selective colour degradation in the water type. For example, priors specific to oceanic water consider red light as the most attenuated. Notable priors for oceanic water are the Underwater Dark Channel Prior~\cite{Drews2013UDCP}, the one proposed by Carlevaris-Bianco {et al.} (CB)~\cite{CB2010InitialResult} and the Automatic Red Channel~(ARC)~\cite{ARC}.
The Underwater Dark Channel Prior~(UDCP)~\cite{Drews2013UDCP} discards the most degraded red colour channel and states that objects closer to the camera, with less degraded colours, have low intensities in either of the blue or green channels. The further an object is, the higher the Underwater Dark Channel intensity. However, UDCP misestimates~$\textbf{T}$ for objects that are naturally high in blue and green colours. For example, a white object close to the camera is considered faraway as the pixels have a high intensity in all colour channels. Alternative to only using blue and green channels, Carlevaris-Bianco {et al.} (CB) proposed to consider the difference between the red colour channel and the less attenuated between blue and green colour channels~\cite{CB2010InitialResult}. 
This CB prior states that the closer the object is to the camera, the larger the colour difference; however, objects naturally low in red colour can still be misestimated.
%
%
%
The Automatic Red Channel Prior (ARC)~\cite{ARC} directly exploits the loss in red intensity, and the increase in blue and green intensity along the range. ARC misestimates objects naturally low in red intensity as located faraway. 




Textures are increasingly blurred along the range by scattering in {\em all} water types. This observation can be used as a prior to estimate~$\textbf{T}$. The degradation of texture can be measured by the variance of intensities in a local window~\cite{Blurriness_and_Light} or by the responses of high-frequency components in the image~\cite{NaturalStatistics_Aude_2003}, which can be identified with a band-pass Laplacian pyramid~\cite{BglightEstimationICIP2018_Submitted}. Texture priors are often combined with colour priors to estimate~$\textbf{T}$ more accurately. For example, a linear combination with UDCP~\cite{Blurriness_and_Light} averages out the error in either priors, whereas a combination that takes the maximum between texture prior and ARC~\cite{BglightEstimationICIP2018_Submitted} avoids underestimating~$\textbf{T}$.

Fig.~\ref{fig:: sample_transmission_map} shows the transmission maps estimated by the aforementioned priors in oceanic and coastal waters. Each of oceanic and coastal water has two images with different degree of colour cast. The images contain coral reefs and fishes at different ranges from the camera. 
All of these priors have their own shortcomings: DCP and UDCP misestimate the white regions on the fish to be faraway from the camera (row 1); Haze-line often estimates the water regions to be close to  the camera; CB overestimates the transmission map values (most regions are estimated to be close to the camera) and fails to estimate the transmission map in most of the colour-casted coastal water (row 4); ARC misestimates coral regions that are naturally high in red colour (row 1) and the water region in coastal water to  be close to the camera (row 4); the texture prior works for textured regions in both oceanic and coastal waters, but misestimates flat regions (row 1). 

To address the different degradation extents in each colour channel caused by selective attenuation, one can derive~$\textbf{T}$ for each colour channel  using Eq.~\ref{eq:: transmission_relationship_wavelength}. The ratio between attenuation coefficients (the exponent) can be extracted from empirical data of the 10 Jerlov water types directly~\cite{BMVC_HL_2017, seathru2019CVPR} or approximated using an empirical relationship between the diffuse attenuation coefficient and background light~\cite{Blurriness_and_Light, BglightEstimationICIP2018_Submitted, MIL}.  Instead of using different~$\textbf{T}$ for each colour channel, an alternative approach is to adjust the estimated~$\textbf{A}$ by the average colour channel intensity when using Eq.~\ref{eq:modelJ}~\cite{ARC}.

To remove the colour cast on objects and restore~$\textbf{J}_\textbf{0}$~(Eq.~\ref{eq:modelJ0}),the illuminant~$\textbf{E}$ can be estimated using colour statistics priors, such as the Gray World assumption~\cite{BUCHSBAUM1980_grayworld} and the Spatial Principal Component Analysis~\cite{Cheng:14ColourConstancy}, that are hypothesised on the reflectances of objects in air~\cite{Emberton2015, BMVC_HL_2017, seathru2019CVPR}. These colour statistics do not consider non-reflective objects (e.g. water mass) and can result in inaccurate estimations when pure water mass is visible in the image.~$\textbf{E}$ can also be estimated directly from the estimated~$\textbf{A}$~\cite{WCID}, which requires the generally unavailable diffuse attenuation coefficients of the water type to be known. Moreover, all these methods remove the colour cast in the entire scene, which could distort the water colour in the restored image.

Finally, using artificial light in-situ introduces a spherical overexposed area, which is not modelled in Eq.~\ref{eq:model}. To avoid overcompensating for these areas, a saturation prior can be used that considers these regions to have a similar intensity in each colour channel~\cite{ARC} or a comparison between the mean luminance of foreground and background~\cite{WCID}. In addition to these direct approaches, a combination of only considering the blue channel and UDCP can also handle these bright regions in oceanic water~\cite{Blurriness_and_Light}.

Table~\ref{Table: restoration} compares the restoration methods we discussed. Most methods are designed for oceanic waters and only a few methods also consider coastal waters~\cite{BMVC_HL_2017, seathru2019CVPR, MIL, Emberton2018}. Furthermore, only a few methods address the degradation caused by the illuminant attenuated through the depth under the water surface~\cite{BMVC_HL_2017,WCID,seathru2019CVPR,Emberton2018}.

\section{Enhancement}
\label{sec: methods_enhancement}
In this section, we discuss enhancement methods that improve the  quality images as perceived by humans or the performance of  computer vision tasks. 

%

%

\subsection{Human vision}\label{sec: methods_enhancement_human_vision}

%
Enhancement methods for human vision operate in the spatial or in the frequency domain and may use colour spaces that are appropriate for human perception, such as CIELab~\cite{CIELab} and HSL~\cite{HSV_patent_Valensi}. Most enhancement methods employ oceanology knowledge or human visual system preferences. Table~\ref{Table: enhancement} summarises the  underwater  enhancement  methods. 
%

\begin{table*}[t!]
\caption{Enhancement methods that aim to improve the quality of underwater images as perceived by humans. The methods improve colour and contrast, and denoise the image. The underwater-specific image quality assessment measure (UW-IQA) used to validate a method is shown in the evaluation column, along with the evaluations for task evaluations, subjective tests and generic IQA measures. 
}
\scriptsize
\begin{minipage}[b]{1.0\linewidth}
\centering
\setlength\tabcolsep{0.8pt}
\begin{tabular}{L{0.5cm}L{2.5cm}|ccccccccccc|ccccccccccccc}
\Xhline{3\arrayrulewidth}
\multicolumn{2}{c|}{{\textbf{Reference}}} &\multicolumn{1}{c}{{\textbf{Fusion}}} &\multicolumn{3}{c}{{\textbf{Optimised aspect}}}  &\multicolumn{2}{c}{\textbf{Domain}} &\multicolumn{3}{c}{\textbf{Colour space}}  &\multicolumn{2}{c|}{\textbf{Water type}} &\multicolumn{4}{c}{{\textbf{Evaluation}}} 
\\
& && \multicolumn{1}{c}{{Colour}} &\multicolumn{1}{c}{{Contrast}} 
& Denoise 
& \multicolumn{1}{c}{{Spatial}} & \multicolumn{1}{c}{{Frequency}} &  \multicolumn{1}{c}{{RGB}} &\multicolumn{1}{c}{{Lab}} &\multicolumn{1}{c}{{HSL}}   & Oceanic & Coastal   & Task & Subjective & Generic & UW-IQA \\ 
\Xhline{3\arrayrulewidth}
\cite{Fu_retinex-based_ICIP2014} & Fu {et al.} & &\checkmark & & &\checkmark & &\checkmark &\checkmark &  &\checkmark &\checkmark & & & & \\
\cite{Tang_retinex_video_SOVP2019} & Tang {et al.}&  &\checkmark & & &\checkmark & &\checkmark &\checkmark &  & & & &  & \checkmark &~\cite{WANG2018_NR_CFF}  \\
\cite{Gao2019_Adaptive_Retinex} & Gao {et al.} & &\checkmark & & &\checkmark & &\checkmark & &  &\checkmark &\checkmark &  \checkmark &  &
  &~\cite{UCIQE, UIQM_Panetta2016}\\
\cite{Zhang2017_multisclae_retinex} & Zhang {et al.} & &\checkmark & & &\checkmark & &\checkmark &\checkmark &  & & & & \checkmark &\checkmark & \\
\cite{Protasiuk2019_localcolourmapping} & Protasiuk {et al.} & &\checkmark & & &\checkmark & &\checkmark & &  & & &  &\checkmark & &~\cite{UIQM_Panetta2016} \\
\cite{Iqbal2010UCC} & Iqbal {et al.} & &\checkmark & & &\checkmark & &\checkmark & &\checkmark  &\checkmark &  & & & & \\
\cite{JTF} & Serikawa and Lu & &\checkmark & &\checkmark &\checkmark & &\checkmark & &  &\checkmark &  &  &  & \checkmark & \\
\cite{Prabhakar2010_wavelet_threshold} &  Prabhakar and Kumar & &\checkmark &\checkmark &\checkmark & &\checkmark & & &  & & &  &  & \checkmark & \\
\cite{Dai2019_uwlowlight} & Dai {et al.} & &\checkmark&\checkmark & &\checkmark &\checkmark &\checkmark & &  &\checkmark &\checkmark && & \checkmark  &~\cite{UCIQE, UIQM_Panetta2016}  \\
\cite{Ancuti2012} & Ancuti {et al.} &\checkmark &\checkmark &\checkmark & &\checkmark & &\checkmark &&  &\checkmark & & \checkmark &  & \checkmark & \\
\cite{Ancuti2018} & Ancuti {et al.} &\checkmark &\checkmark &\checkmark & &\checkmark & &\checkmark & &  &\checkmark &\checkmark &  \checkmark
 & 
 & \checkmark &~\cite{UCIQE, UIQM_Panetta2016, Lu2016_descattering_quality_ICIP} \\
\cite{Khan2016_wavelet} & Khan {et al.} &\checkmark  &\checkmark &\checkmark & &\checkmark &\checkmark &\checkmark & &\checkmark  & & & & 
& \checkmark & \\
\cite{VASAMSETTI2017_wavelet_variational} & Vasamsetti {et al.} &\checkmark &\checkmark &\checkmark & &\checkmark &\checkmark &\checkmark & &\checkmark  & & &  & & \checkmark & \\
\cite{WaterNET_TIP2019} & Li {et al.} &\checkmark &\checkmark  &  &  &\checkmark &   &\checkmark  &  & &   &  &   & &  \checkmark & \\
\Xhline{3\arrayrulewidth} 
\end{tabular}
\label{Table: enhancement}
\end{minipage}
\end{table*}

%


The colour cast can be removed by estimating the illuminant in the scene using white balancing algorithms, such as Gray World~\cite{BUCHSBAUM1980_grayworld} and MaxRGB~\cite{Land:71Max_RGB}.
White balancing algorithms may operate in the RGB~\cite{Tang_retinex_video_SOVP2019, Ancuti2012,WaterNET_TIP2019} or CIELab~\cite{Fu_retinex-based_ICIP2014,Zhang2017_multisclae_retinex} colour spaces.
%
However, white balancing alone cannot handle the range-dependent degradation extent. The colour degradation along the range can be compensated by assuming  that the colour channels in the opponent colour pairs~\cite{OpponentColour_Hering} have similar intensity values in an undegraded image~\cite{Ancuti2018}. 
%
To increase contrast, approaches such as gamma correction~\cite{Ancuti2012,Ancuti2018,  WaterNET_TIP2019}, histogram equalisation (e.g. CLAHE~\cite{CLAHE_Zuiderveld}) in RGB and HSL colour spaces~\cite{Khan2016_wavelet}, and adjusting each colour channel's value according to the overall colour cast factor~\cite{VASAMSETTI2017_wavelet_variational} are also used. To address the blurring caused by scattering, unsharp masking can be employed to emphasise the edges~\cite{unsharpmask_adaptive_Polesel_2000}   
in the spatial domain or by directly enhancing the edges, represented by high-frequency components, in the frequency domain~\cite{Dai2019_uwlowlight}. A degradation linked to sharpness is the noise, normally caused by backscattering or the image acquisition process. 
While denoising can be achieved by low-pass filtering in the frequency domain~\cite{Prabhakar2010_wavelet_threshold}, this might introduce additional blurring to the edges. A balance between denoising and edge preservation can be achieved with a joint trilateral filter~\cite{JTF}.





Fusion is also used to combine differently filtered versions of a degraded image~\cite{Ancuti2012,Ancuti2018,  WaterNET_TIP2019}, in order to apply the most suitable enhancement filter for different image regions. 
%
The inputs to a fusion algorithm can be combined to optimise contrast, saturation and edge details~\cite{Ancuti2018}, and also to suppress over- or under-exposure~\cite{Ancuti2012}. The inputs can be combined using weights estimated by neural networks trained on a dataset  collected from subjective tests~\cite{WaterNET_TIP2019} or derived from saliency~\cite{Ancuti2012, Ancuti2018}. Fusion can also be performed on the decomposition of the wavelet coefficients in frequency space, before converting the enhanced image back to RGB colour space~\cite{VASAMSETTI2017_wavelet_variational,Khan2016_wavelet}.

\subsection{Computer vision tasks}
\label{sec: methods_enhancement_application}

Underwater computer vision tasks include the detection, tracking, and recognition of fish~\cite{chuang2014supervised,wang2016closed,Lakshmi2020_fishDetCountClass, DetectingFishInLowQualityVideo_Concetto_2008}, other sea animals~\cite{corgnati2014automated, mehrnejad2014towards}, and plankton~\cite{hirata2016plankton};  the classification of coral reefs~\cite{soriano2001image} and the positioning of vehicles~\cite{noaa_rov}.  These tasks enable marine species discovery~\cite{telegraph_new_species}, coral preservation~\cite{CV_analysis_behaviour, NYTimes_CoralZoomIn_2016, Intel_Coral_2020}, ocean exploration with unmanned vehicles~\cite{noaa_rov} and archaeological excavation~\cite{NG_uw_archaeology, Advances_Robotics_oceanExploration_2000},. However,  the algorithms for these tasks are generally designed for in-air images and hence do not perform well on the degraded underwater images~\cite{feautreDetection_BMVC_2002}. 

Underwater computer vision tasks may focus on objects of interest (detection, counting, recognition and tracking) or consider the whole scene (3D scene reconstruction).
{\em Objects of interest} include marine animals (detection)~\cite{mehrnejad2014towards,Lakshmi2020_fishDetCountClass}, corals (recognition), cables~\cite{ortiz2002vision}, pipelines~\cite{Foresti2000}  or other man-made objects~\cite{lee2012vision, Rizzini2015, Kallasi15}. 
{\em Scene reconstruction} is the basis for scene mapping, which is important for archaeology and excavation as knowing the scene in advance can optimise the time divers stay underwater.
Scene reconstruction typically involves establishing the correspondence between two images by matching a set of detected feature points~\cite{3Dreconstruction_Principles}. 

Most methods aim to make the condition of the image closer to those taken in-air, by correcting the colour degradations and deblurring the image.  Colour degradations can be compensated by removing the colour cast with white balancing~\cite{3DFromRobots_JohnsonRobertson_2010,Kallasi15} and increasing the contrast with histogram equalisation~\cite{skarlatos2012open}. A few methods also address the wavelength-dependent attenuation   by exponentially increasing the intensity in each colour channel with respect to the attenuation coefficients obtained from empirical experiments~\cite{Foresti2000, lee2012vision}. Moreover, image noise caused by backscattering of suspended particles can be removed using bilateral filter~\cite{Lopez2020_animalClassification_Cabled} or morphological opening and closing of the segmented foreground binary image in object recognition~\cite{Lakshmi2020_fishDetCountClass}, whereas the edges can be enhanced by unsharp masking in the luminance channel of the CIELab colour space~\cite{Rizzini2015}, or with the Wallis filter~\cite{WallisFilter1976,skarlatos2012open}. Table~\ref{Table: applications} compares the enhancement targets of the aforementioned methods. 

\begin{table}
\caption{Enhancement methods that aim to improve the performance of a computer vision algorithm by enhancing colour (col.), edges, or by denoising (den.). Computer vision tasks include object detection (det.), counting (count.), recognition (recog.), tracking (track.) and scene reconstruction (reconst.).}
\centering
\setlength\tabcolsep{1pt}
\scriptsize
\begin{tabular}{ll|ccc|cccccL{5cm}ccccccccccccccc|c}
\Xhline{3\arrayrulewidth}
\multicolumn{2}{c|}{\multirow{3}{*}{\textbf{Reference}}}  & \multicolumn{3}{c|}{\textbf{Enhancement}}  &\multicolumn{5}{c}{\textbf{Task}}\\
 &  & col. & edge & den. & det. & count. & recog. & \multicolumn{1}{c}{track.} & reconst.\\
\Xhline{3\arrayrulewidth}
\cite{Lakshmi2020_fishDetCountClass} & Lakshmi and Krishnan &  &  &\checkmark &\checkmark &\checkmark &\checkmark &   &  \\
\cite{mehrnejad2014towards} & Mehrnejad {et al.} &\checkmark &  &  &\checkmark &  &  &   & \\
\cite{Rizzini2015} & Rizzini {et al.} &\checkmark &\checkmark &  &\checkmark &  &  &  & \\
\cite{Foresti2000} & Foresti {et al.} &\checkmark &  &  &\checkmark &  &  &  & \\
\cite{Kallasi15} & Kallasi {et al.} &\checkmark &  &  &\checkmark &  &  &  & \\
\cite{lee2012vision} & Lee {et al.} &\checkmark &  &  &\checkmark &  &  & \checkmark & \\
\cite{Lopez2020_animalClassification_Cabled} & Lopez-Vazquez {et al.} &\checkmark &  &\checkmark &  &  &\checkmark &  & \\
\cite{skarlatos2012open} & Skarlatos {et al.} &\checkmark &\checkmark &  &  &  &  &  &\checkmark \\
\cite{3DFromRobots_JohnsonRobertson_2010} & Johnson-Robertson {et al.} &\checkmark &  &  &  &  &  &  &\checkmark\\
\Xhline{3\arrayrulewidth}
\end{tabular}%
\label{Table: applications}
\end{table}

Unlike methods for restoration (Sec.~\ref{sec: methods_restoration}) and enhancement for human vision (Sec.~\ref{sec: methods_enhancement_human_vision}), enhancement methods for computer vision tasks are still at an early stage. Most methods combine image filtering techniques in the approach but without validating the method in real life deployment. Due to the lack of ground-truths, these methods are mainly evaluated with laboratory recreated environments. 

\section{Learning-based methods}
\label{sec: neural_network}

\begin{table*}[t!]
\centering
\begin{threeparttable}
\caption{Learning-based methods. The target images can be synthesised using a physics-based model (Physics) for the colour degradation along the range (R) or along both range and depth (R+D), blurring along the range (S), synthesised using neural network (NN), or collected via subjective tests (Test). Network types include convolutional neural networks (CNN), Auto-encoders, Generative Adversarial Networks (GAN) and  Cycle-Consistent Adversarial Networks (CycleGAN). Note that we list here only the the networks that filter a degraded image but not those synthesise the target images. In addition to architecture-specific terms (Additional), the terms of the loss function encourage similarities between the filtered and target image, as measured in terms of colour differences, using the~$L_1$ or~$L_2$ norm, or the mean square error (MSE), deblurring through texture details with  Structural SIMmilarity (SSIM)~\cite{wang2004ssim}, Multi-scale SSIM (MS-SSIM)~\cite{MSSIM} or gradient difference (Grad-Diff)~\cite{image_Gradient_loss_2016_Mathieu}, or features extracted by trained neural network using a perceptual loss~\cite{perceptual_loss}. All discriminators in the GANs are PatchGAN~\cite{pix2pix2017}. All GANs and CycleGANs include  a GAN loss~\cite{GANGoodfellow} or cycle-consistency loss~\cite{CycleGAN_2017_Zhu}, respectively. The underwater-specific image quality assessment measure (UW-IQA) used to validate a method is shown in the evaluation column, along with the evaluations for task evaluation, subjective tests and generic IQA measures. }
\scriptsize
\centering
\setlength\tabcolsep{1.4pt}  
\begin{tabular}{ll|l|lcc|c|cccc|cccc}
\Xhline{3\arrayrulewidth}
 \multicolumn{2}{c|}{\textbf{Reference}} & \multicolumn{1}{c|}{\textbf{Acronym}} &  \multicolumn{3}{c|}{\textbf{Target images}} & \textbf{Type} & \multicolumn{4}{c|}{\textbf{Loss function}} & \multicolumn{4}{c}{\textbf{Evaluation}} \\
 &  &  & {Physics} & {NN} & {Test} &  & {Colour} & {Deblur} & {Perceptual} & {Additional} & {Task} & {Subjective} & {Generic} & {UW-IQA} \\
\Xhline{3\arrayrulewidth}  \cite{Li2018WaterGANUG} & Li {et al.} & WaterGAN & R &  &  & CNN  &~$L_2$ &  &  &  & \checkmark & \checkmark &  &  \\
\cite{Anwar2018DeepUI_UWCNN} & Li et al. & UWCNN & R &  &  & CNN &~$L_1$ & SSIM &  &  &  & \checkmark &  &  \\
\cite{UIEDAL_Uplavikar_2019_CVPRW} & Uplavikar et al. & UIE-DAL & R &  &  & Auto-encoder & MSE &  &  &  & \checkmark &  &  &  \\
\cite{Yu2019_UnderwaterGAN_PRIF} & Yu et al. & Underwater-GAN & R &  &  & GAN &  &  & \checkmark & GAN &  &  & \checkmark &  \\
\cite{Li_2020_Cast-GAN} & Li and Cavallaro & Cast-GAN & R+D, S &  &  & GAN &~$L_2$ & SSIM & \checkmark & GAN &  & \checkmark &  &  \\
\cite{Hashisho_UNet_ISISPA_2019} & Hashisho et al. & UDAE &  & \checkmark &  &  Auto-encoder  &~$L_1$ & MS-SSIM &  &  &  &  &\checkmark &  \\
\cite{DenseGAN_Guo_2020} & Guo et al. & UWGAN* &  & \checkmark &  & GAN &  &  &  & GAN & \checkmark &  &  & \cite{UCIQE, UIQM_Panetta2016} \\
\cite{UGAN_Fabbri_2018} & Fabbri et al. & UGAN &  & \checkmark &  & GAN &~$L_1$ & Grad-Diff &  & GAN & \checkmark &  &  &  \\

\cite{li2019fusion} & Li et al. & FGAN &  & \checkmark &  & GAN &~$L_1$ &  &  & GAN &  &  &  & \cite{UIQM_Panetta2016, UCIQE} \\
\cite{Islam2019FastUI4ImprovedPerception} & Islam et al. & FUnIE-GAN &  & \checkmark &  \checkmark & GAN &~$L_1$ &  & \checkmark & GAN &  & \checkmark &  & \cite{UIQM_Panetta2016} \\
\cite{P2P_Sun_IET_2019} & Sun et al. & P2P &  &  & \checkmark & CNN & MSE &  &  &  &  &  & \checkmark &  \\
\cite{UWGAN_emerging_SPL_2018} & Li et al. & UWGAN* &  &  & \checkmark & CycleGAN &  & SSIM &  & CycleGAN & \checkmark & \checkmark &  &  \\
\cite{MCycleGAN_Lu_Laser2019} & Lu et al. & M-Cycle GAN &  &  & \checkmark & CycleGAN &  & MS-SSIM &  & CycleGAN &  &  &  & \cite{UIQM_Panetta2016} \\
\Xhline{3\arrayrulewidth}
\end{tabular}
\label{Tab: learning-based}
\begin{tablenotes}
\setlength{\columnsep}{0.8cm}
\setlength{\multicolsep}{0cm}
\scriptsize
\item[*] 
\textcolor{black!80}{Note that the two networks are different, even if they have the same acronym. }
\end{tablenotes}
\end{threeparttable}
\end{table*}

Learning-based methods that use neural networks have been increasing employed for underwater images~\cite{Li2018WaterGANUG,UIEDAL_Uplavikar_2019_CVPRW,Yu2019_UnderwaterGAN_PRIF,Hashisho_UNet_ISISPA_2019,Islam2019FastUI4ImprovedPerception,Li_2020_Cast-GAN,P2P_Sun_IET_2019,UGAN_Fabbri_2018,Anwar2018DeepUI_UWCNN,DenseGAN_Guo_2020,URCNN_ICIP2018_Hou, UWGAN_emerging_SPL_2018, MCycleGAN_Lu_Laser2019, li2019fusion}.
In this section, we discuss the {target} images used to train neural networks, the chosen network architectures and loss functions. 
Table~\ref{Tab: learning-based} compares learning-based methods and their features.

{\em Target images}  used in training may be images free of degradation to be used for underwater image restoration~\cite{Anwar2018DeepUI_UWCNN, Li2018WaterGANUG, Li_2020_Cast-GAN, UIEDAL_Uplavikar_2019_CVPRW,Yu2019_UnderwaterGAN_PRIF} or images selected for quality to be used for underwater image enhancement~\cite{Islam2019FastUI4ImprovedPerception, DenseGAN_Guo_2020, ICIP2017UWCNN, UGAN_Fabbri_2018, Hashisho_UNet_ISISPA_2019, li2019fusion, WaterNET_TIP2019, MCycleGAN_Lu_Laser2019, UWGAN_emerging_SPL_2018}. Most methods' training require a target image for each of the degraded image  (supervised)~\cite{Li2018WaterGANUG,UIEDAL_Uplavikar_2019_CVPRW,Yu2019_UnderwaterGAN_PRIF,Hashisho_UNet_ISISPA_2019,Islam2019FastUI4ImprovedPerception,Li_2020_Cast-GAN,P2P_Sun_IET_2019,UGAN_Fabbri_2018,Anwar2018DeepUI_UWCNN,DenseGAN_Guo_2020,URCNN_ICIP2018_Hou, li2019fusion}, whereas a few methods only require specifying the degraded and target image {\em datasets}  but not the correspondence between images (weakly supervised)~\cite{MCycleGAN_Lu_Laser2019, UWGAN_emerging_SPL_2018}. 

Because of the difficulty of capturing the underwater scene without the water medium, target images free of degradation are difficult to obtain. Degraded images are instead synthesised using physics-based models~\cite{Anwar2018DeepUI_UWCNN, Li2018WaterGANUG, Li_2020_Cast-GAN}. The corresponding target images are taken from existing datasets, such as~\mbox{NYU-D} dataset~\cite{Silberman:NYUD:ECCV12}, with natural indoor scenes, and SINTEL~\cite{SINTEL_Background},  with computer-generated images (see Sec.~\ref{sec: Datasets}). 
Several methods use images from~\mbox{NYU-D} dataset as  target~$\textbf{J}$ and synthesise~$\textbf{I}$ with Eq.~\ref{eq:model}~\cite{Anwar2018DeepUI_UWCNN, Li2018WaterGANUG, Yu2019_UnderwaterGAN_PRIF, UIEDAL_Uplavikar_2019_CVPRW}. Because the attenuated illuminant and scattering are not modelled, networks trained on these images only compensate for the range but not the depth nor blurred textures. Furthermore, the limited colour palette  of~\mbox{NYU-D} images results in a similarly  limited colour palette in the processed images~\cite{UGAN_Fabbri_2018}.

To compensate for the degradation through the depth, the degraded image~$\textbf{I}$ can be synthesised with~\mbox{$\textbf{J} =\textbf{E}\cdot\textbf{J}_\textbf{0}$} in Eq.~\ref{eq:model}, where the attenuated illuminant~$\textbf{E}$ at depth is synthesised using Eq.~\ref{eq:ambLight}~\cite{Li_2020_Cast-GAN}. Moreover, the range-dependent scattering can be modelled by applying successive Gaussian blurring (Eq.~\ref{eq:: width_sigma}) to~$\textbf{J}$~\cite{Li_2020_Cast-GAN}. To obtain a more diverse colour palette in the target images, $\textbf{J}_\textbf{0}$ can be derived from SINTEL~\cite{SINTEL_Background} by modifying the colours of each object, so that they follow the colour distributions of natural, in-air images (e.g.~the Berkeley Segmentation Dataset~\cite{Berkeley_Segmentation_Dataset})~\cite{Li_2020_Cast-GAN}. 
The aforementioned restoration methods synthesise~$\textbf{I}$ for the 10 Jerlov water types using the diffuse attenuation~\cite{Jerlov} of the water type~\cite{Anwar2018DeepUI_UWCNN, Li2018WaterGANUG, Yu2019_UnderwaterGAN_PRIF, UIEDAL_Uplavikar_2019_CVPRW}, or also
the beam attenuation coefficients~\cite{Solonenko:15:inherit} from empirical data~\cite{Li_2020_Cast-GAN}.

Target images can also be selected  via subjective tests~\cite{Yu2019_UnderwaterGAN_PRIF,  ICIP2017UWCNN, UGAN_Fabbri_2018,  li2019fusion, WaterNET_TIP2019}. 
For example, the target images used to train  WaterNET~\cite{WaterNET_TIP2019} are the most chosen (by 50 participants) among images filtered with 12 enhancement methods. Each participant was asked  '{\em which one is better}' in a double-stimulus subjective test, where the degraded image was also shown but could not be selected. 
{Alternatively}, the target images can be synthesised after learning the enhancement with a Generative Adversarial Network~\cite{GANGoodfellow}, trained on small datasets whose images are annotated by humans as 'good' or 'bad' qualities~\cite{Hashisho_UNet_ISISPA_2019, Islam2019FastUI4ImprovedPerception}. 
Moreover, in weakly supervised training, the selection of target image is no longer restricted to images capturing the same scene as the degraded image. The target image datasets can then contain 'less degraded' underwater images, chosen from a  subjective test~\cite{UWGAN_emerging_SPL_2018}, or even in-air images~\cite{MCycleGAN_Lu_Laser2019}.  

Underwater neural networks trained in a supervised manner can employ end-to-end Convolutional Neural Networks (CNNs) that  process a degraded image and use skip connections~\cite{Anwar2018DeepUI_UWCNN, li2019fusion,P2P_Sun_IET_2019}. 
The CNNs can also be guided by additional information such as the range, which can be estimated by earlier network layers~\cite{Li2018WaterGANUG}. 
In this case, the error in range estimation may accumulate through the subsequent layers and affect the output image.  Auto-encoders for underwater image filtering typically use the U-Net~\cite{U-net_2015_Ronneberger} architecture with skip connections between the encoder and the decoder to maintain contextual information in the reconstructed image~\cite{UIEDAL_Uplavikar_2019_CVPRW,Hashisho_UNet_ISISPA_2019}. Several networks employ Generative Adversarial Network (GAN~\cite{GANGoodfellow}) framework~\cite{Yu2019_UnderwaterGAN_PRIF,Islam2019FastUI4ImprovedPerception,Li_2020_Cast-GAN,UGAN_Fabbri_2018,Anwar2018DeepUI_UWCNN,DenseGAN_Guo_2020, li2019fusion}. The GAN consists of a generator network, that filters the degraded image, and a discriminator network, that determines whether the an image is the target image or has been filtered by network.    
 Moreover, networks trained in a weakly supervised manner typically use a CycleGAN~\cite{CycleGAN_2017_Zhu}, that consists of a first GAN that learns the filtering from the degraded to the target image set, and a second GAN that learns to process the filtered image back to the original degraded image. However, without explicitly stating the target image of each degraded image, CycleGAN is less stable and more difficult to train~\cite{CycleGAN_2017_Zhu}. 

{\em Loss functions} for underwater image filtering generally measure the similarity between a target image and the network-filtered image. The similarity is quantified through colour differences or by comparing the structure (texture) in  local windows.  Colour differences are measured pixel-by-pixel using the~$L_1$ norm~\cite{Hashisho_UNet_ISISPA_2019,Anwar2018DeepUI_UWCNN,Islam2019FastUI4ImprovedPerception,UGAN_Fabbri_2018,li2019fusion}, the~$L_2$ norm~\cite{Li2018WaterGANUG, Hashisho_UNet_ISISPA_2019} or the mean square error~\cite{UIEDAL_Uplavikar_2019_CVPRW, P2P_Sun_IET_2019}. Image structures are measured using Structural SIMilarity~(SSIM~\cite{wang2004ssim} and multi-scale SSIM~\cite{MSSIM})~\cite{Hashisho_UNet_ISISPA_2019, Anwar2018DeepUI_UWCNN, Li_2020_Cast-GAN,MCycleGAN_Lu_Laser2019,UWGAN_emerging_SPL_2018} or the image gradient~\cite{image_Gradient_loss_2016_Mathieu} difference~\cite{UGAN_Fabbri_2018}. Furthermore, a perceptual loss function~\cite{perceptual_loss} is often used to suppress artifacts that are common in images processed by neural network~\cite{Li_2020_Cast-GAN, Islam2019FastUI4ImprovedPerception,Yu2019_UnderwaterGAN_PRIF}. Finally, networks employing a GAN or CycleGAN framework include the corresponding GAN or CycleGAN loss term~\cite{Islam2019FastUI4ImprovedPerception, P2P_Sun_IET_2019, UWGAN_emerging_SPL_2018, MCycleGAN_Lu_Laser2019}.  A review of underwater image filtering entirely dedicated to neural networks is presented by Anwar and Li~\cite{Anwar_Li_2020_Survey}.

\newcommand\WholeCell[1]{%
  \multicolumn{2}{c}{{#1}}
}
\newcommand\InfoCell[1]{%
  \multicolumn{1}{c}{\footnotesize \cellcolor{black}\textcolor{white}{#1}}
}

\section{Datasets}
\label{sec: Datasets}

Datasets are indispensable for the development of  methods~\cite{UnbiasedLookonDataset, DeeperLookDatasetBias_Tommasi2017} and the evaluation of their performance~\cite{LIVE_Image_R2}. 
In this section, we discuss datasets categorises into natural (Sec.~\ref{sec:natDatasets}) and chemically or digitally degraded (Sec.~\ref{sec:artDatasets}). Table~\ref{Table: datasets} summarises the datasets and the associated information. Fig.~\ref{fig:: natural_dataset_images} shows examples of notable datasets.

\subsection{Natural datasets}
\label{sec:natDatasets}

\begin{table*}
\centering 
\caption{Underwater datasets categorised as natural, digitally degraded and chemically degraded. 
The website hosting each dataset is accessible via the hyperlink embedded in the name of the dataset.
}
\scriptsize
\centering 
\setlength\tabcolsep{3.8pt}
\begin{tabular}{lllllrc}
\Xhline{3\arrayrulewidth}
\multicolumn{1}{c}{\textbf{Category}} &\multicolumn{1}{c}{\textbf{Reference}}  & \multicolumn{1}{c}{\textbf{Dataset}} 
 &\multicolumn{1}{c}{\textbf{Annotation/ Information}} &
 \multicolumn{1}{c}{\textbf{Content}} &~\textbf{Number of images} 
\\ 
\Xhline{3\arrayrulewidth}
\multirow{7}{*}{{Natural}}
&\multirow{3}{*}{\cite{liu19_RUIE}}
 & 
\href{https://github.com/dlut-dimt/Realworld-Underwater-Image-Enhancement-RUIE-Benchmark}{RUIE UIQS} & Quality determined by IQA~\cite{UCIQE} &\multirow{3}{*}{Marine life}  & 3,630 
\\ 
 & 
& \href{https://github.com/dlut-dimt/Realworld-Underwater-Image-Enhancement-RUIE-Benchmark}{RUIE UCCS} & Colour cast &  & 300 
\\ 
 & 
& \href{https://github.com/dlut-dimt/Realworld-Underwater-Image-Enhancement-RUIE-Benchmark}{RUIE UHTS} & Marine life location \& type  & & 300 
\\ 
\cline{2-6}
&~\cite{Islam2019FastUI4ImprovedPerception} & 
\href{https://li-chongyi.github.io/proj_benchmark.html}{EUVP unpair} & Quality determined by subjective test & Marine life and divers   &  6,665   
\\
&~\cite{WaterNET_TIP2019} & 
\href{http://irvlab.cs.umn.edu/resources/euvp-dataset}{UIEB} &\textit{Preferred} image  by subjective test & Marine life and divers   & 890  
\\ 
&~\cite{CLEF_initiative}  & \href{https://www.imageclef.org/lifeclef/2015/fish}{LifeCLEF Fish/ Coral task}  & Fish/ Coral species and locations & Fish/ Coral   &~Over 20,000  \\ 
&~\cite{BermanArvix_UWHL_dataset} & 
\href{http://csms.haifa.ac.il/profiles/tTreibitz/datasets/ambient_forwardlooking/index.html}{SQUID} & Range, colour chart & River bed, rocks, wrecks    & 57 
\\
&~\cite{seathru2019CVPR}  & 
\href{http://csms.haifa.ac.il/profiles/tTreibitz/datasets/sea_thru/index.html}{Sea-thru} & Range, colour chart & River bed, rocks, corals  & 1,157 
\\
\Xhline{2\arrayrulewidth}
\multirow{3}{*}{{Digitally degraded}} &~\cite{Anwar2018DeepUI_UWCNN} & 
\href{https://li-chongyi.github.io/proj_underwater_image_synthesis.html}{PR 2019 Synthetic}  & Reference image  & Indoor scenes & 14,490 \\
&~\cite{UGAN_Fabbri_2018} & \href{http://irvlab.cs.umn.edu/enhancing-underwater-imagery-using-generative-adversarial-networks}{Fabbri {et al.}} & Reference image & Marine life and divers  & 6,143  
\\ 
&~\cite{Islam2019FastUI4ImprovedPerception} & 
\href{https://li-chongyi.github.io/proj_benchmark.html}{EUVP pair} & Reference image  & Marine life and divers    &  24,848  
\\
\Xhline{2\arrayrulewidth}
{{Chemically degraded}} 
&~\cite{Duarte2016} & 
\href{http://amandaduarte.com.br/turbid/}{TURBID}   & Reference image &  Man-made objects  & 112 
\\
\Xhline{3\arrayrulewidth}
\end{tabular}
\label{Table: datasets}
\end{table*}
 
Natural datasets capture scenes that are naturally degraded by the water medium. The annotations of these datasets include the perceived quality of images,  images  characteristics,  or computer  vision  results (e.g.~bounding boxes and segmentation masks).

The Real-world Underwater Image Enhancement (RUIE) dataset~\cite{liu19_RUIE} consists of three subsets, namely Underwater Image Quality Set (UIQS), Underwater Colour Cast Set (UCCS) and Underwater Higher-level Task-driven Set (UHTS). The images were captured by 24 cameras mounted at fixed locations over three hours. The changing tide and sunlight over time create varying scenes' depth and illumination conditions in the dataset. UIQS contains 3,630 images for evaluation purposes, separated into 5 groups according to their quality quantified by an underwater image quality measure (UCIQE~\cite{UCIQE}, see Sec.~\ref{sec: evaluation}) (Fig.~\ref{fig:: natural_dataset_images}(a)).  UCCS consists of 300 images from UIQS grouped by the overall colour cast, measured by the average b channel intensity in the CIELab colour space (Fig.~\ref{fig:: natural_dataset_images}(d)). UIQS and UCCS are suitable for evaluating the filtered images' quality.
UHTS contains 300 images with bounding boxes of three marine species (scallop, sea cucumbers and sea urchins), and is mainly suitable for evaluating the performance of specific applications such as detection and classification.

The Enhancement of Underwater Visual Perception dataset (EUVP) unpair collection~\cite{Islam2019FastUI4ImprovedPerception} and the Underwater Image Enhancement Benchmark (UIEB)~\cite{WaterNET_TIP2019} are  annotated for image quality.
The EUVP unpair collection consists of 6,665 images~\footnote{There is a mismatch between the image numbers mentioned in the paper and the website, here we consider the numbers in website.}~(Fig.~\ref{fig:: natural_dataset_images}(b)), each annotated as 'good' or 'bad' quality by 6 participants in a subjective test that asked {\em 'whether the background and foreground are identifiable'}. 
The paired images are provided with reference images generated by a neural network that learnt the domain transformation between the undegraded and distorted images from a subjective test, but no analysis was performed on the fidelity of these images with respect to the human perception.
UIEB presents a large-scale natural dataset with corresponding reference images that can be used to train neural networks or evaluate IQA measures~(Fig.~\ref{fig:: natural_dataset_images}(c)). The dataset contains 890 images collected from those used in publications  and online repositories or captured by the authors. 
The {\em 'reference'} image is the most chosen by 50 participants in a subjective test among a set of images processed by 12 filtering methods (9 methods were proposed for underwater images~\cite{Ancuti2018, Fu_retinex-based_ICIP2014, Drews2013UDCP, Hybrid_LI_2017,two-step_Fu_2017, generalised_dcp_Peng_2018, ARC, MIL, Blurriness_and_Light}, 2 for dehazing ~\cite{DCP, Ren-MSCNN-ECCV-2016} and 1 image filtering software~\cite{dive_plus}). Participants were asked '{\em which one is better}' in a double-stimulus of filtered images, where the degraded image is also shown (but cannot be selected). 
Note that the selection of reference images was not confirmed by a statistical significance test.  

The application-specific LifeCLEF Fish/ Coral task~\cite{CLEF_initiative} provides annotations of the species and their locations. The Fish task dataset comprises 93 videos manually annotated with bounding boxes for the fish and a label for their species. The Coral task dataset contains the segmentation masks of 13 types of coral substrates in the image. 

The  Stereo Quantitative Underwater Image Dataset (SQUID)~\cite{BermanArvix_UWHL_dataset} and Sea-thru~\cite{seathru2019CVPR} include scene range information and colour charts to facilitate the evaluation of colour accuracy. SQUID~\cite{BermanArvix_UWHL_dataset} has 4 sets of images captured by a stereo camera with colour charts placed in the scene at different ranges. One set of the images captures a shipwreck and three sets capture coral reefs. The range information in SQUID was calculated from a checkerboard of known size with the MATLAB stereo calibration toolbox and Epicflow~\cite{EpicFlow}. However, some ranges are missing due to occlusions~(Fig.~\ref{fig:: natural_dataset_images}(e)). 
Sea-thru~\cite{seathru2019CVPR} comprises 5 sets totalling 1,157 images in clear and turbid water, with different viewing angles, and at ranges between 4 and 10 metres~(Fig.~\ref{fig:: natural_dataset_images}(f)). The ranges were obtained, using Structure from Motion algorithm, from stereo images where an object of known size is placed in the scene\footnote{An object of known size is required since Structure from Motion only estimates the range in relative scales.}. 

\subsection{Digitally or chemically degraded datasets}
\label{sec:artDatasets}

Degradations can be generated  digitally or with chemical reagents added to a water body. Digitally degraded images are synthesised with physics-based models (e.g. Eq.~\ref{eq:model}) or neural networks. Chemically degraded datasets add reagents to pure water to mimic degradations occurring naturally underwater. These datasets have reference degradation-free images that can be used for quantitative evaluation.

\begin{figure}[t!]
\setlength\tabcolsep{0.3pt}
\centering
\renewcommand{\arraystretch}{0.05}
\begin{tabular}{ccccc}
\begin{tabular}{ccccc}
\includegraphics[width=0.24\columnwidth]{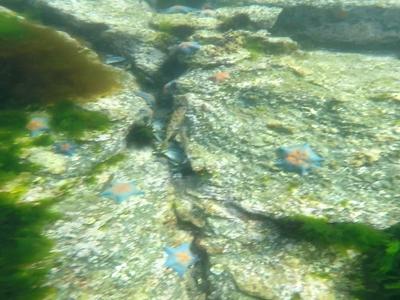} 
&\includegraphics[width=0.24\columnwidth]{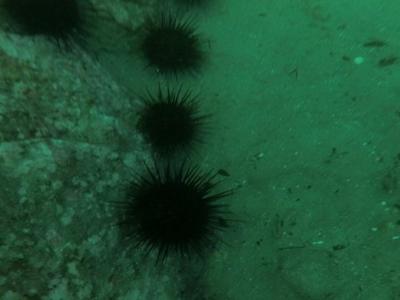} & & 
\includegraphics[width=0.24\columnwidth]{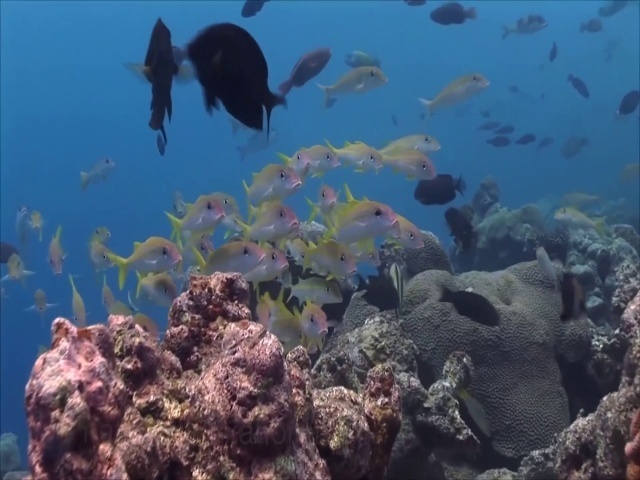}&
\includegraphics[width=0.24\columnwidth]{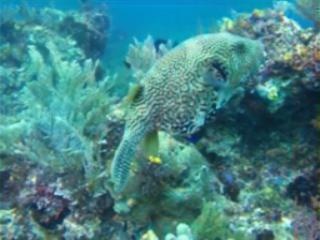}
\\ 
\InfoCell{best quality}  & \InfoCell{worst quality} & & \InfoCell{good quality} &    \InfoCell{bad quality}
\\\\\\\\
\WholeCell{(a) RUIE UIQA~\cite{liu19_RUIE}} & & \WholeCell{(b)  {EUVP unpair}~\cite{Islam2019FastUI4ImprovedPerception}} \\
\vspace{3pt}
\end{tabular}
\\
\begin{tabular}{ccccc}
 \includegraphics[width=0.24\columnwidth]{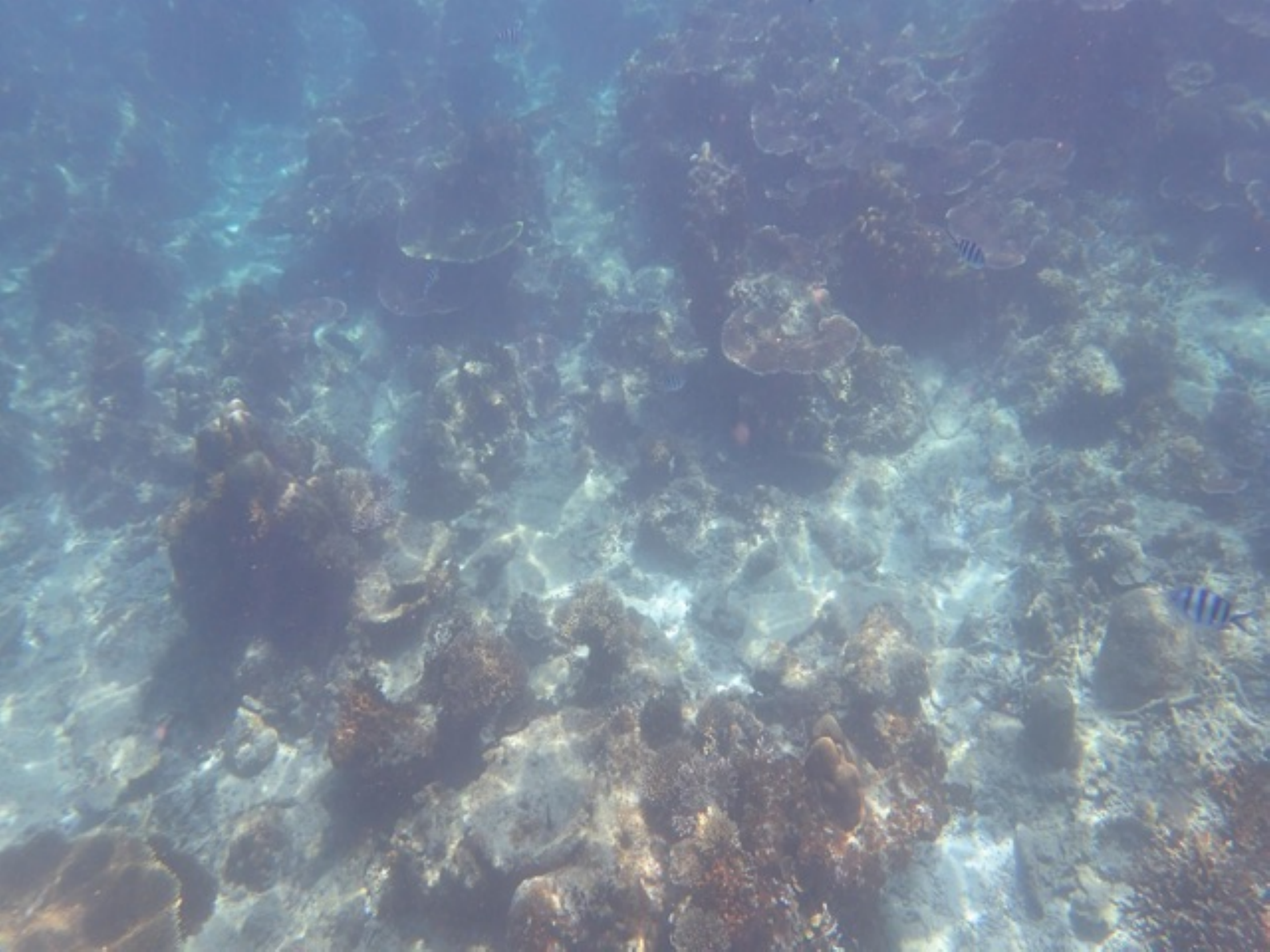} 
 &
\includegraphics[width=0.24\columnwidth]{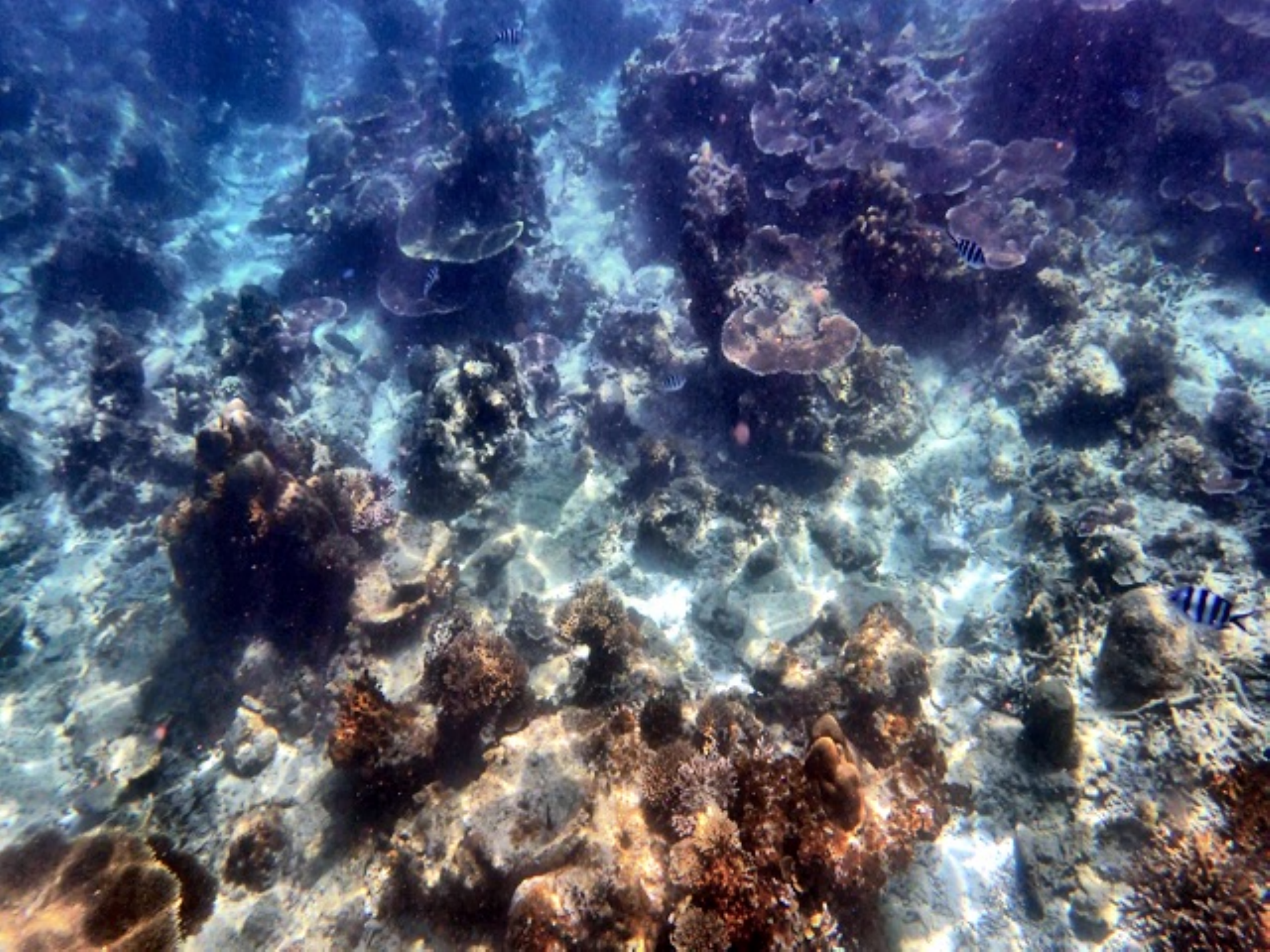}
& & 
\includegraphics[width=0.24\columnwidth]{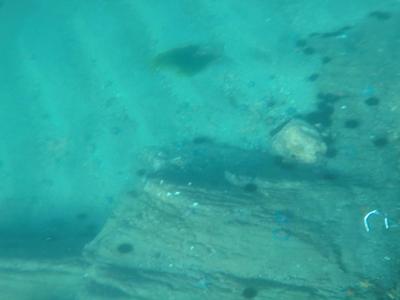} & 
\includegraphics[width=0.24\columnwidth]{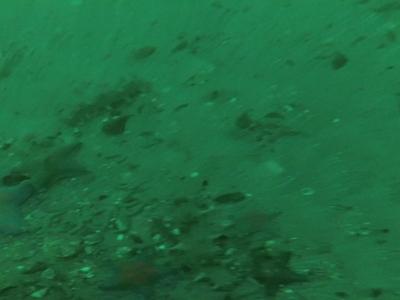} 
\\ 
\InfoCell{image} &   \InfoCell{preferred} & &
\InfoCell{blue cast} &   \InfoCell{green cast}
\\\\\\\\
\WholeCell{(c) {UIEB}~\cite{WaterNET_TIP2019}} & & \WholeCell{(d)  RUIE UCCS~\cite{liu19_RUIE}} 
\\
\vspace{3pt}
\end{tabular}
\\
\begin{tabular}{ccccc}
\includegraphics[width=0.24\columnwidth]{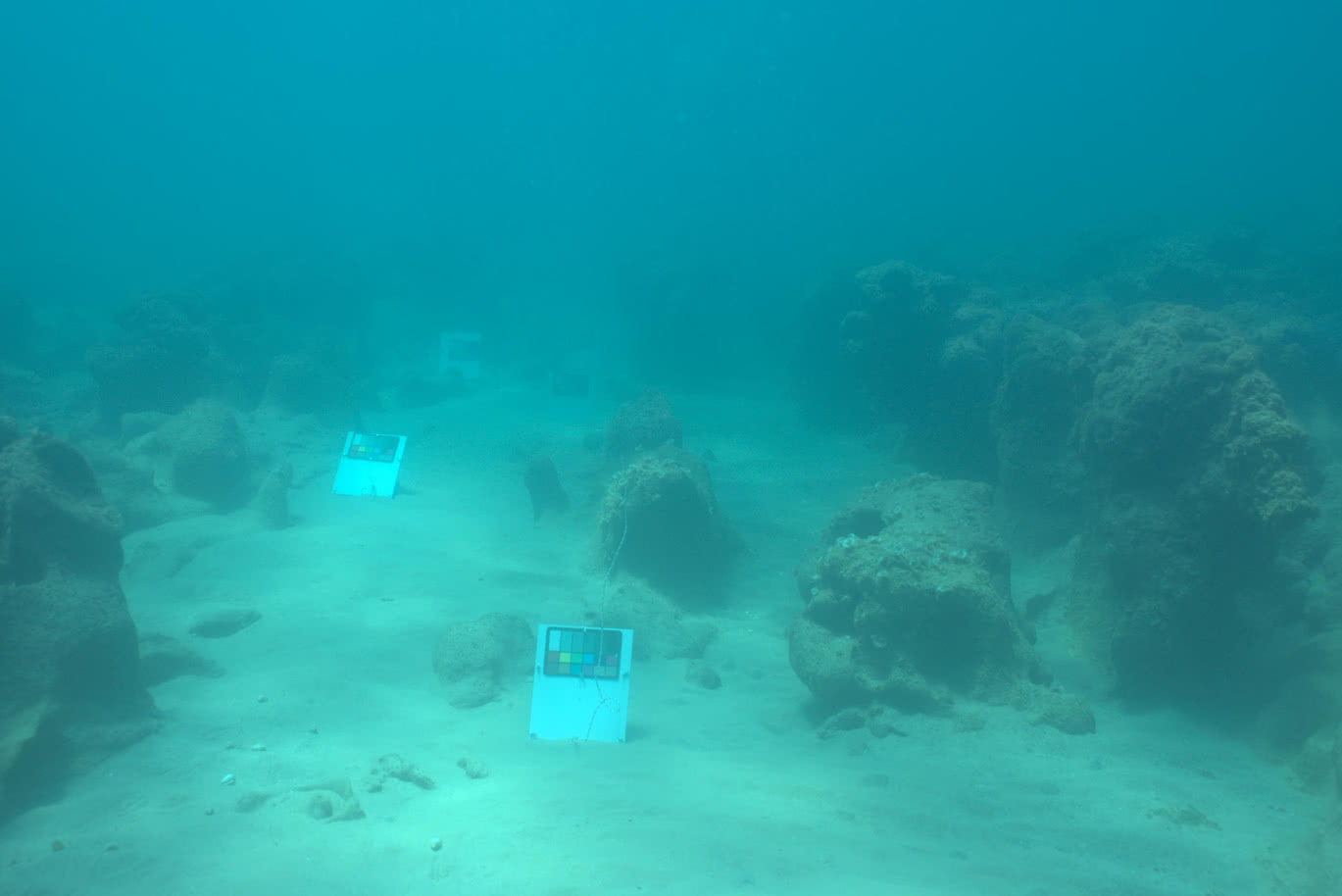}
& \includegraphics[width=0.24\columnwidth]{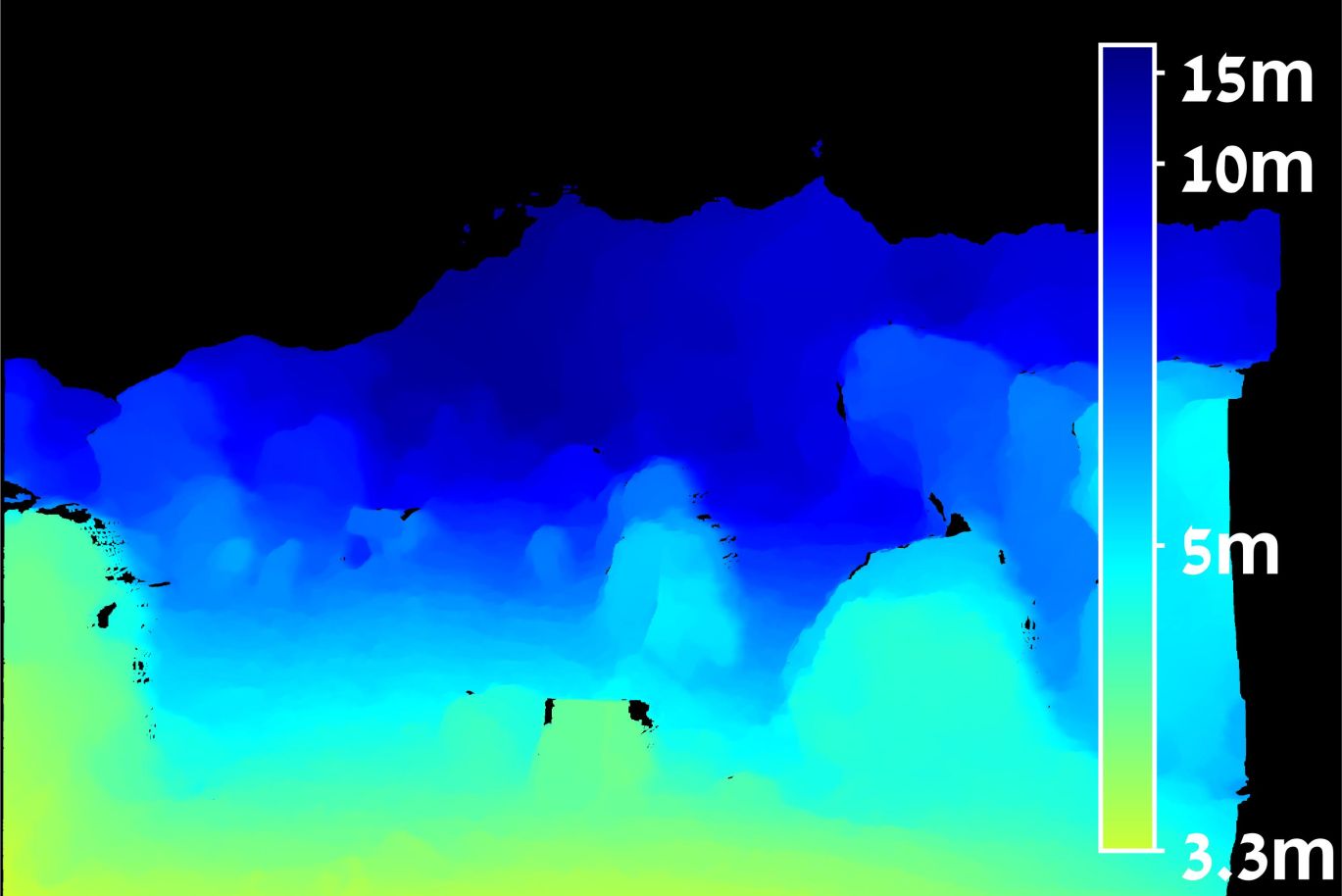} & &
\includegraphics[width=0.24\columnwidth]{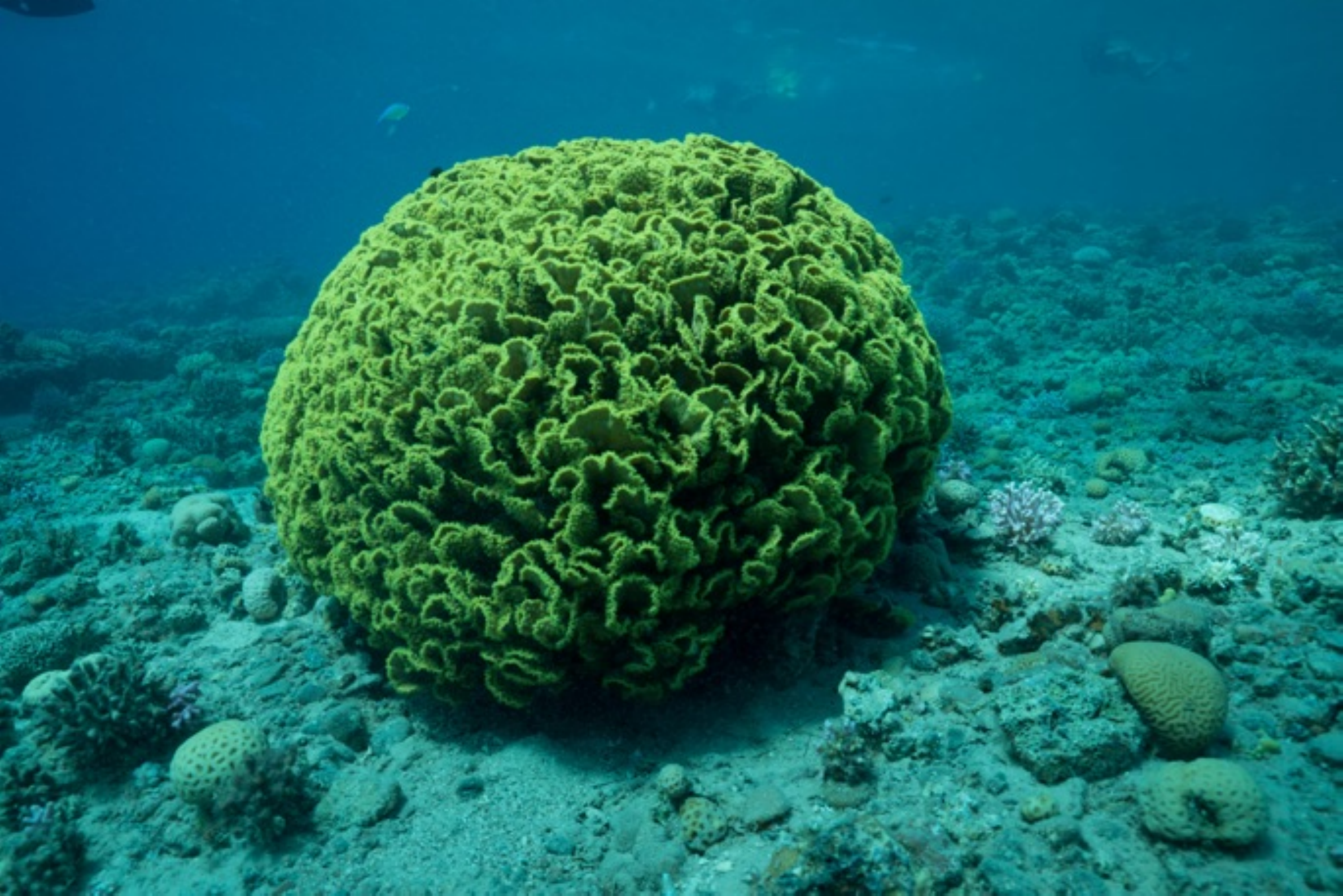} & 
\includegraphics[width=0.24\columnwidth]{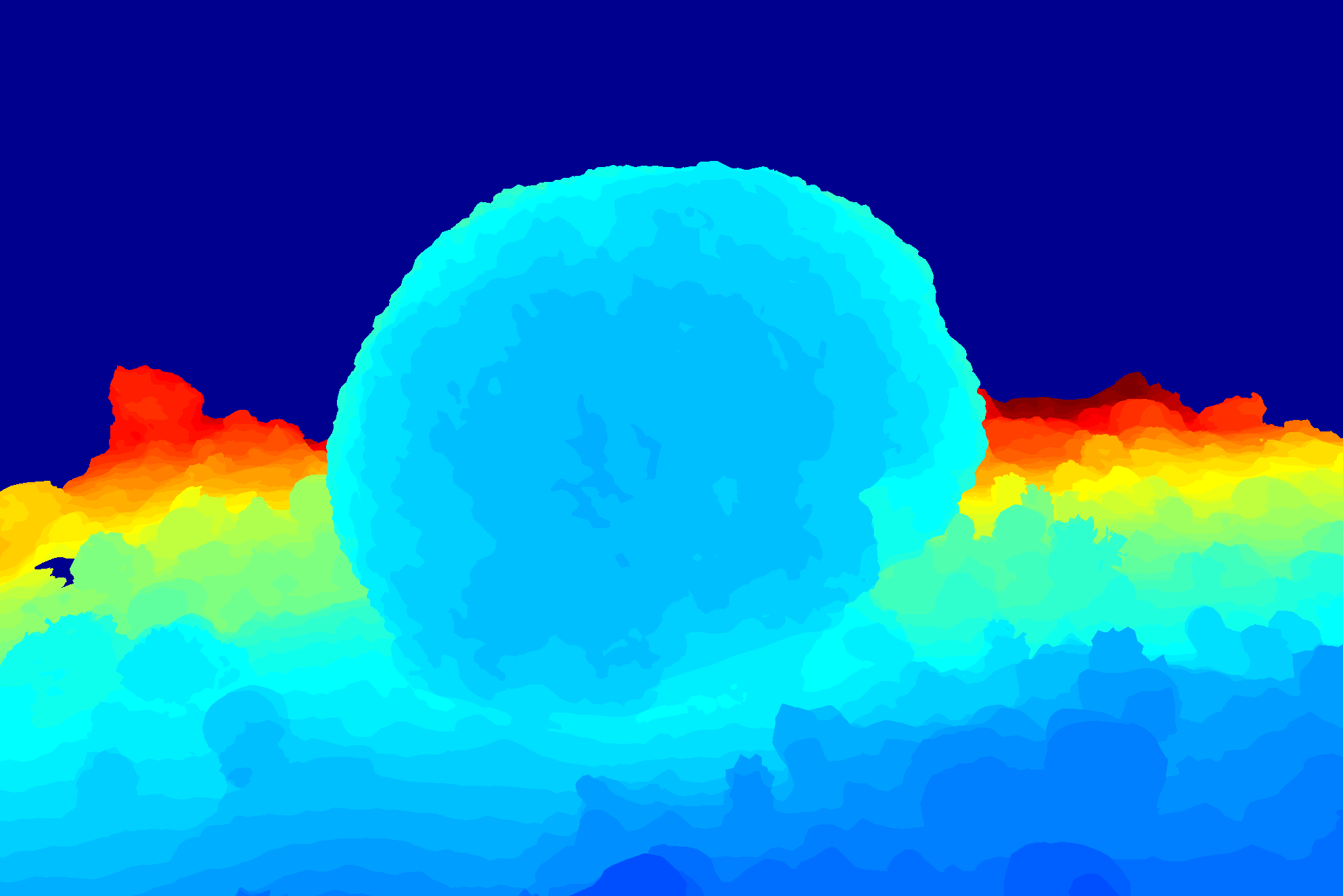}
\\
\InfoCell{image} & \InfoCell{range} & &\InfoCell{image} & \InfoCell{range}\\\\\\\\
\WholeCell{(e) {SQUID}~\cite{BermanArvix_UWHL_dataset}} && \WholeCell{(f) {Sea-thru}~\cite{seathru2019CVPR}}  \\\\
\vspace{3pt}
\end{tabular}
\\
\begin{tabular}{ccccc}
\includegraphics[width=0.203\columnwidth]{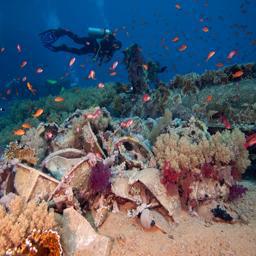}&\includegraphics[width=0.203\columnwidth]{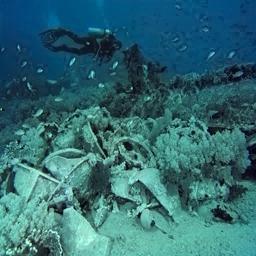} 
& &  \includegraphics[width=0.275\columnwidth]{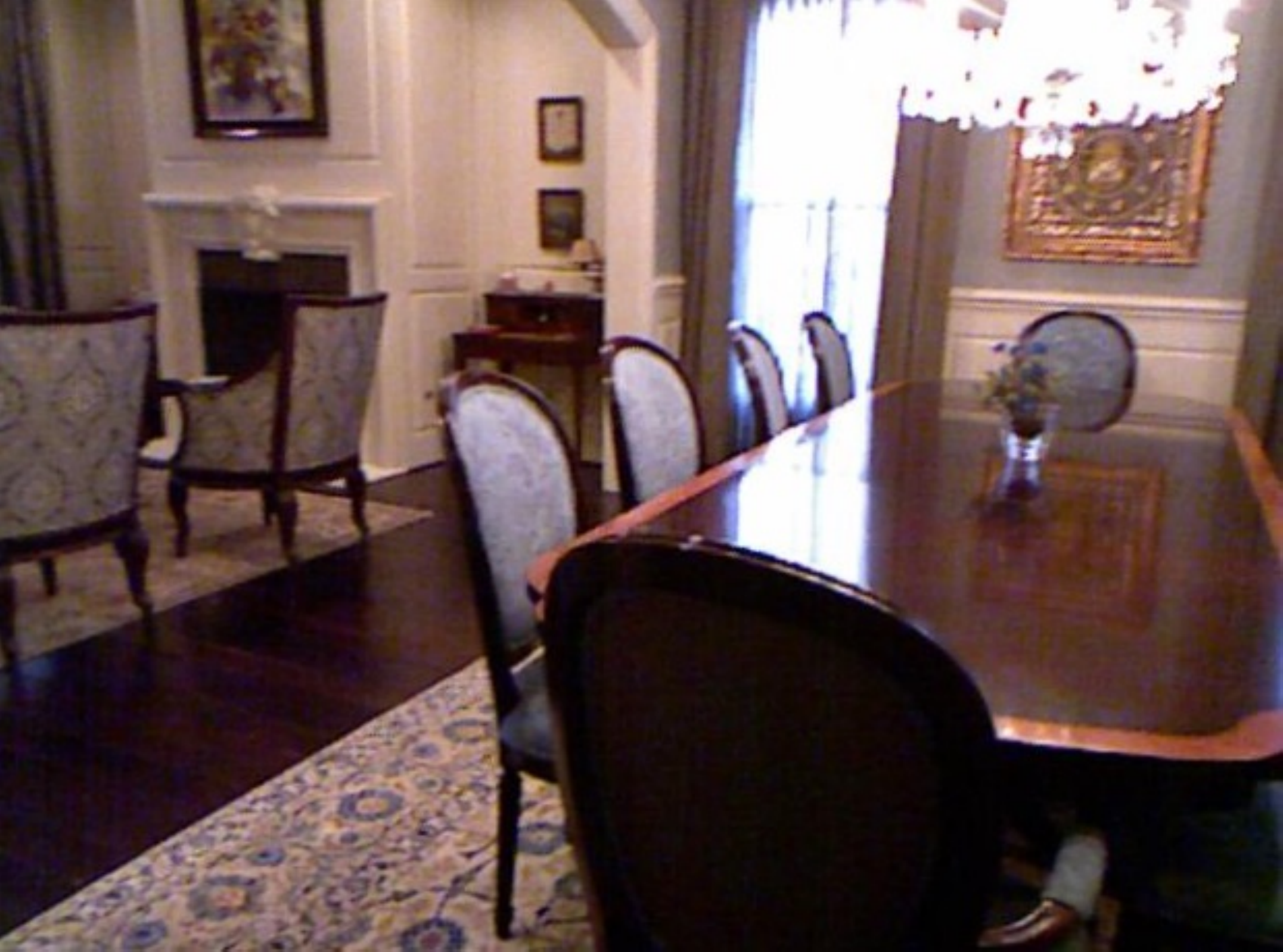}
&\includegraphics[width=0.275\columnwidth]{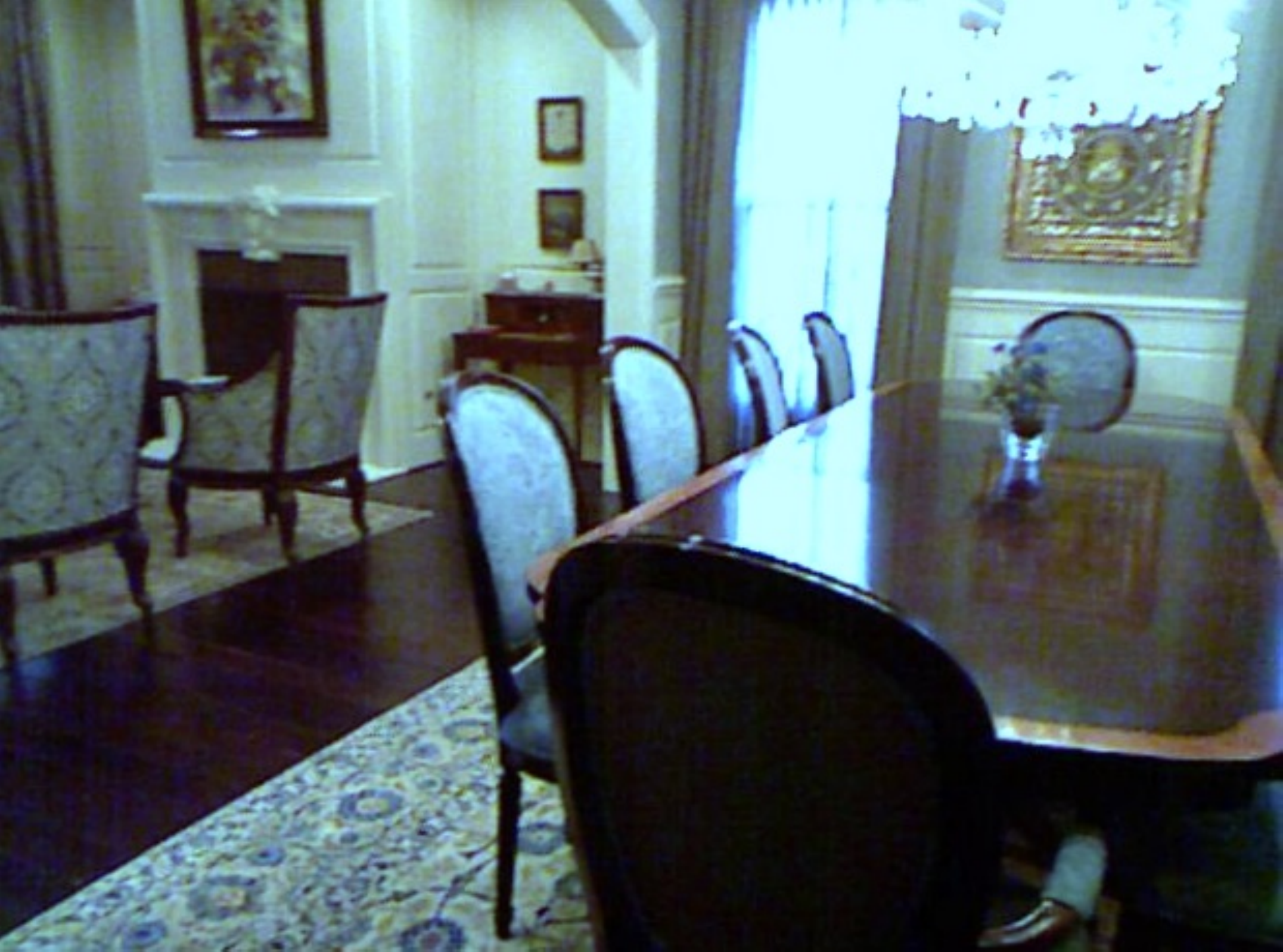} \\
\InfoCell{reference} & \InfoCell{degraded} & & \InfoCell{reference} & \InfoCell{degraded}\\ \vspace{3pt}\\
\WholeCell{(g) {EUVP pair}~\cite{Islam2019FastUI4ImprovedPerception}} &&\WholeCell{(h) {PR synthesised}~\cite{Anwar2018DeepUI_UWCNN}}\\\\
\vspace{3pt}
\end{tabular}
\\
\begin{tabular}{ccccc}
\includegraphics[width=0.24\columnwidth]{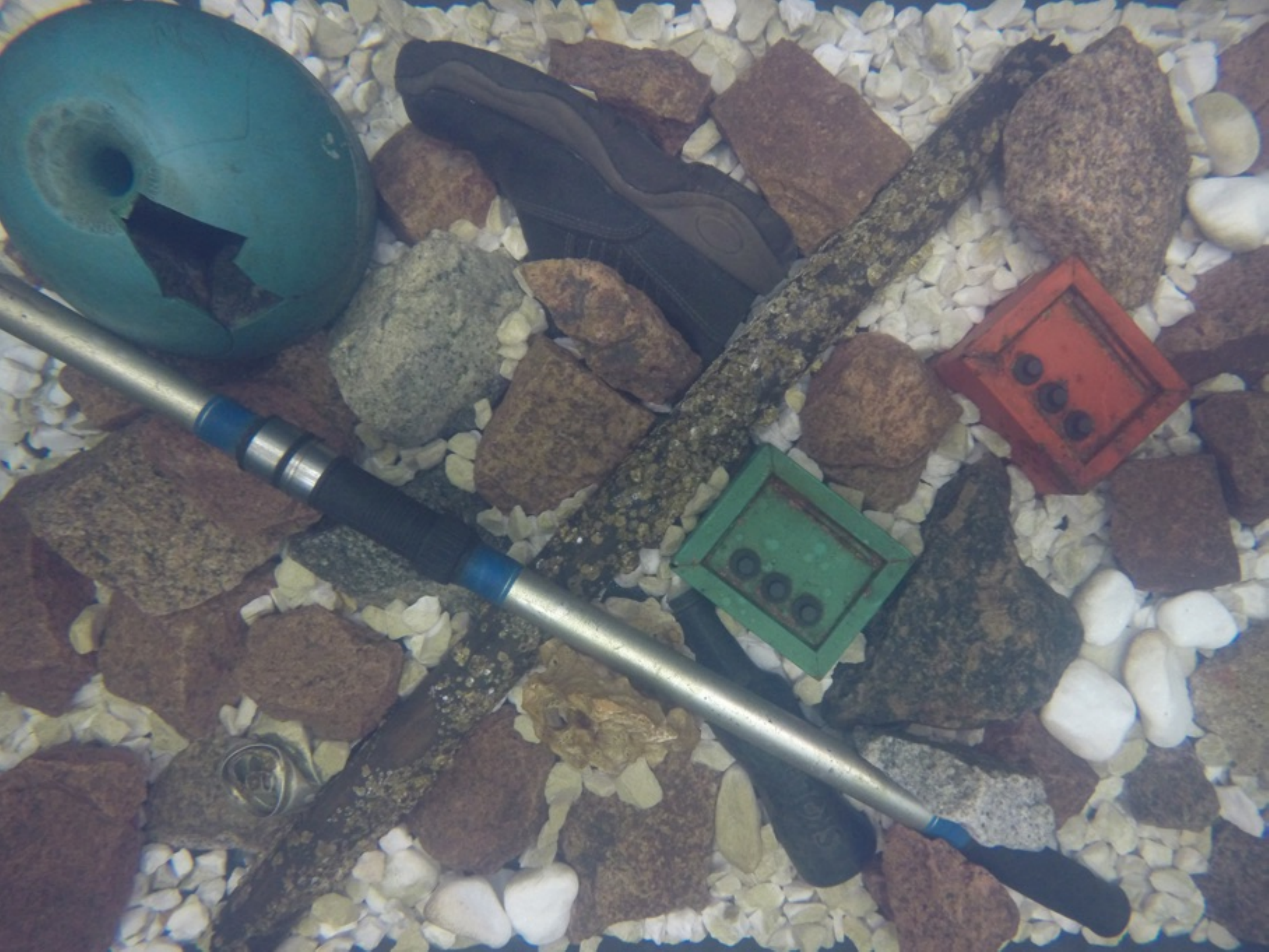}
& \includegraphics[width=0.24\columnwidth]{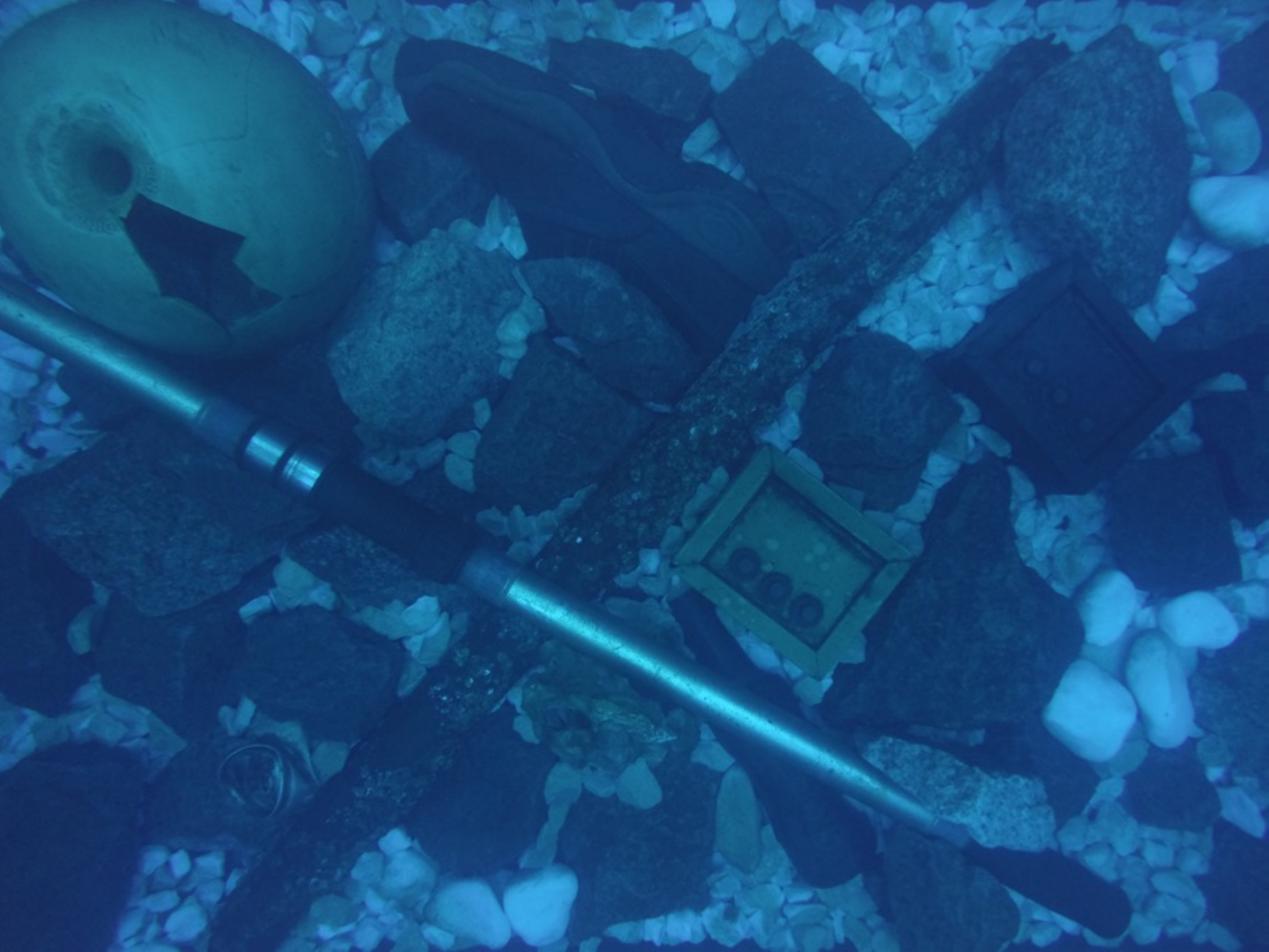} 
& & 
\includegraphics[width=0.24\columnwidth]{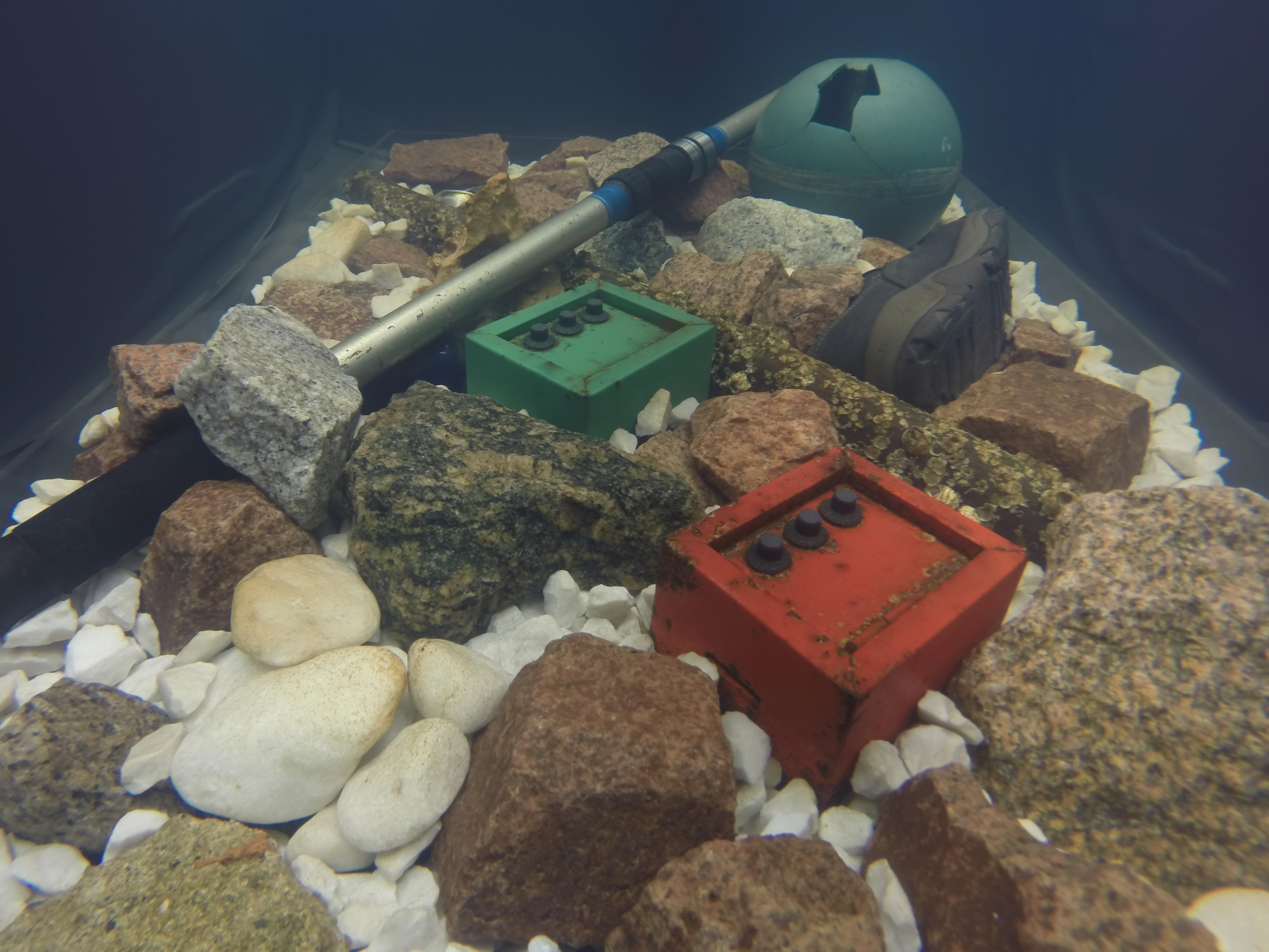} &
\includegraphics[width=0.24\columnwidth]{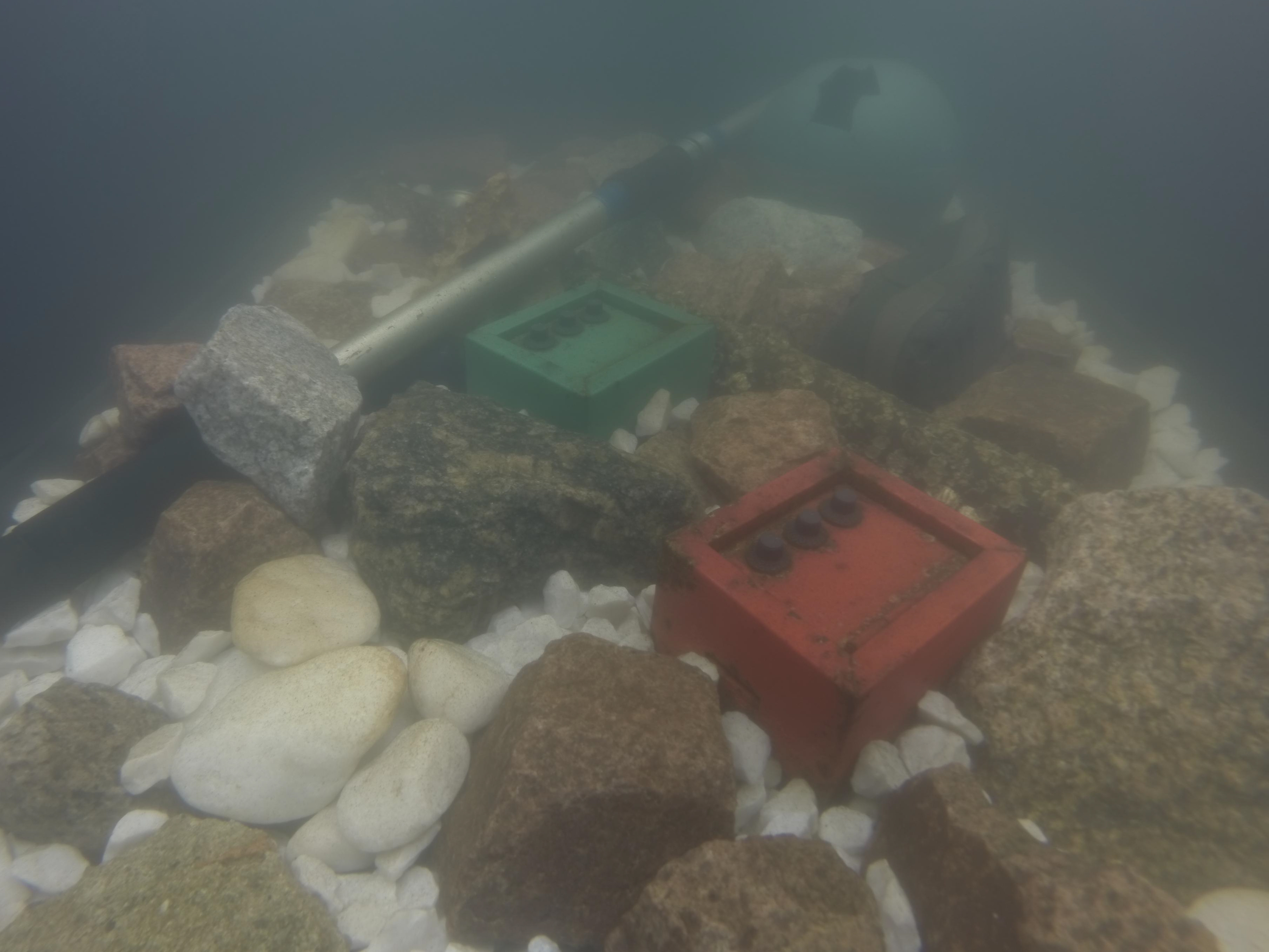} 
\\
\InfoCell{reference} & \InfoCell{degraded} &  & \InfoCell{reference} & \InfoCell{degraded}\\
\vspace{3pt}
\\
\WholeCell{(i) {TURBID DeepBlue}~\cite{Duarte2016}} & &\WholeCell{(j) {Chlorophyll}~\cite{Duarte2016}} \\
\end{tabular}
\end{tabular}
\caption{Sample images from underwater datasets. The first three rows  show images from {\em natural} datasets and their annotation on quality (RUIE UIQA~\cite{liu19_RUIE} and EUVP unpair~\cite{Islam2019FastUI4ImprovedPerception}) and colour cast (RUIE UCCS~\cite{liu19_RUIE}), correspondences between degraded and reference images (UIEB~\cite{WaterNET_TIP2019}), or with the scene's range provided (SQUID~\cite{BermanArvix_UWHL_dataset} and Sea-thru~\cite{seathru2019CVPR}). The fourth row shows images from  {\em digitally degraded} datasets, where the degradations are synthesised by a neural network (EUVP pair~\cite{Islam2019FastUI4ImprovedPerception}) or a physics-based model (PR synthesised for Jerlov type I water~\cite{Anwar2018DeepUI_UWCNN}). The fifth row shows images from~{\em chemically degraded} datasets, where the degradation is created by adding chemical reagents to water~(TURBID DeepBlue and Chlorophyll~\cite{Duarte2016}).}
\label{fig:: natural_dataset_images}
\end{figure}

As for digitally degraded datasets, the PR 2019 Synthetic Underwater Image Dataset~\cite{Anwar2018DeepUI_UWCNN} contains 1,449 images synthesised from the~\mbox{NYU-D} indoor dataset~\cite{Silberman:NYUD:ECCV12} with Eq.~\ref{eq:model}~(Fig.~\ref{fig:: natural_dataset_images}(h)). 
The dataset is synthesised for 10 Jerlov water types and different degradation levels. 
Fabbri {et al.}~\cite{UGAN_Fabbri_2018} presented a dataset of underwater images for training a CycleGAN~\cite{CycleGAN_2017_Zhu} using degraded images from ImageNet and manual annotations for distortion. 6,143 images with distortion and 1,817 images without distortion were used for training. CycleGAN then generated the image pair with no distortion for the 6,143 images. 
The EUVP pair collection~\cite{Islam2019FastUI4ImprovedPerception} uses the same approach and training dataset as Fabbri et al.~\cite{UGAN_Fabbri_2018}~(Fig.~\ref{fig:: natural_dataset_images}(g)). A total of 24,840 reference underwater images are generated by the trained CycleGAN. 
The former approach does not model the range-dependent scattering, whereas the latter lacks proof that all the characteristics of water distortion are accurately synthesised in datasets. 

A notable chemically degraded dataset is TURBID~\cite{Duarte2016} that captures artificial objects (such as plastic corals and metal boxes) placed on a sand floor in a water tank. To recreate the degradations occurring in natural water, milk was added for turbidity, methylene blue for a blue water appearance (DeepBlue subset, Fig.~\ref{fig:: natural_dataset_images}(i)) and chlorophyll for a green water appearance (Chlorophyll subset, Fig.~\ref{fig:: natural_dataset_images}(j)).  
TURBID provides reference images captured in a controlled water environment. However, the added reagents do not fully replicate the contributing factors of natural water environments, and hence may not be representative of underwater images captured in natural oceanic or costal waters.

\section{Evaluation}\label{sec: evaluation}

In this section, we discuss evaluation approaches to validate a task with objective measures~(Sec.~\ref{sec:targetEval}), as well as  {subjective} tests~(Sec.~\ref{sec:subjEval}) and automatic quality assessment measures, which aim to produce results that align with human judgement (Sec.~\ref{sec:imgQualAssessment}).

\subsection{Task evaluation}\label{sec:targetEval}

Task evaluation consists of applying computer vision algorithms to quantify the improvement of underwater image filtering results or in assessing these results for a specific reference object or image region.

Several authors used the performance of computer vision {\em algorithms} to assess the effectiveness of filtering methods. Examples include feature point  matching, employed in~\cite{Ancuti2018, Gao2019_Adaptive_Retinex}; image segmentation,  employed in~\cite{Ancuti2018}; object detection, employed in~\cite{Islam2019FastUI4ImprovedPerception, UGAN_Fabbri_2018};  and pose estimation, employed in~\cite{Islam2019FastUI4ImprovedPerception}. These evaluations are generally presented and discussed for a few images only and, in absence of a ground truth, no quantitative analysis can be performed. 

The accuracy of {\em colour reproduction} can be evaluated, using a standard colour chart, by measuring the discrepancy between the colour of a patch in a filtered image and the colour of the same patch as its appears in a scene captured under a white illuminant. The colour difference can be measured as angular error~\cite{BMVCAngularError} (in~\cite{BMVC_HL_2017, Gao2019_Adaptive_Retinex}), Euclidean distance (in~\cite{Li2018WaterGANUG}), or CIE2000~\cite{sharma2005ciede2000} (in~\cite{Ancuti2018, Gao2019_Adaptive_Retinex}). This evaluation depends on the availability of images containing a colour chart and is limited to a specific water environment. 

To assess the extent of water colour preservation, the  change in {\em water colour} between the degraded and filtered image can measured, considering that the colour of the water mass should not be substantially modified during filtering. This evaluation can be used to assess, for images with visible pure water regions, the quality of the restoration along the {range} (i.e.~the distance between an object and the camera). The change can be measured as the mean square error between the RGB intensities in pixels of the annotated water mass regions~\cite{BglightEstimationICIP2018_Submitted}.

\subsection{Subjective tests}\label{sec:subjEval}

Subjective tests for underwater images are challenging to design and, unlike standard quality assessment~\cite{ITURec500-14,NetflixVideo}, no established protocol exists. However, a set of specific instructions may help participants focus on what to evaluate. 

Table~\ref{Table: evaluation_measures_subj} summarises the existing subjective tests for underwater image filtering results. 
\begin{table*}[t!]
\caption{Subjective tests to evaluate restoration (R) and enhancement (E) methods, and for dataset annotation (DS). For each test we list the numbers of images, the number of methods under evaluation and the number of participants (Part.).  Text in {\em italics} reports, when available,  the exact question the participants were asked. In ranking and selection, the participant evaluates multiple images for each response, whereas in scoring, the participant evaluates one image for each response. } 
\scriptsize
\begin{minipage}[b]{1\linewidth}
\centering
\setlength\tabcolsep{1.5pt}
\begin{tabular}{cllccccrrrllcc}
\Xhline{3\arrayrulewidth}
&\multicolumn{2}{c}{\textbf{Reference}} & \multicolumn{2}{c}{\textbf{Rank/select}}  &\multicolumn{1}{c}{\textbf{Score}}  & {\textbf{Against}} & 
\multicolumn{3}{c}{\textbf{Number of}} & 
&\multicolumn{1}{c}{\textbf{Task/ Question}}\\
\textbf{} &\multicolumn{2}{c}{\textbf{}}  & Double & Multi & & {\textbf{degraded}} & \multicolumn{1}{c}{{Images}} &{{Methods}} &\multicolumn{1}{r}{Part.} &\multicolumn{1}{c}{{}}&   \\
\hline
\multirow{6}{*}{R} 
&\cite{Emberton2015} & Emberton {et al.}&\checkmark & &  & & 10 & 4 & 10 &   & select the preferred image out of 4\\
&\cite{Emberton2018} & Emberton {et al.}   &\checkmark &   &   & & 100 & 4 & 12  &  &  select the preferred image out of 4\\
&~\cite{Li_2020_Cast-GAN} & Li and Cavallaro&\checkmark &  &  &\checkmark & 12 & 3 & 21 & & '{\em select the image that looks more like it is taken under a white illuminant}' \\
&~\cite{Anwar2018DeepUI_UWCNN} & Li {et al.} &  &\checkmark &  & & 20 & 5 & 20 &  &  rank 5 images based on their perceived visual quality  \\
&~\cite{BglightEstimationICIP2018_Submitted} & Li and Cavallaro   &  &\checkmark & &\checkmark & 26 & 5 & 23 &  & '{\em select the most attractive image}' out of 5 images\\
&\cite{Hou2020_DCPguided_TV} & Hou {et al.}&  &  & \checkmark & & 50 & 7 & 20  &  &  score the 7 images from worst to best (1-7) based on preference \\
\hline
\multirow{3}{*}{E}
&\cite{Protasiuk2019_localcolourmapping} & Protasiuk {et al.}& &\checkmark &  & & 4 & 10 & 10 &  &  rank 4 images from least to most realistic (1-4)\\
&~\cite{Islam2019FastUI4ImprovedPerception} & Islam {et al.}   &  &\checkmark &   &\checkmark & 4 & 9 & 9 &  & out of 9 images, rank the top 3 based on quality\\ 
&~\cite{Zhang2017_multisclae_retinex} & Zhang {et al.} &  & &\checkmark &  & 4 & 1 & 30 &  &  '{\em  whether colour patches on a colour chart can be distinguished}'\\

\hline\hline
\multirow{2}{*}{DS}
&~\cite{WaterNET_TIP2019} & UIEB &\checkmark & &  &\checkmark & 890 & 12 & 50 &   &  '{\em which one is better}' \\
&~\cite{Islam2019FastUI4ImprovedPerception} & EUVP   &  & &\checkmark  &  & 6,665 & 1 & 6 &  &  '{\em whether the background and foreground are identifiable}'\\ 
\Xhline{3\arrayrulewidth}
\end{tabular}
\label{Table: evaluation_measures_subj}
\end{minipage}
\end{table*}
Comparative experiments aim at the selection of the {\em preferred} images through double-stimulus tests~\cite{Emberton2015, Emberton2018}, multi-selection tests~\cite{BglightEstimationICIP2018_Submitted}, or ranking of multiple filtered images~\cite{Hou2020_DCPguided_TV, Anwar2018DeepUI_UWCNN}. 
Other subjective tests~\cite{Protasiuk2019_localcolourmapping, Zhang2017_multisclae_retinex,Islam2019FastUI4ImprovedPerception} asked participants to score images based on '{\em how realistic they are}'~\cite{Protasiuk2019_localcolourmapping} or to indicate whether the {\em colour patches on a colour chart can be distinguished}~\cite{Zhang2017_multisclae_retinex}. These experiments evaluated a few images only  (4 in~\cite{Zhang2017_multisclae_retinex} and 40 in~\cite{Protasiuk2019_localcolourmapping}). To validate the extent of colour cast removal, participants can be requested to select, in a double-stimulus test, the image that '{\em looks more like it is taken under a white illuminant}'~\cite{Li_2020_Cast-GAN}. 

Fig.~\ref{fig:: experiment_natural_images} compares the filtering results of eight representative state-of-the-art methods, of which  five are for restoration  (Automatic Red Channel (ARC)~\cite{ARC}, Depth-compensated Background Light Restoration (DBL)~\cite{BglightEstimationICIP2018_Submitted}, Wavelength Compensation Image Dehazing (WCID)~\cite{WCID}, Underwater Haze-line (UWHL)~\cite{BMVC_HL_2017} and UWCNN~\cite{Anwar2018DeepUI_UWCNN}) and three for enhancement (Colour Balance and Fusion by Ancuti et al. (Fusion)~\cite{Ancuti2018}, FUnIE~\cite{Islam2019FastUI4ImprovedPerception} and WaterNET~\cite{WaterNET_TIP2019}). The selected images are typically used in publications and cover the two main Jerlov oceanic and coastal water categories, with scenes with  non-uniform water colour and objects at a varying range from the camera.
Among the restoration methods, WCID and UWHL are the only two that aim to compensate for the degradation through the depth under the water surface. WCID and DBL address the non-uniform water colour. ARC and DBL are designed for oceanic waters. UWHL is designed for all water types by iteratively restoring the images using the properties of the 10 Jerlov water types and selects the restored image that best fits the gray-world assumption. UWCNN also presents 10 networks trained for the 10 Jerlov water types, and we show the results for oceanic (type I) and coastal (type 3) water types.  

ARC slightly darkens the images, owning to the use of weighted background light to replace the colour-channel dependent transmission map. Methods designed for oceanic water (ARC and DBL) have a limited performance on images taken in coastal waters.  Although DBL fails to compensate for the degradation or remove any colour cast in green water, it compensates for the colour degradation along the range. 
Moreover, DBL successfully addresses the non-uniform water colour (Fig.~\ref{fig:: experiment_natural_images} row 2). 
UWHL can compensate for the colour degradations in oceanic water, although it may produce over-saturated images~(Fig.~\ref{fig:: experiment_natural_images} row~1). UWHL can also remove the colour cast in heavily green casted images~\mbox{(Fig.~\ref{fig:: experiment_natural_images} rows~5 to~7)}. However, the global white balancing used to remove the cast distorts the water colour and introduces unnatural reds in the scene. 
UWCNN, which was trained for oceanic type I waters, introduces a pink cast to oceanic images. This might be caused by the synthesised training dataset~(PR synthesised dataset~\cite{Anwar2018DeepUI_UWCNN}), which does not contain objects naturally present in water: the network learned the colour distribution of indoor objects and overcompensates for the loss in red. While this network fails - as expected - to compensate for the colour casts in coastal waters, UWCNN trained for coastal type 3 fails to restore any images taken in either coastal or oceanic water. This confirms that, because of the lack of appropriate training datasets, neural networks, which are to date trained on synthetic datasets, are still underperforming in the underwater image restoration task. 

The above aims to provide an unbiased analysis of the methods' performances. However, readers might still ignore the visible artifacts and consider some of the said failure images to be of better qualities. We note that subjective tests are time-consuming to conduct and can be potentially biased by the pool of participants, if the pool is not selected and trained carefully. To overcome this problem, automatic assessment measures have been developed.
\subsection{Automatic  assessment}\label{sec:imgQualAssessment}
 
\begin{table}[t!]
\caption{Underwater-specific image quality assessment measures aim to automatically evaluate the image quality in agreement with human judgements. Based on whether a reference image is required, the measures are categorised (Cat.) as full-reference (FR) or no-reference (NR). The measure can be applied (App.) to score an image or to rank the relative quality between two images. Weights that combine (Comb.) the attributes can be  derived from the subjective test results using Support Vector Machines (SVM) or regression (reg.). Note that $Q_u$ was not validated with subjective test. We list the number of images (Img.), the number of participants (Part.), and the task in the test.} 
\scriptsize
\centering
\setlength\tabcolsep{0.5pt}
\begin{tabular}{llllcclrcccccccccccccccccc}
\Xhline{3\arrayrulewidth} 
\multicolumn{1}{c}{\textbf{Cat.}} & 
\multicolumn{2}{c}{\textbf{Measure}} &\multicolumn{1}{c}{\textbf{App.}} &\multicolumn{2}{c}{\textbf{attributes}} & {\textbf{Comb.}} &\multicolumn{3}{c}{\textbf{Subjective Test}}\\
\multicolumn{2}{c}{\textbf{}} &\multicolumn{2}{c}{} &\multicolumn{1}{c}{Colour}&\multicolumn{1}{c}{{Visibility}} & & {{Img.}}  & {{Part.}} & {{Task}}\\
\hline

\multirow{2}{*}{FR} 
&~\label{row:quality_Lu2016ICIP}\cite{Lu2016_descattering_quality_ICIP} & $Q_u$  & score & \checkmark &\checkmark &  equal &\multicolumn{3}{c}{\cellcolor{black!40}not applicable}\\
&~\label{row:PKU-EAQA}\cite{PKU}  & PKU-EAQA  &  rank &  \checkmark &\checkmark & SVM & 400  & 30 & compare\\
\hline
\multirow{3}{*}{NR} 
&~\label{row:UCIQE}\cite{UCIQE} & UCIQE & score &  \checkmark & & reg. & 44 & 12 & score\\
&~\label{row:CFF}\cite{WANG2018_NR_CFF} & CCF & score&  \checkmark
&\checkmark & reg. & 87 &  {{20}} & {{score}}\\
&~\label{row:UIQM}\cite{UIQM_Panetta2016} & UIQM & score &  \checkmark &\checkmark & reg. & 126  & 10 & score\\
\Xhline{3\arrayrulewidth}
\end{tabular}
\label{Table: evaluation_measures_obj}
\end{table}

We discuss {generic} {Image Quality Assessment} (IQA) measures  used in underwater literature and IQA measures specifically designed for underwater images (Table~\ref{Table: evaluation_measures_obj}).
IQA  measures can be categorised as no-reference~(NR) or full-reference~(FR). NR measures require only the image under consideration, whereas FR measures require a reference image to measure the relative degradations.

\begin{figure*}[t!]
\setlength\tabcolsep{0.5pt}
\centering
\begin{tabular}{ccccccccc}
\includegraphics[height=0.047\textheight]{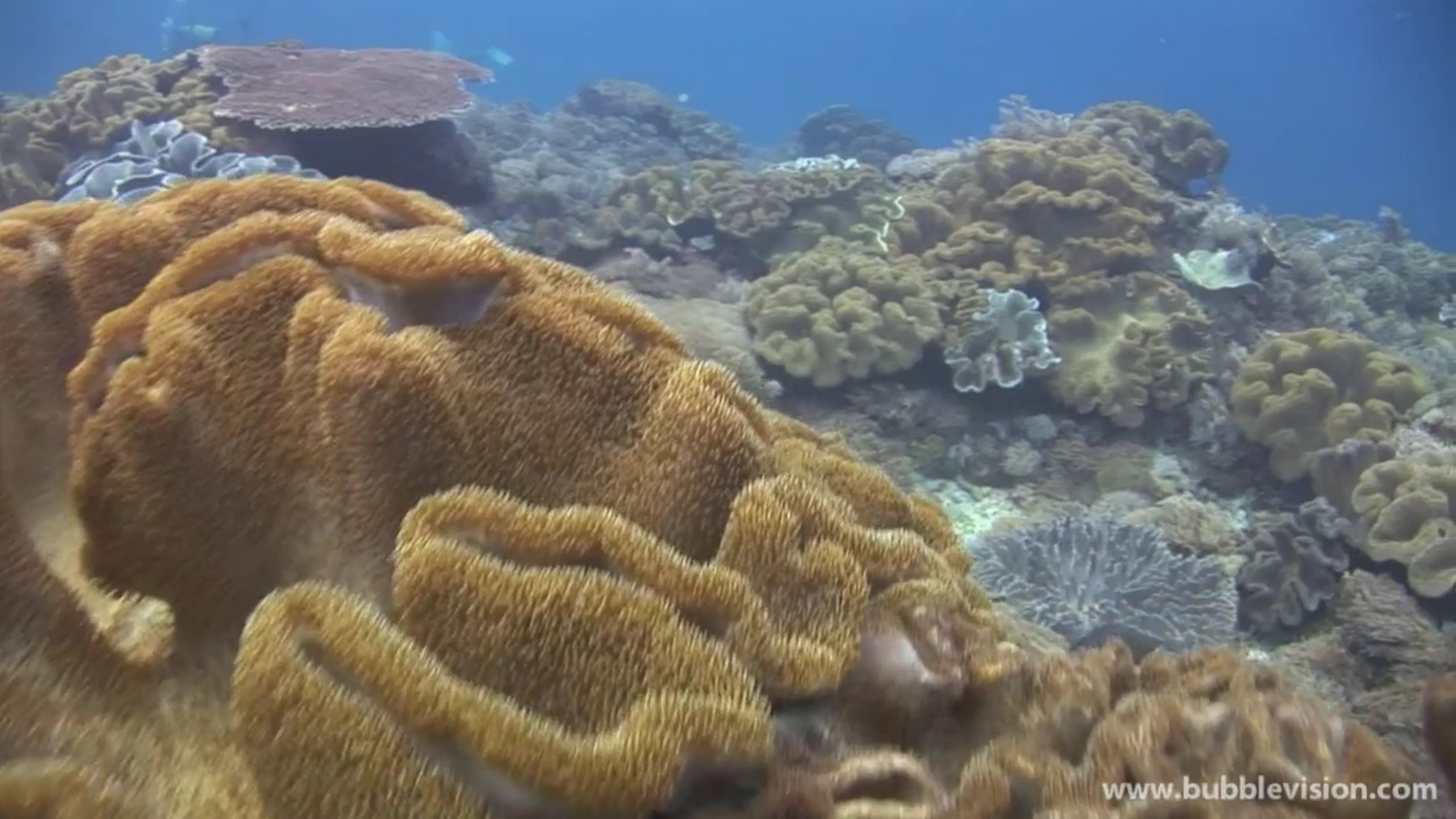}&\includegraphics[height=0.047\textheight]{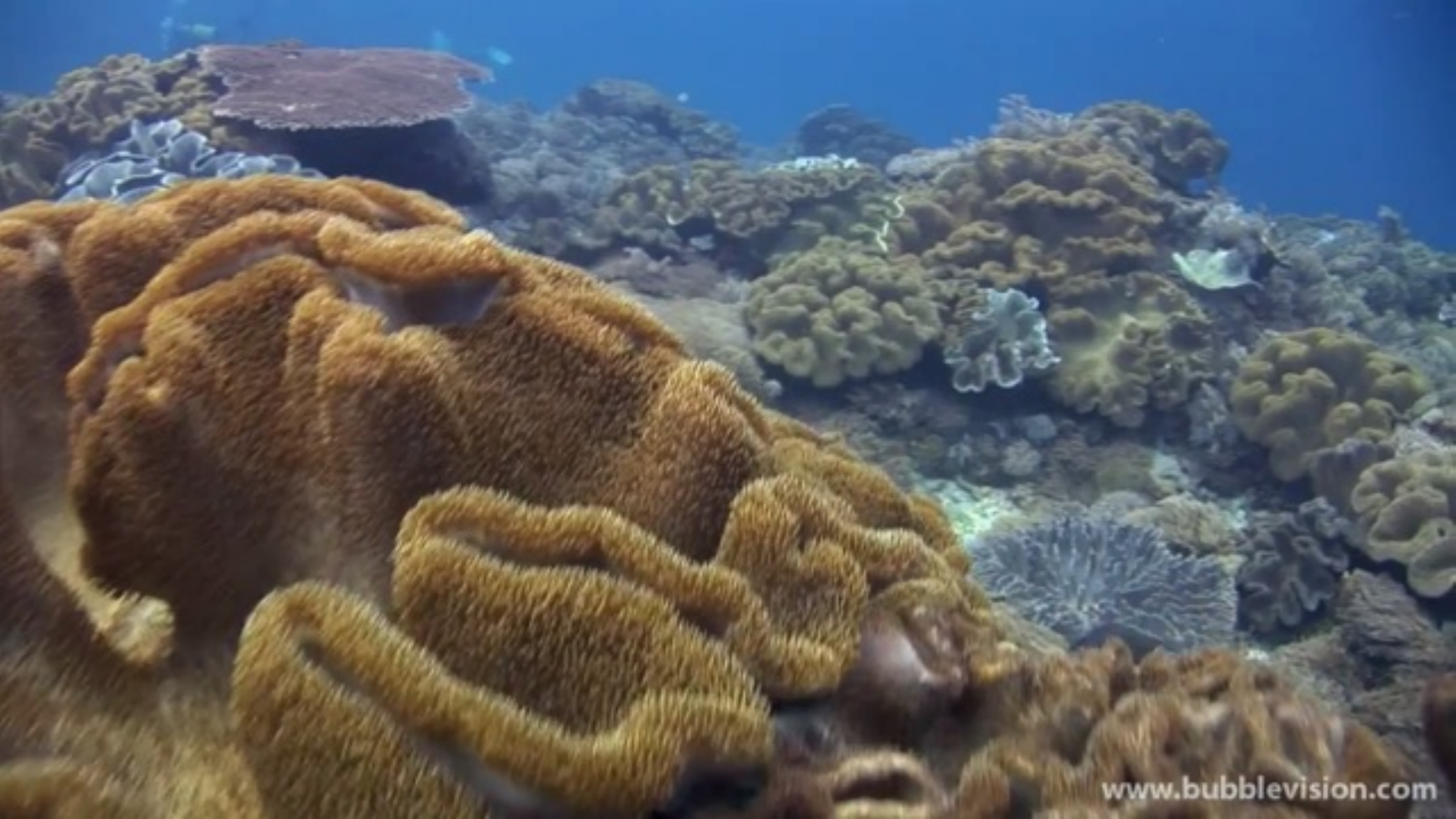}&\includegraphics[height=0.047\textheight]{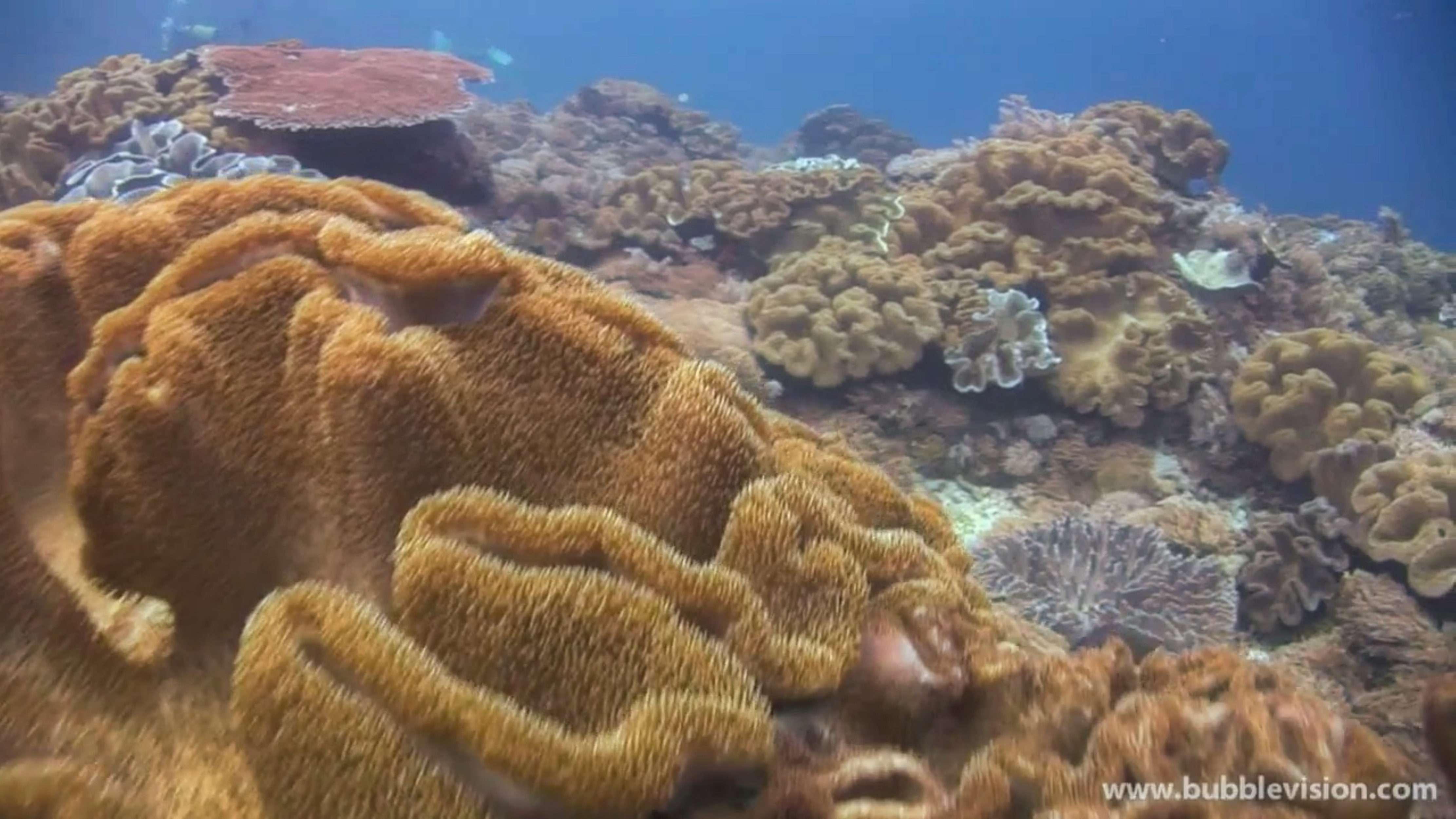}&\includegraphics[height=0.047\textheight]{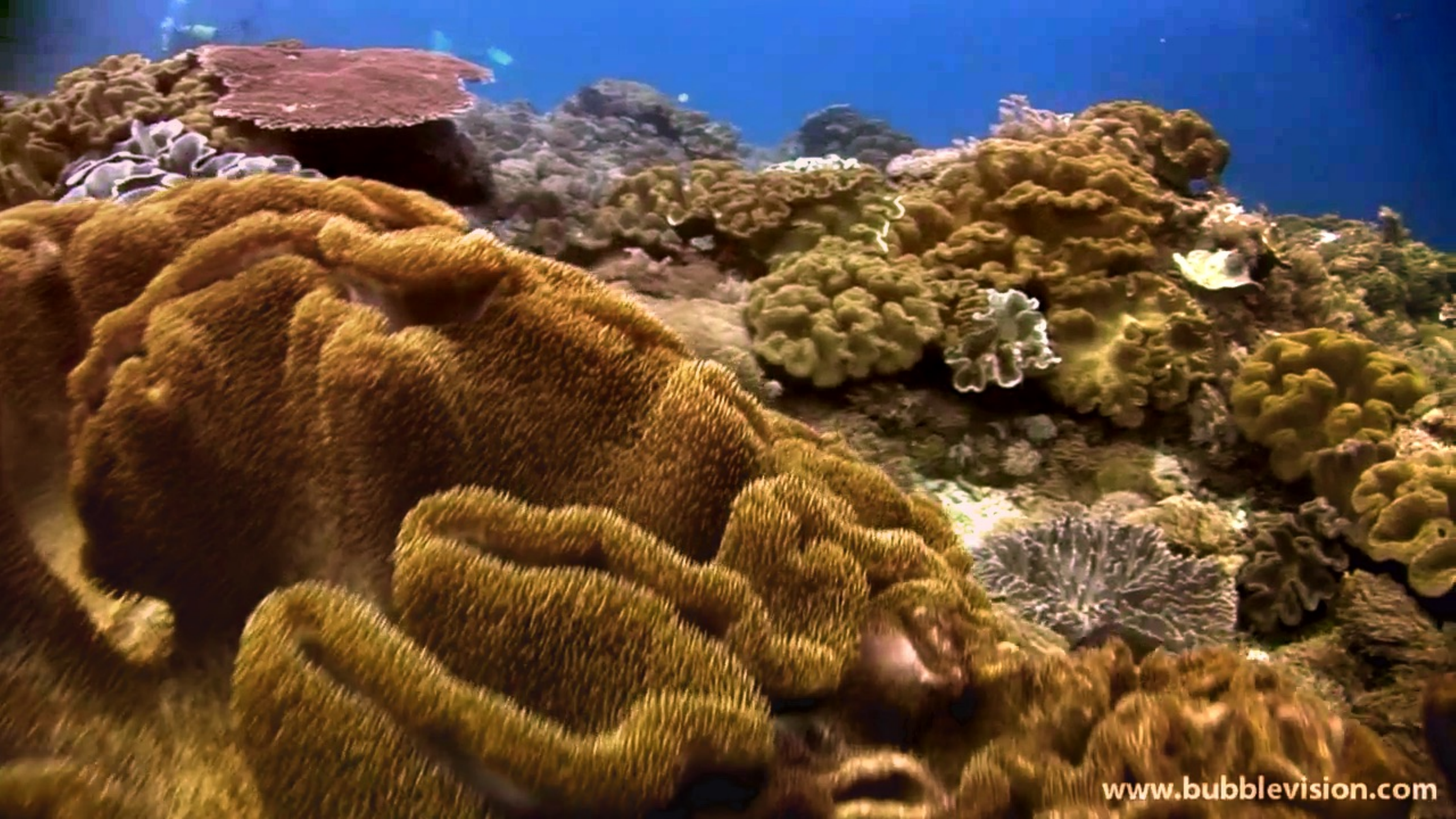}&\includegraphics[height=0.047\textheight]{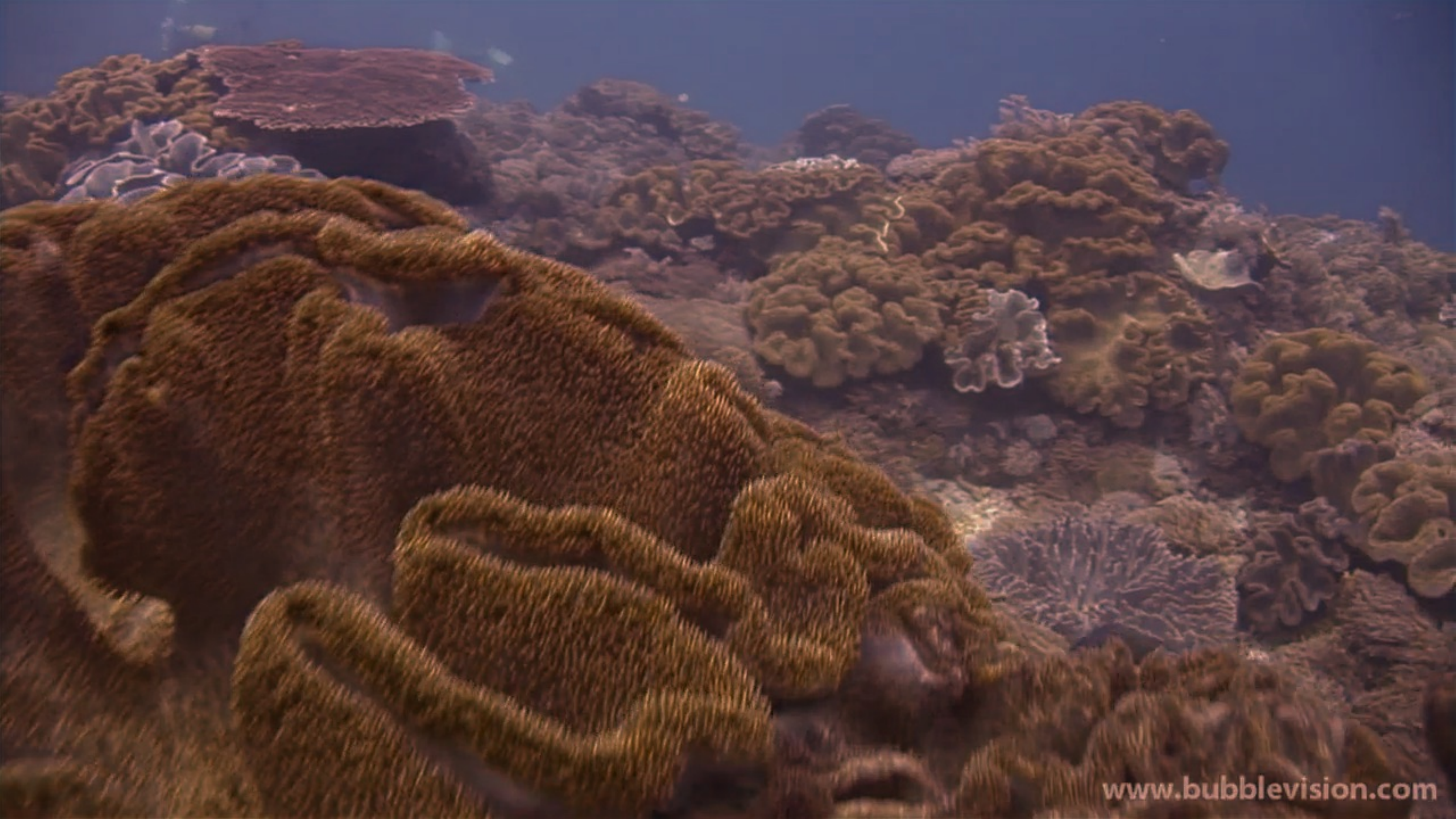}&\includegraphics[height=0.047\textheight]{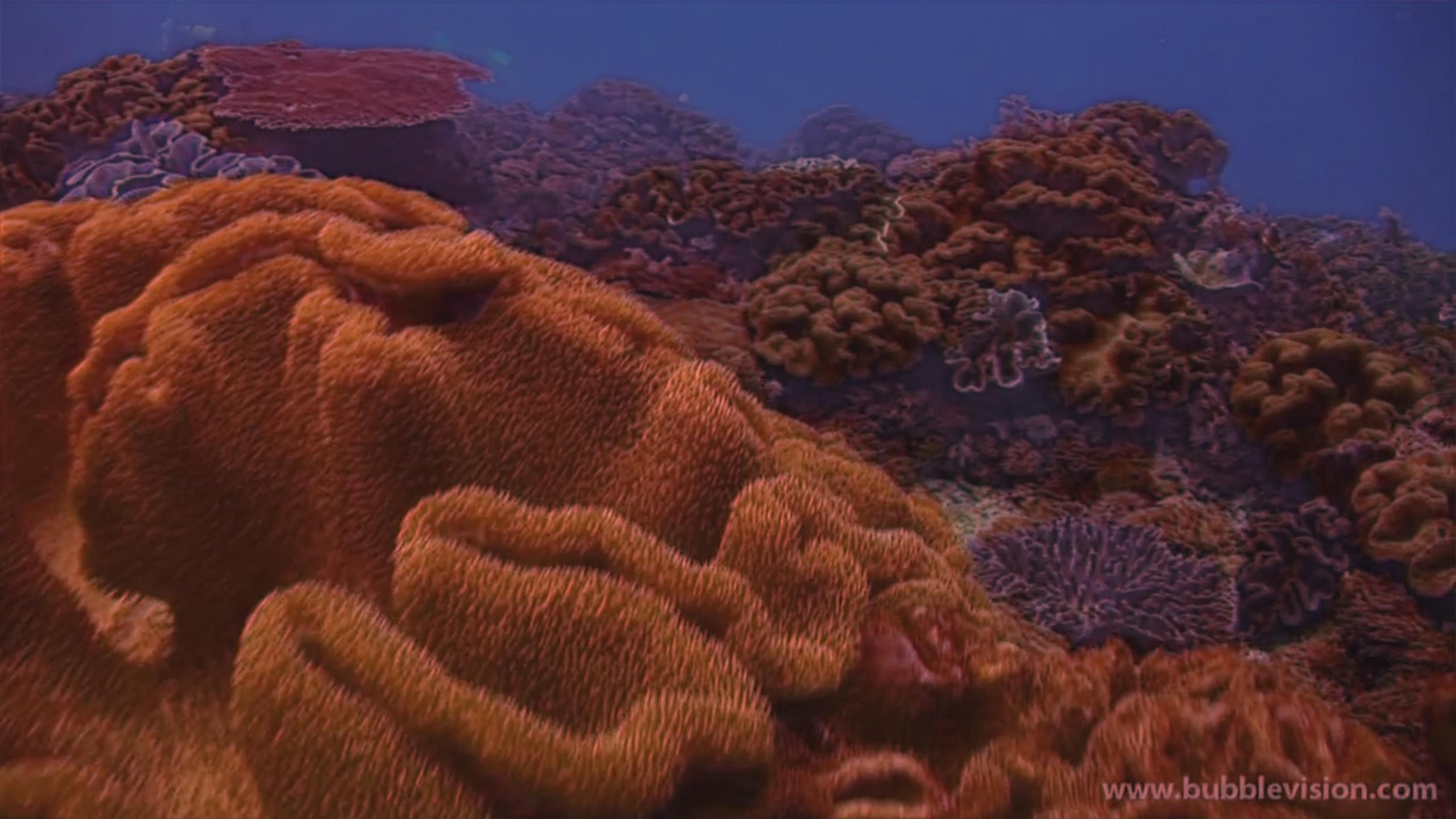}&\includegraphics[height=0.047\textheight]{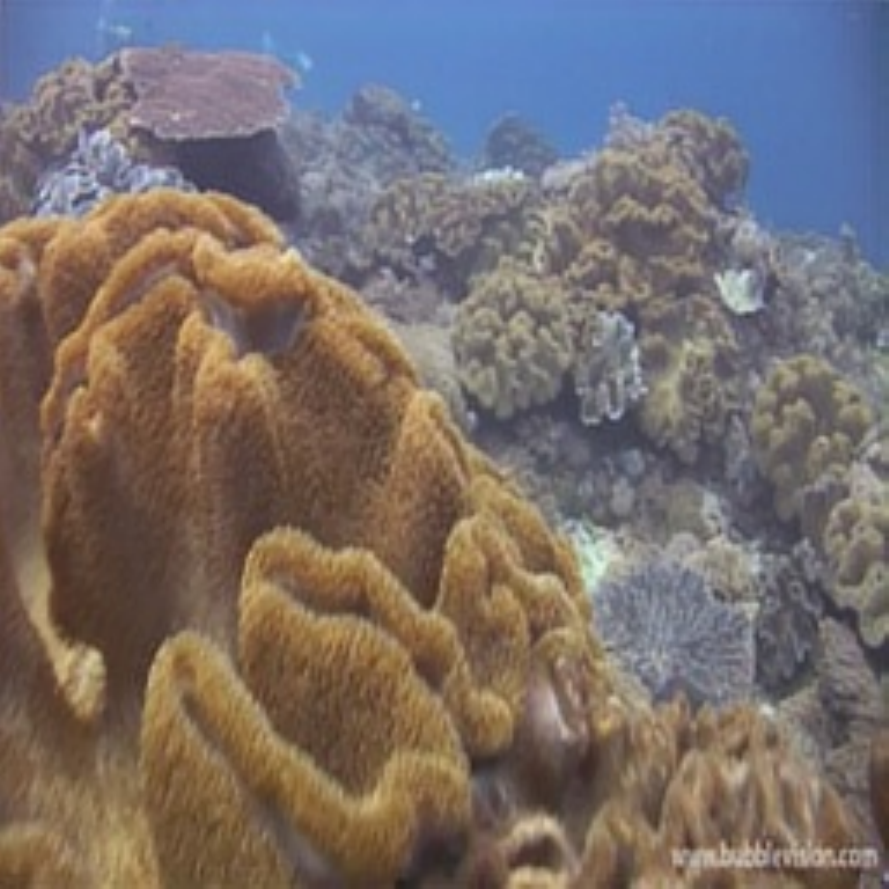}&\includegraphics[height=0.047\textheight]{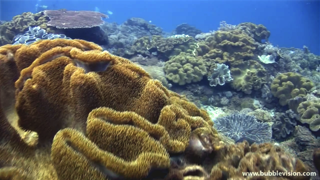}&\includegraphics[height=0.047\textheight]{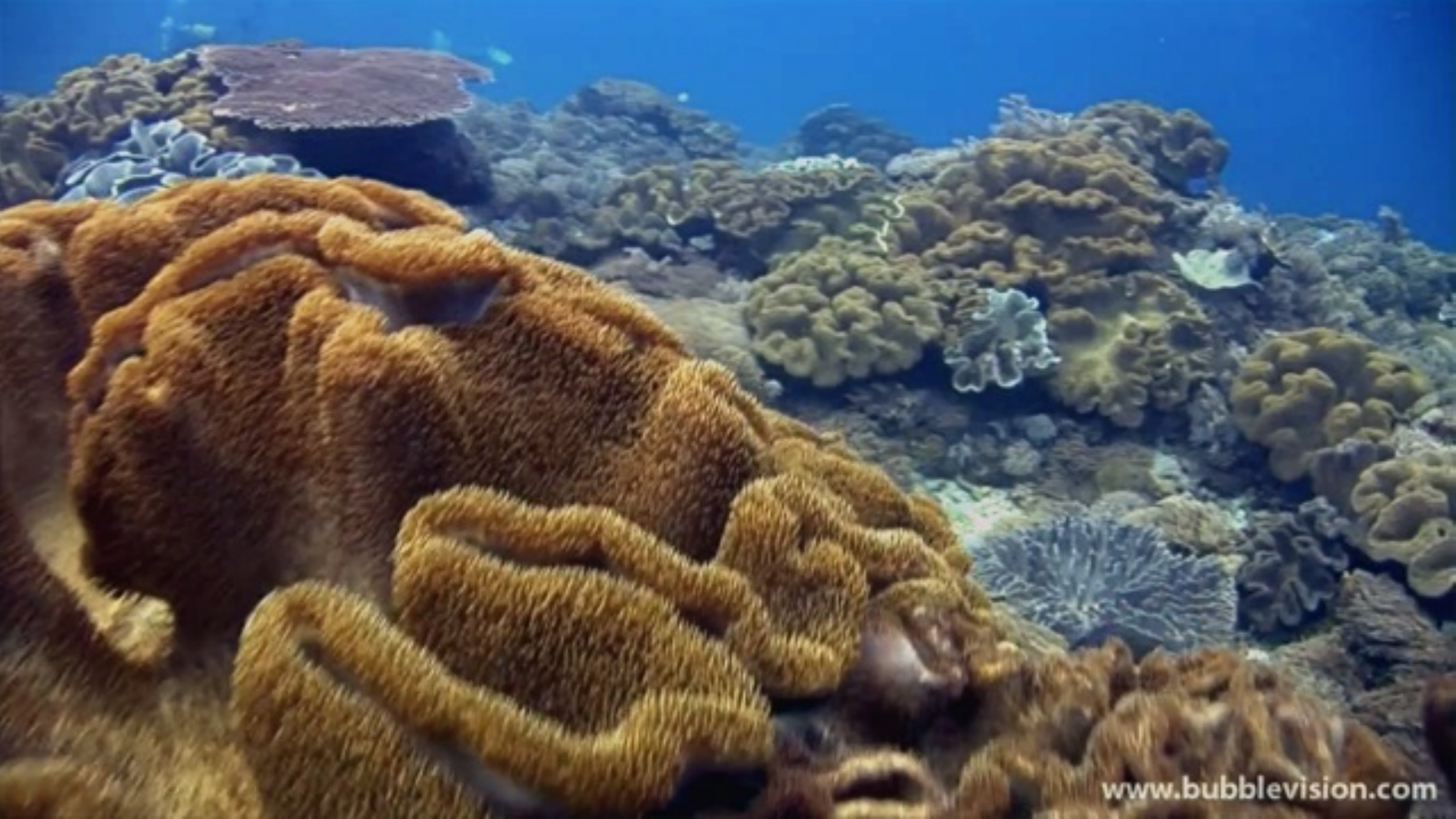} \\
\includegraphics[height=0.047\textheight]{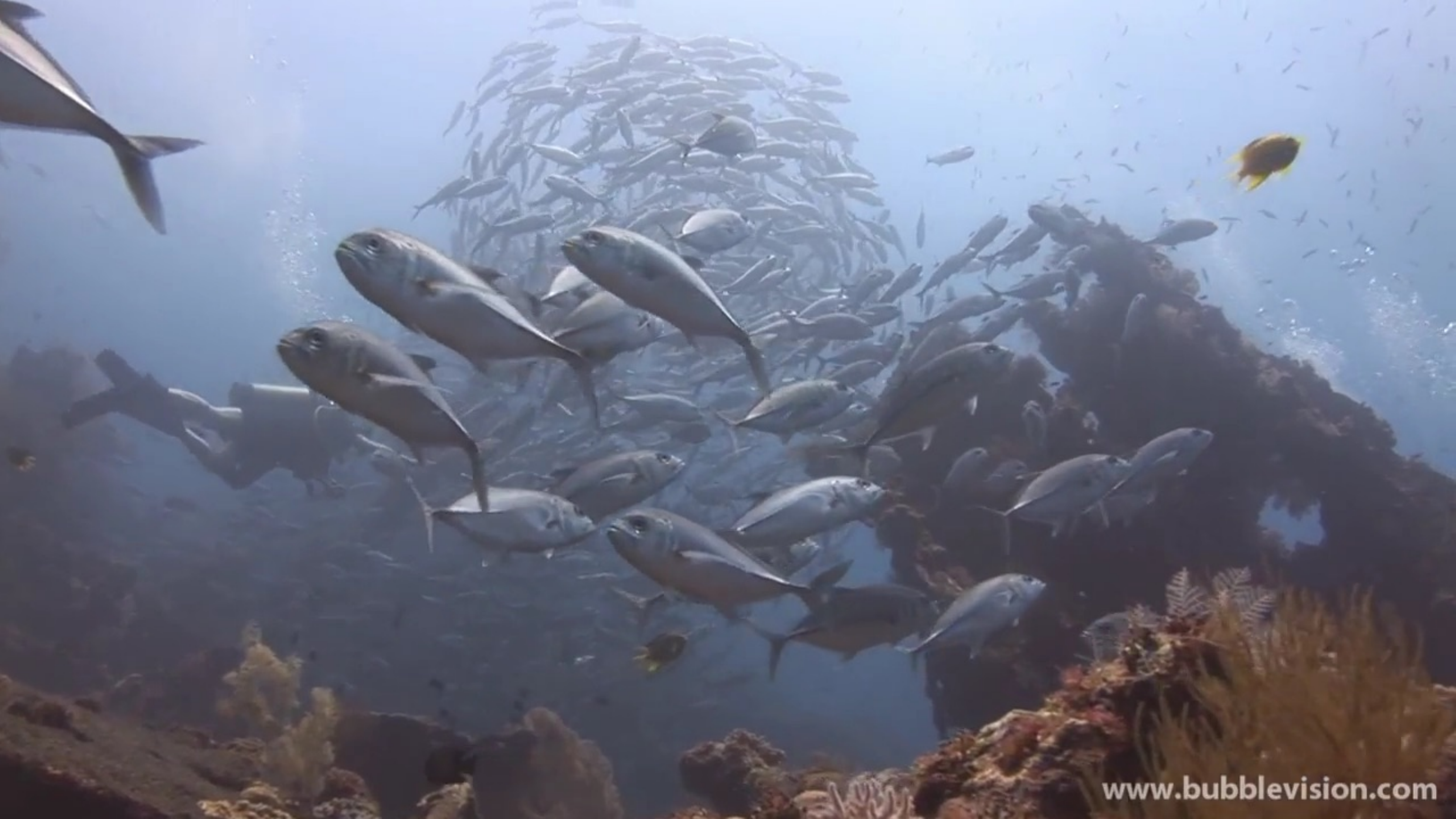}&\includegraphics[height=0.047\textheight]{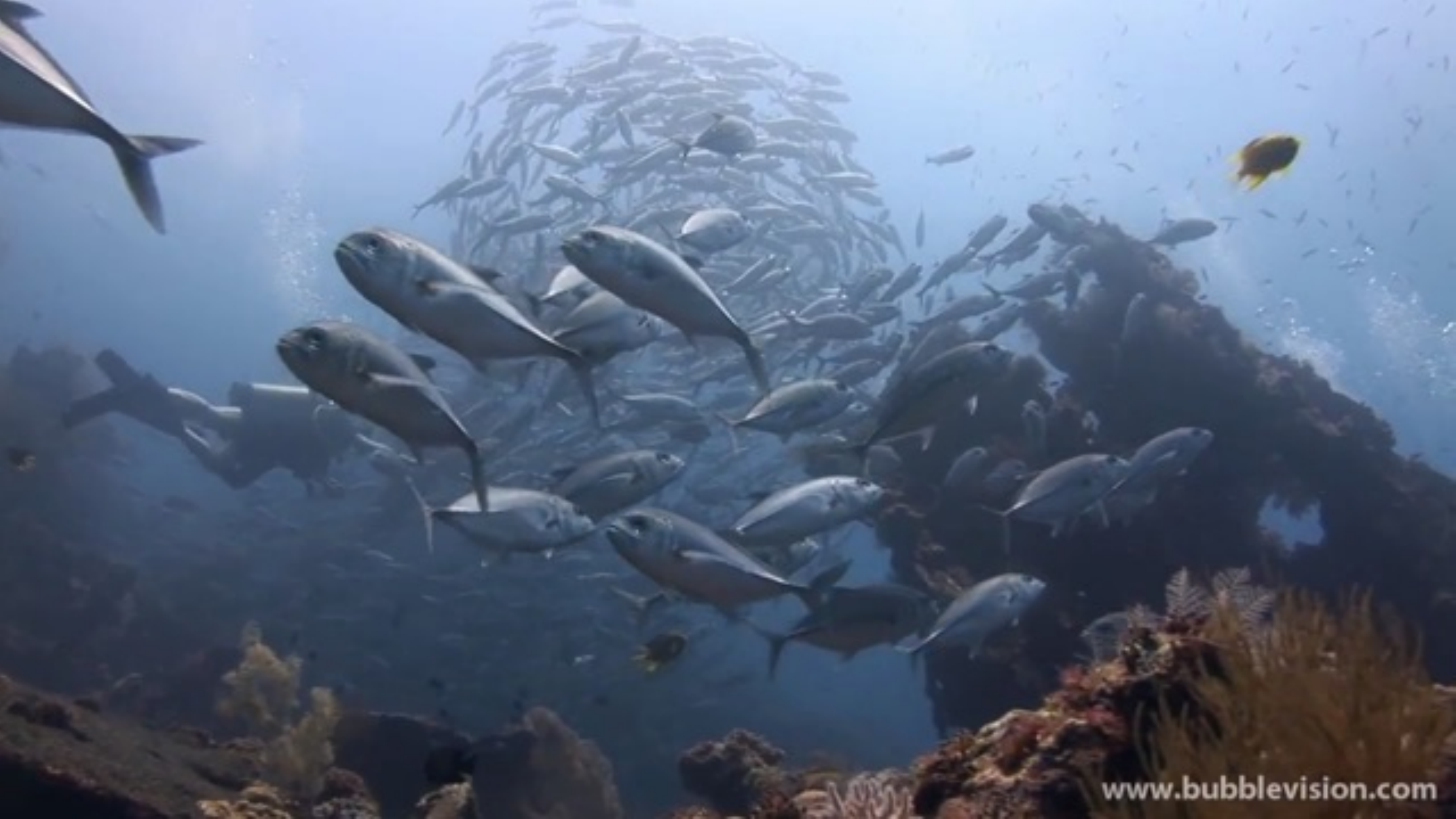}&\includegraphics[height=0.047\textheight]{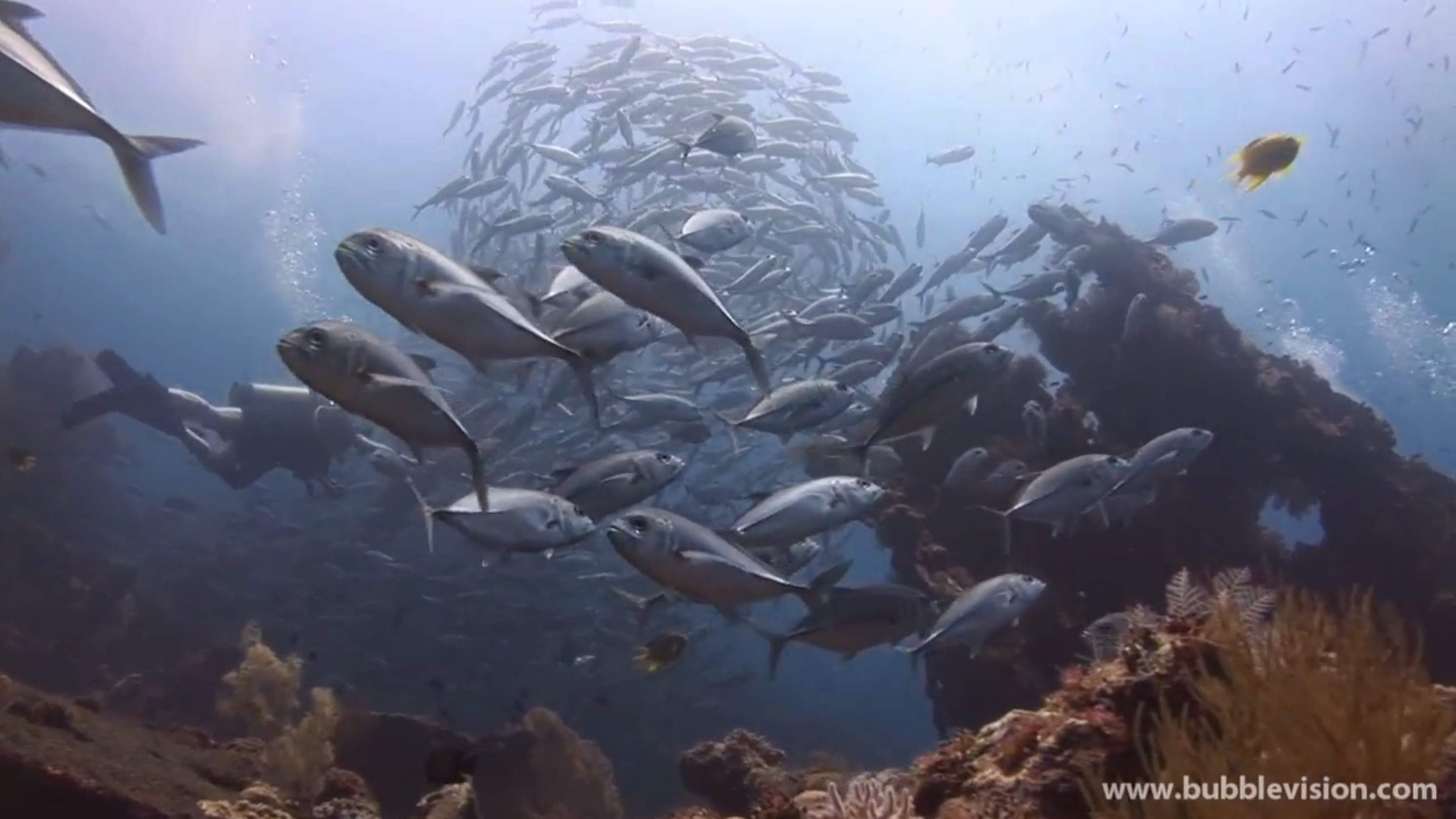}&\includegraphics[height=0.047\textheight]{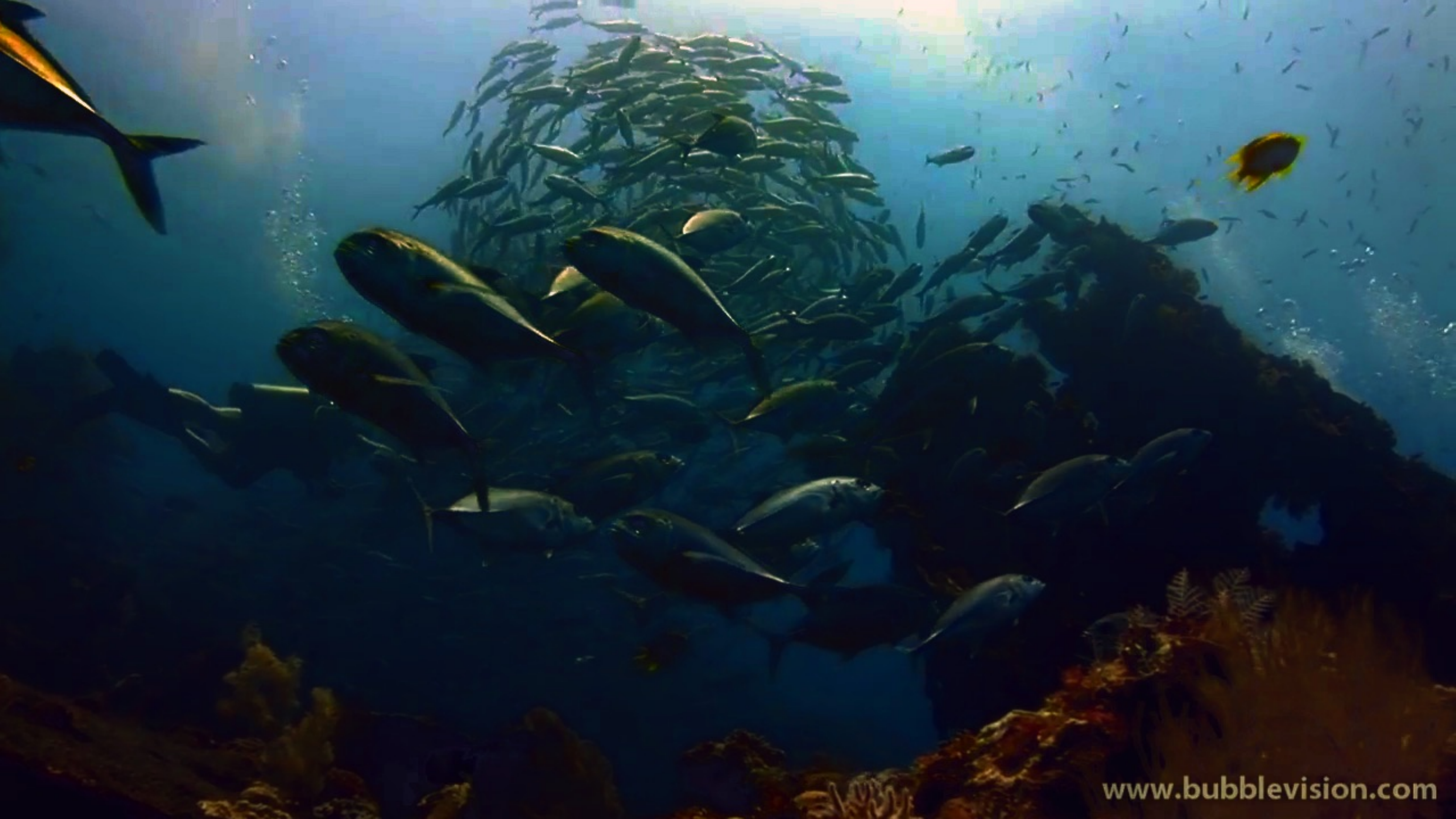}&\includegraphics[height=0.047\textheight]{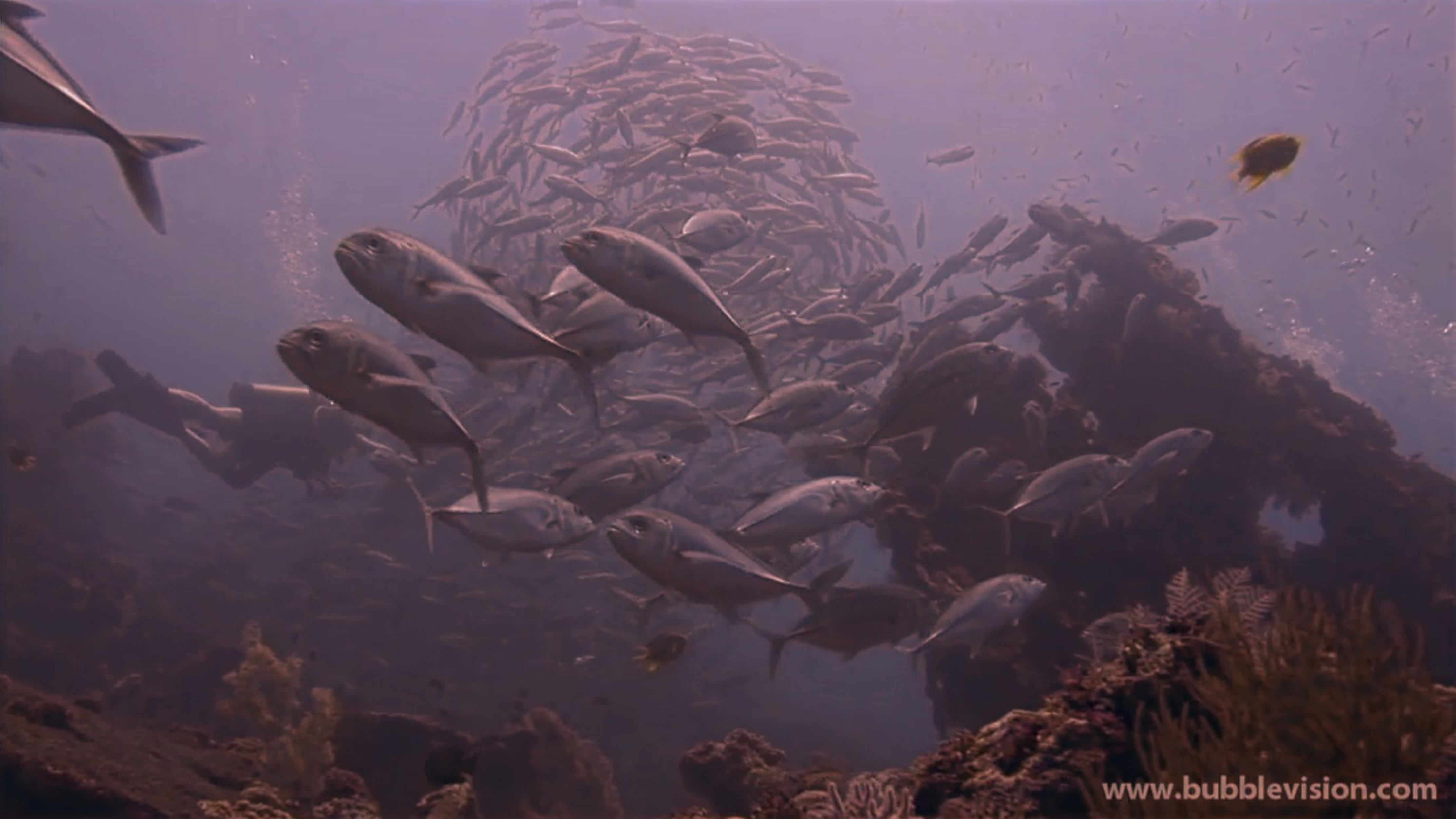}&\includegraphics[height=0.047\textheight]{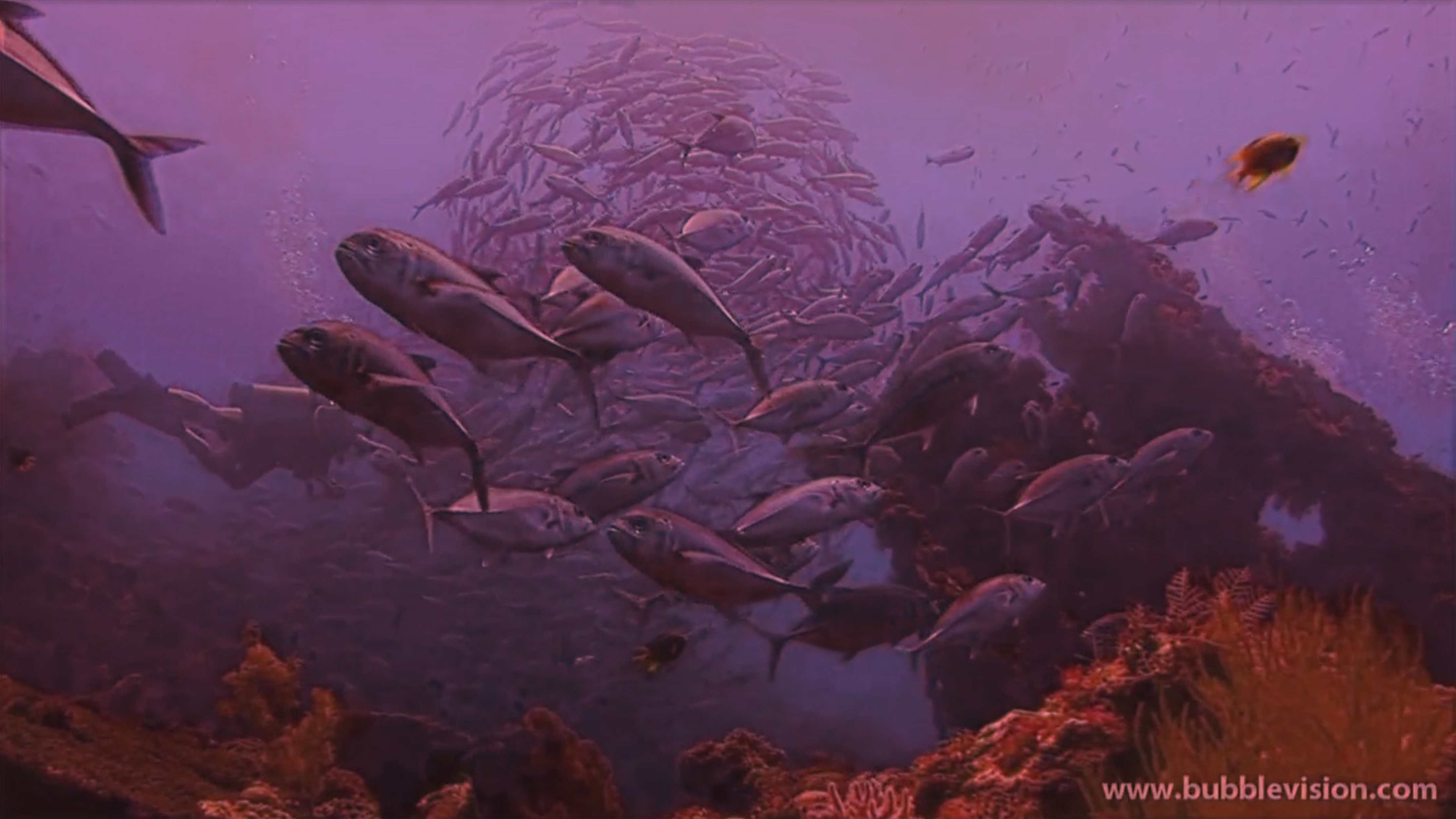} &\includegraphics[height=0.047\textheight]{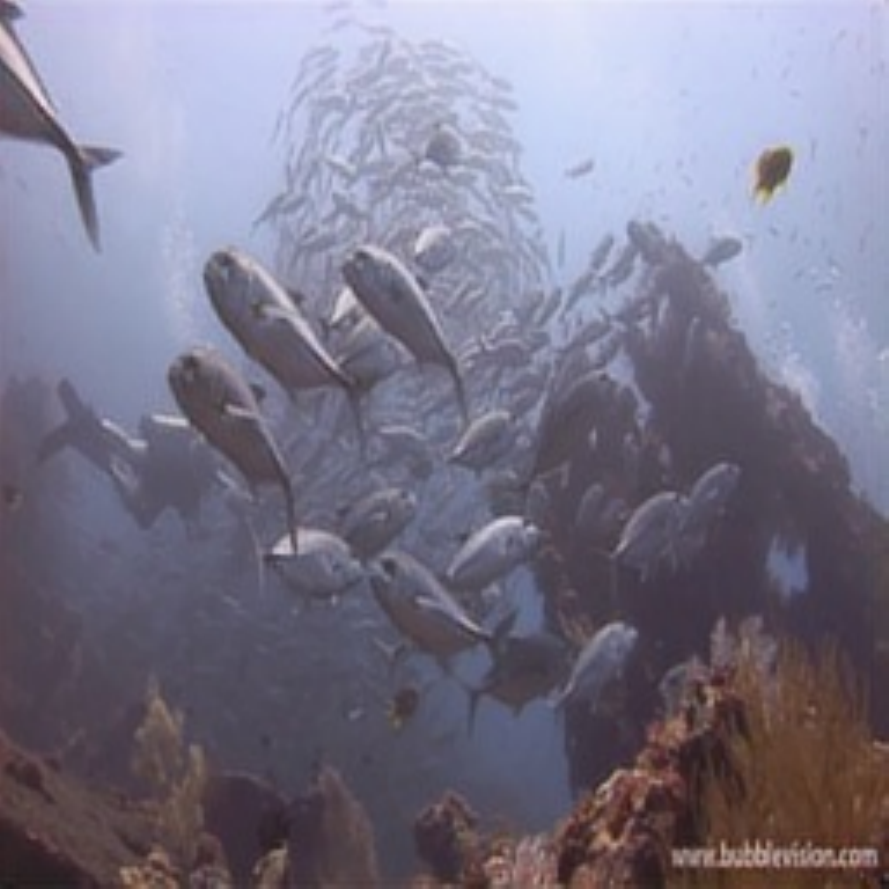}&\includegraphics[height=0.047\textheight]{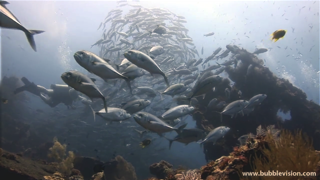}&\includegraphics[height=0.047\textheight]{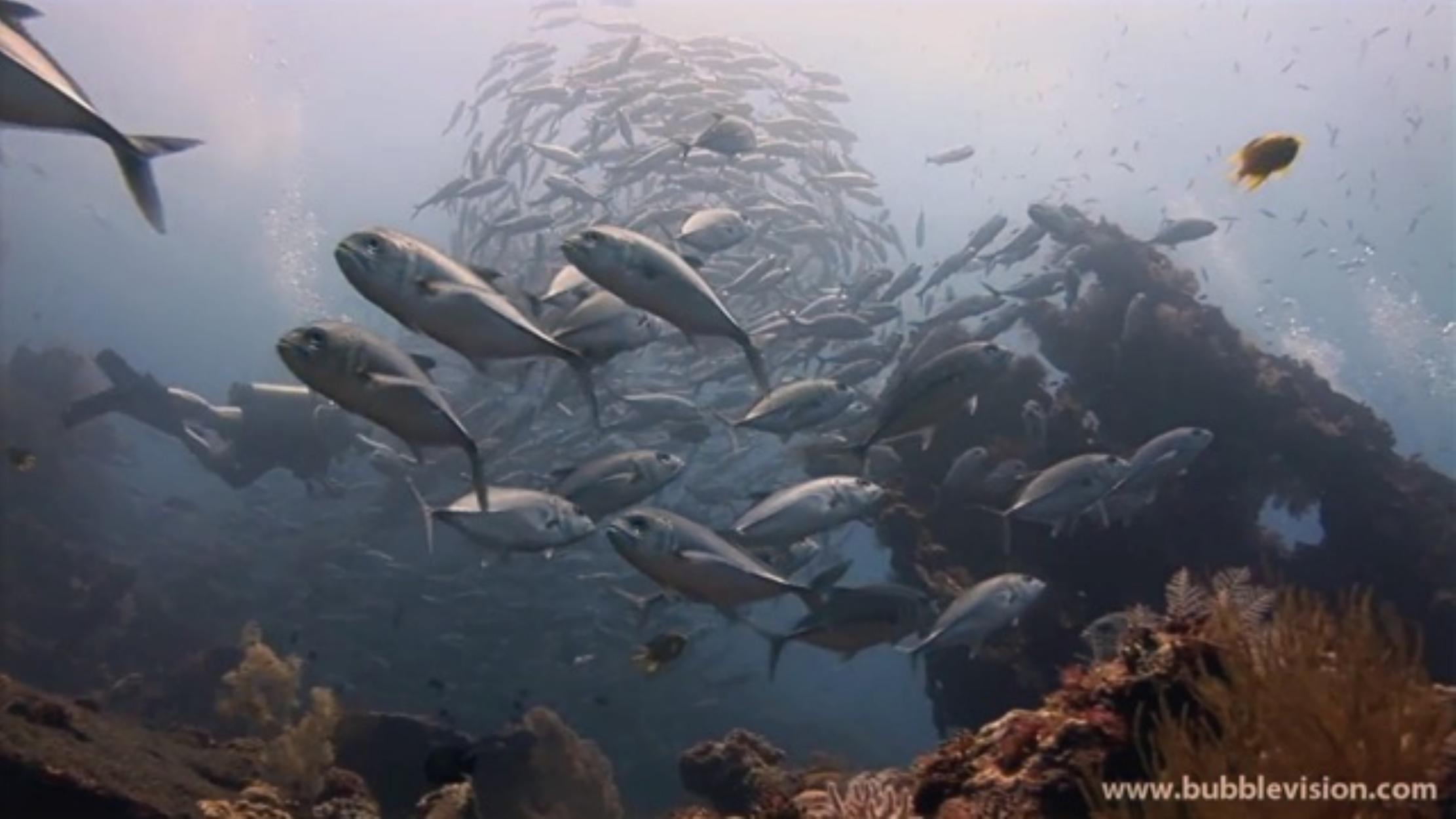}\\
\includegraphics[height=0.064\textheight]{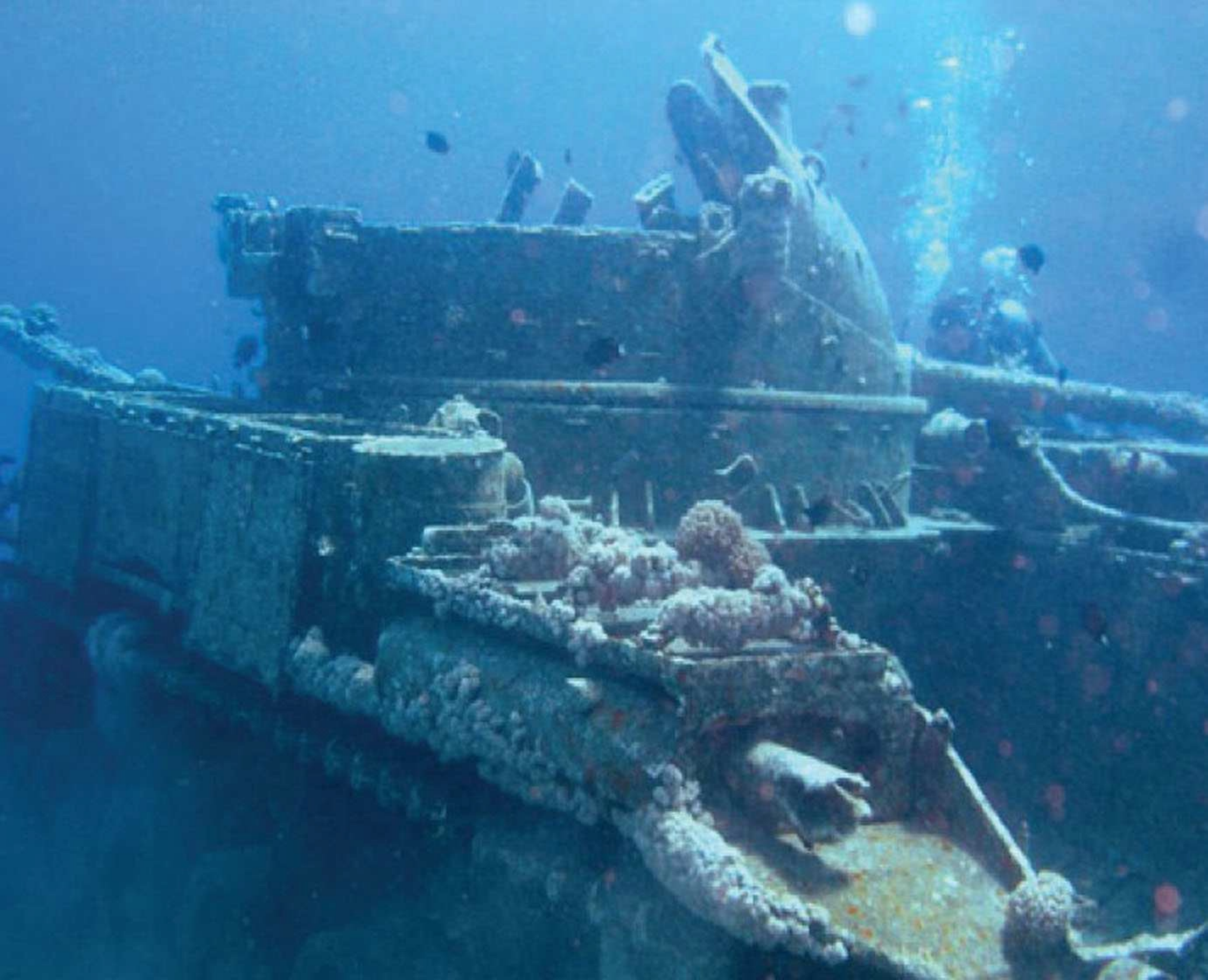}&\includegraphics[height=0.067\textheight]{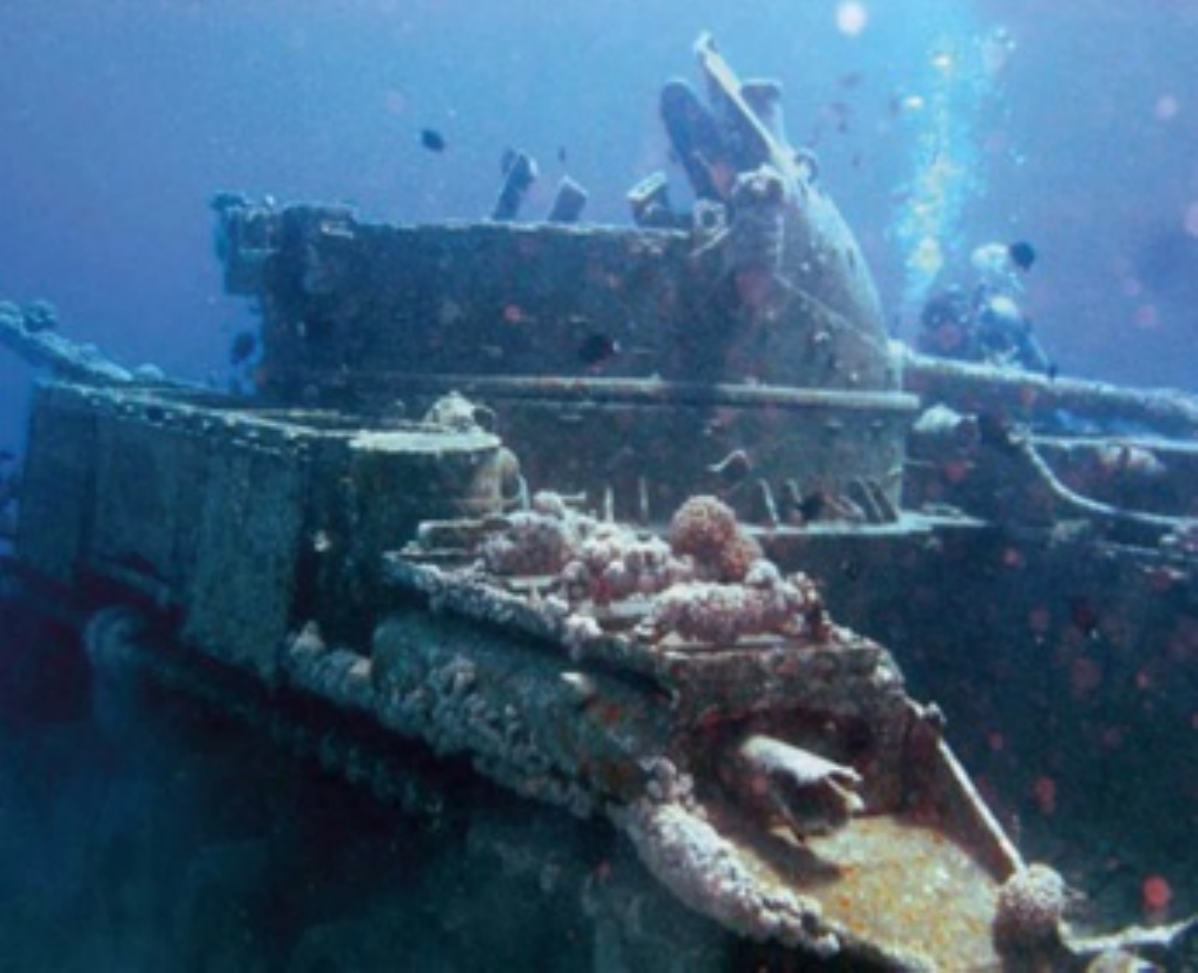}&\includegraphics[height=0.067\textheight]{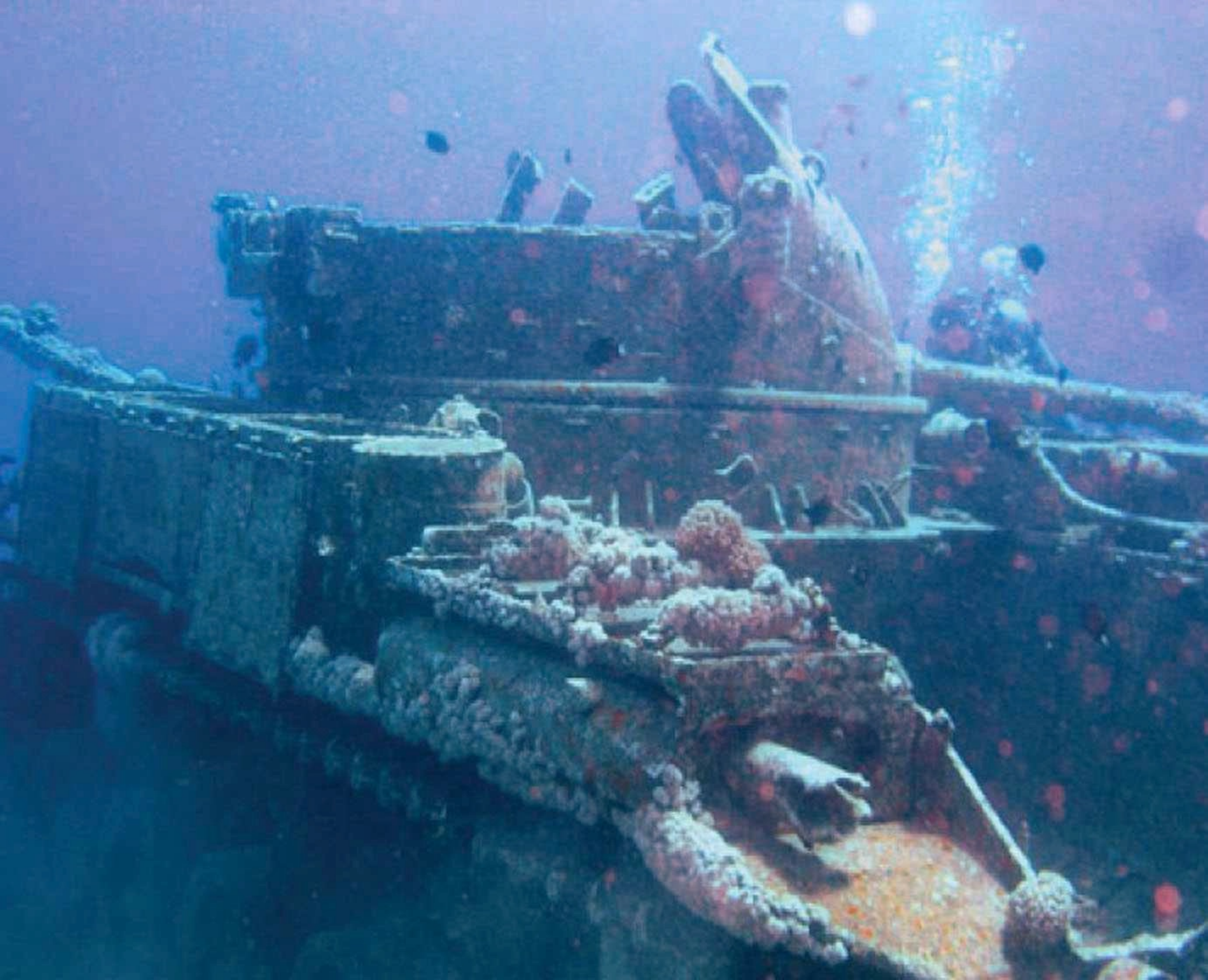}&\includegraphics[height=0.067\textheight]{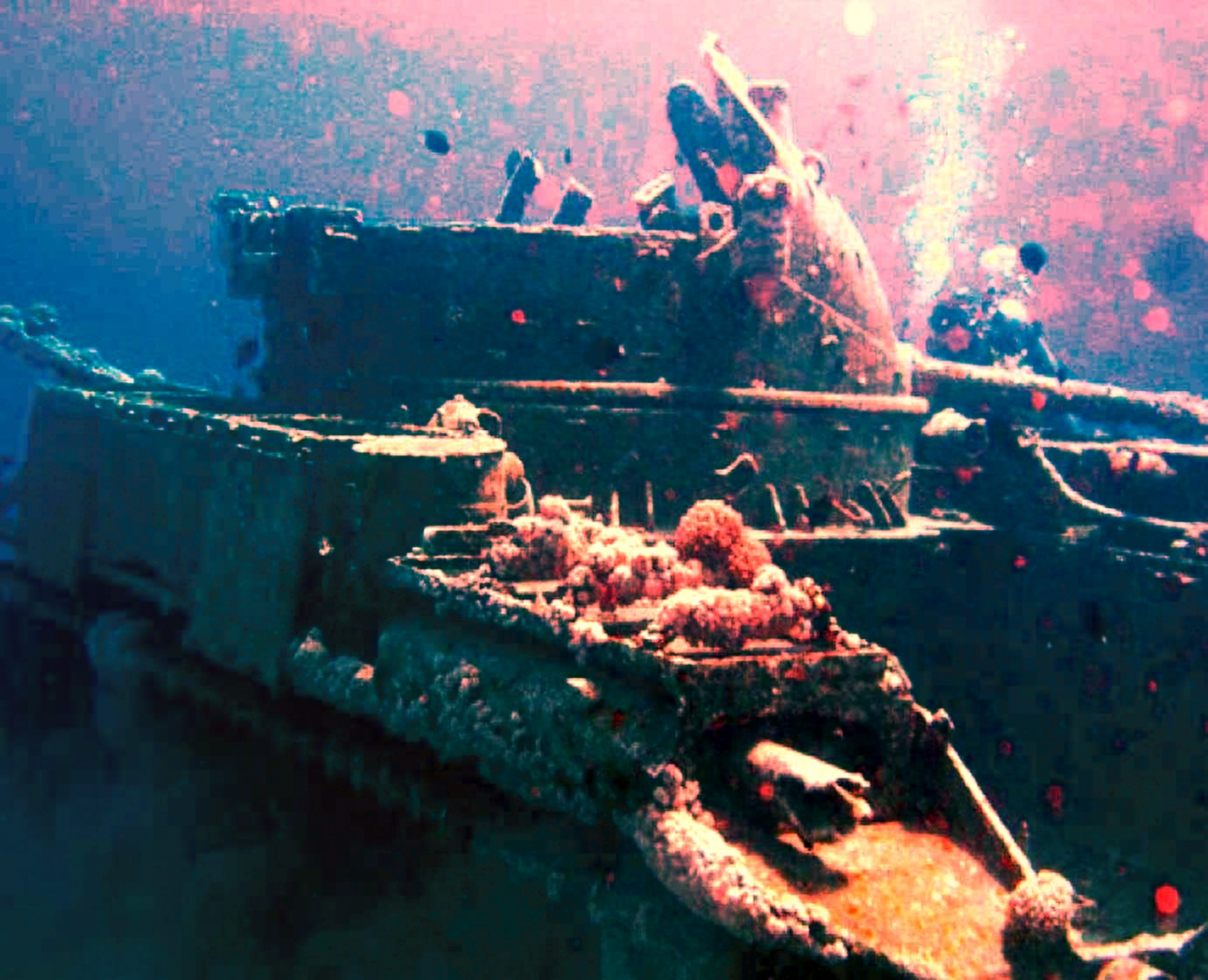}&\includegraphics[height=0.067\textheight]{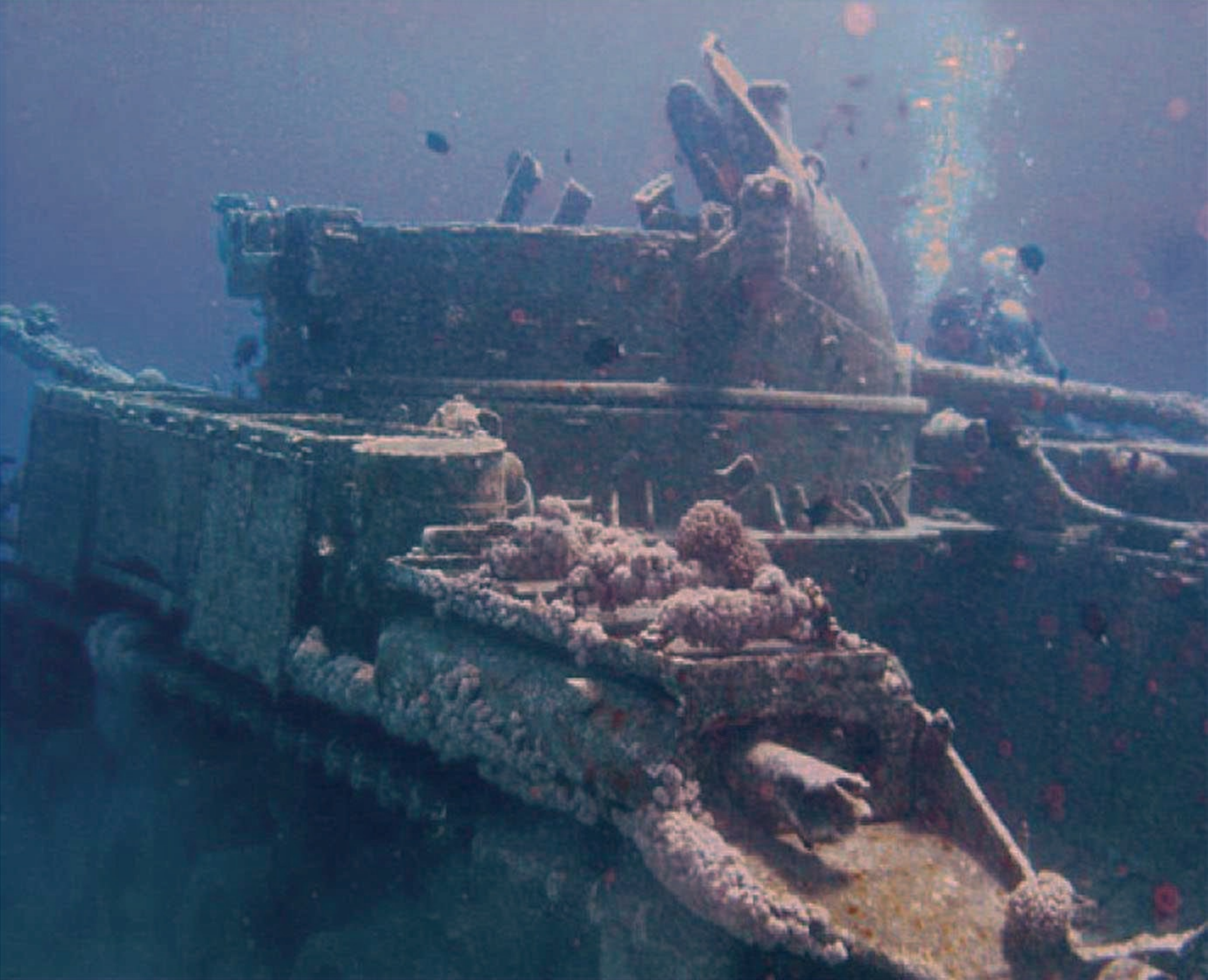}&\includegraphics[height=0.067\textheight]{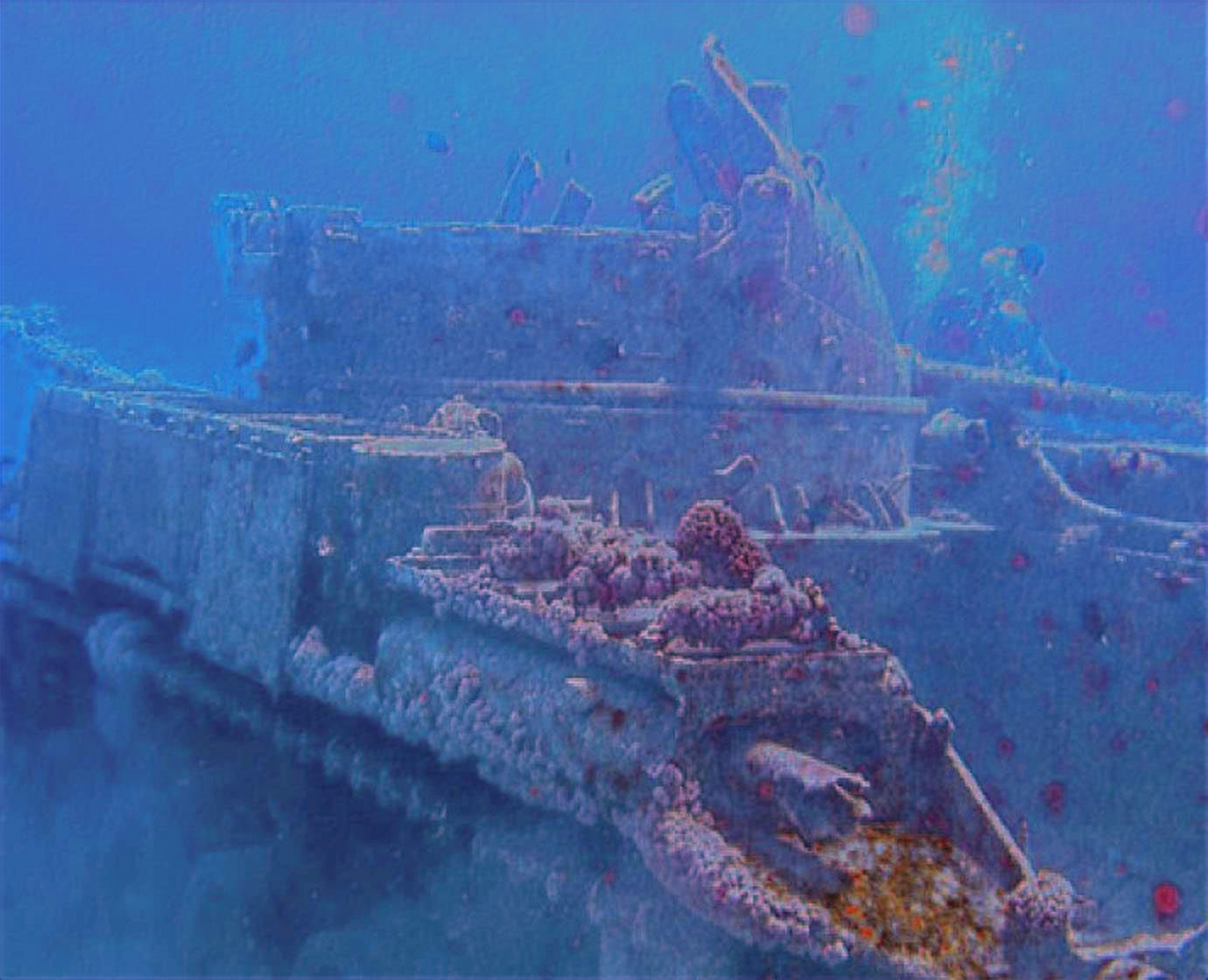}&\includegraphics[height=0.067\textheight]{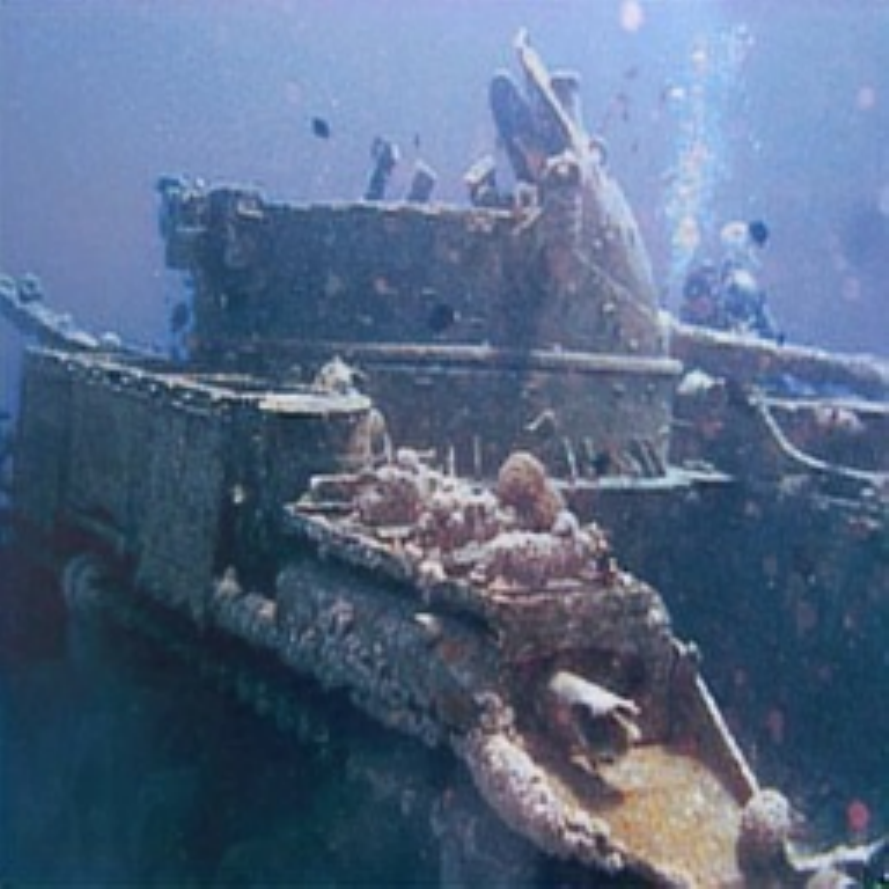}&\includegraphics[height=0.067\textheight]{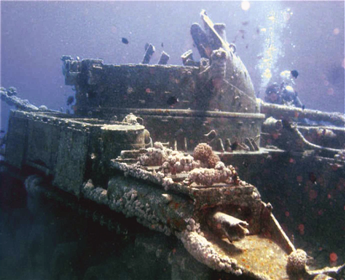}&\includegraphics[height=0.067\textheight]{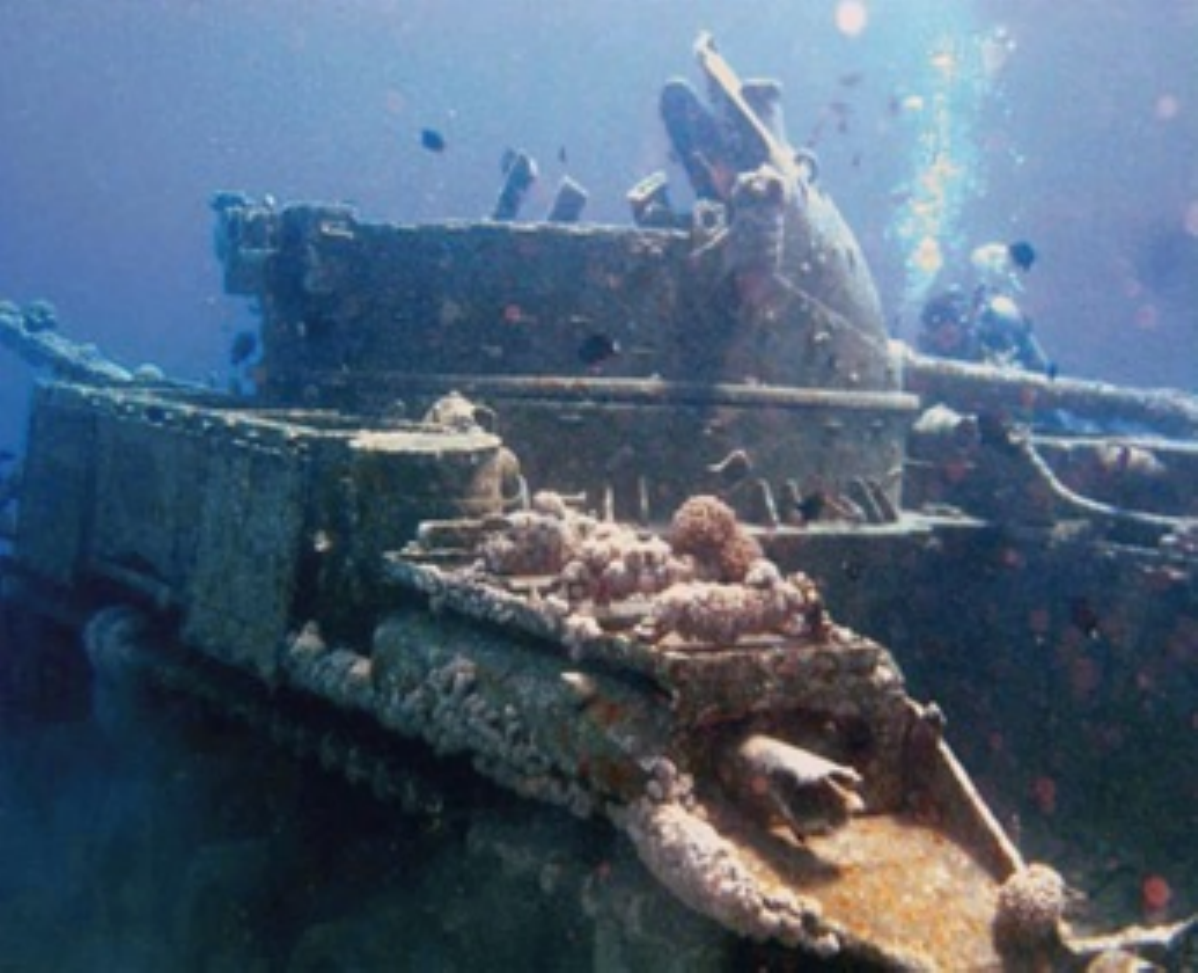} \\
\includegraphics[height=0.059\textheight]{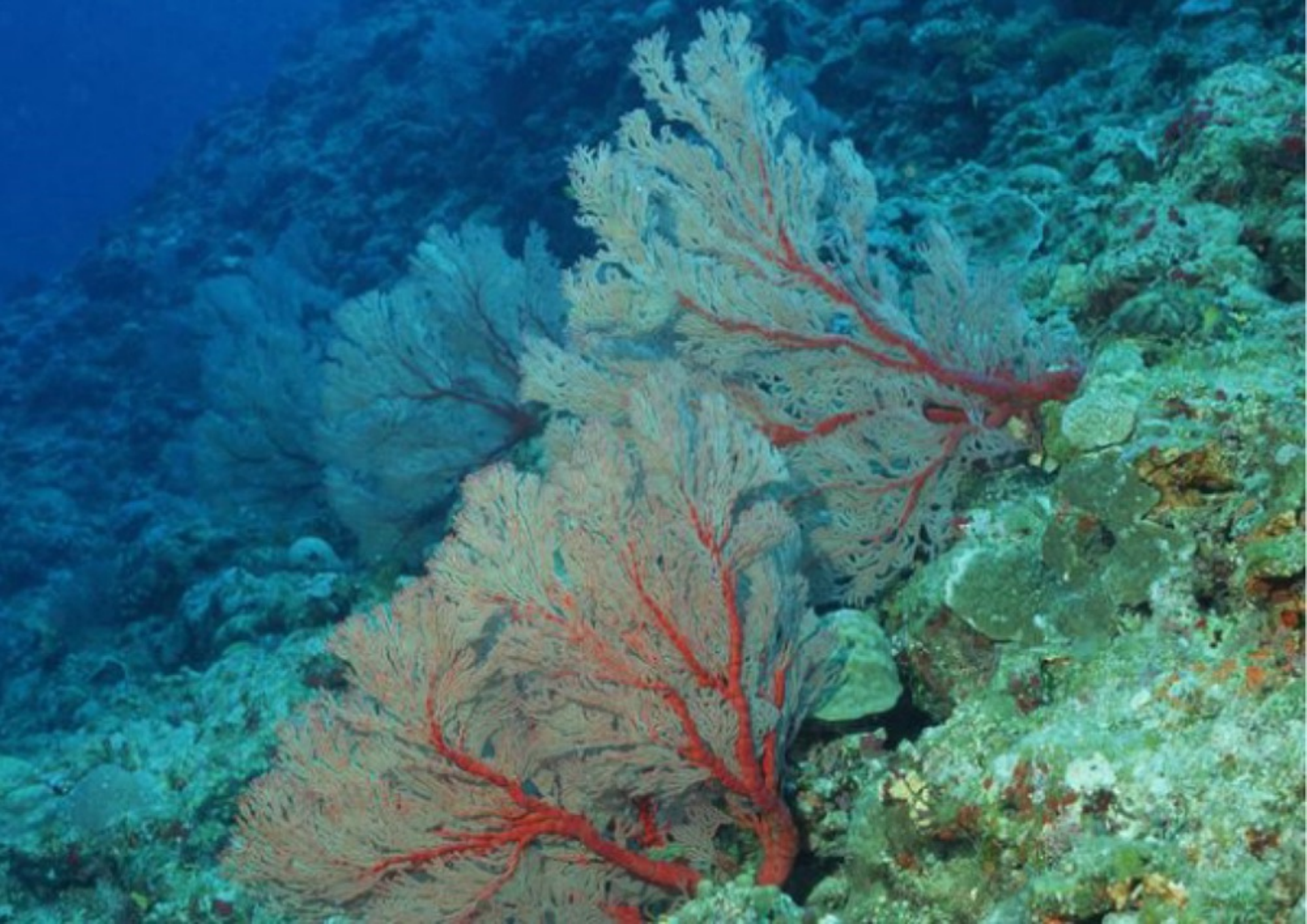}&\includegraphics[height=0.059\textheight]{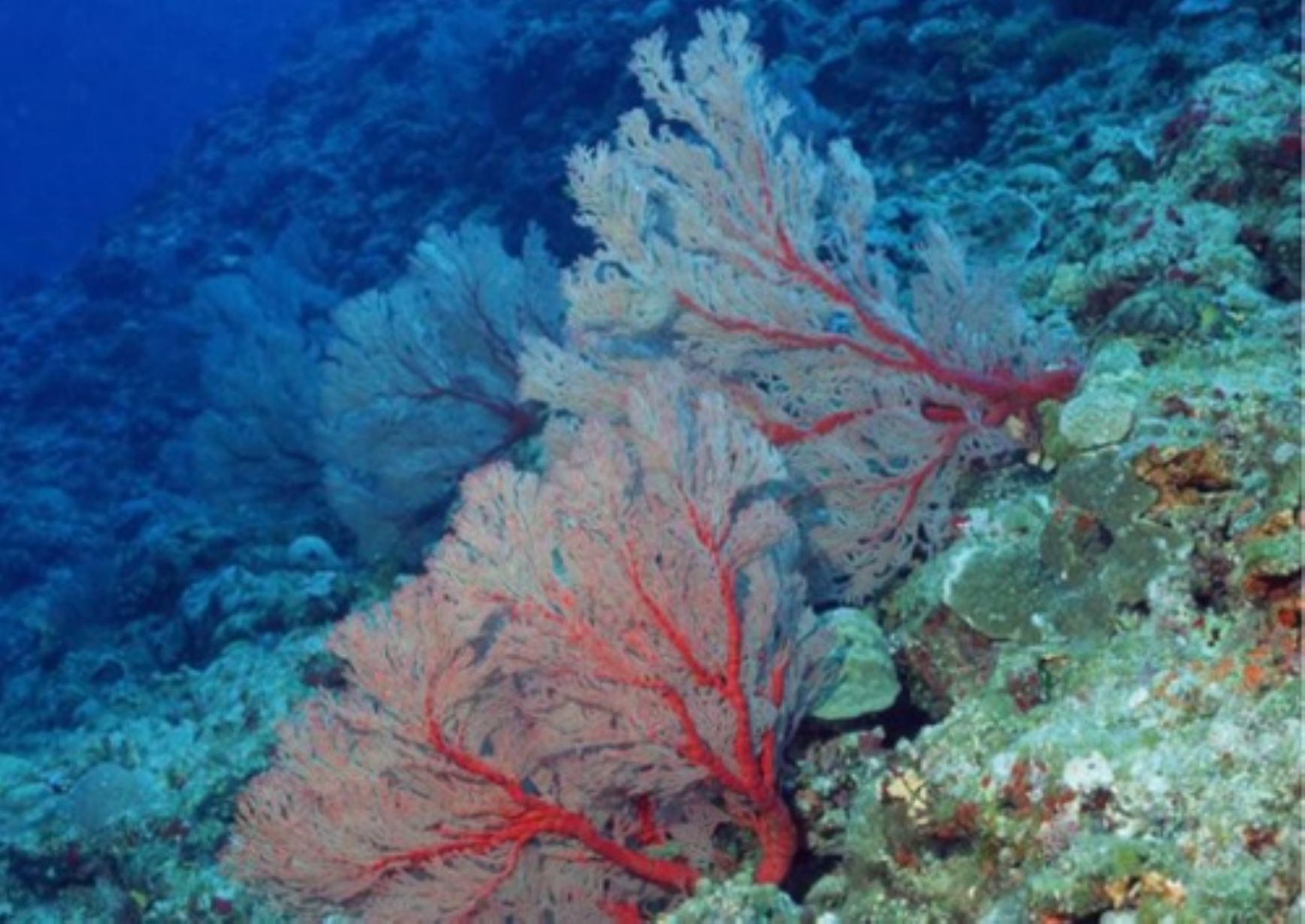}&\includegraphics[height=0.059\textheight]{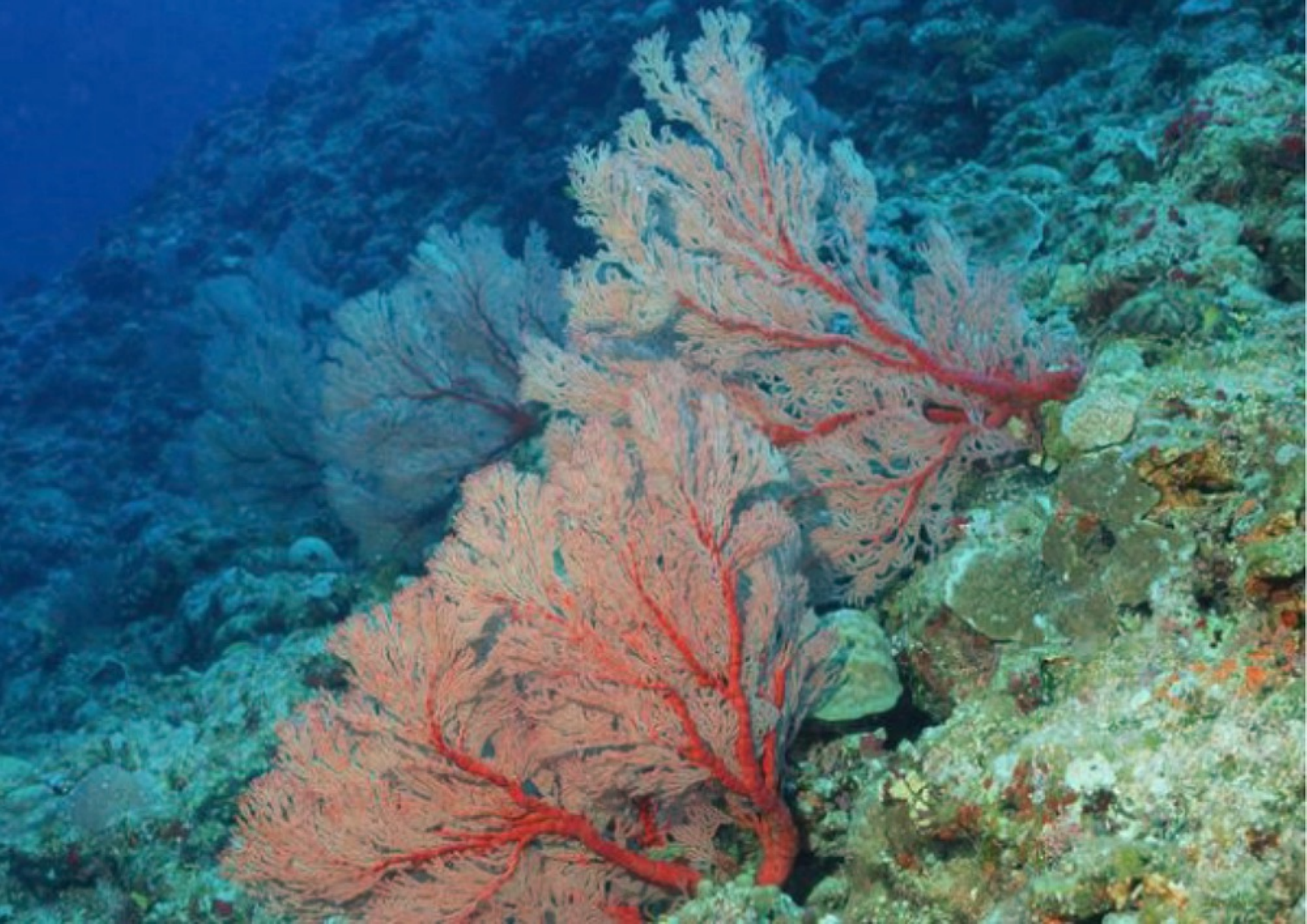}&\includegraphics[height=0.059\textheight]{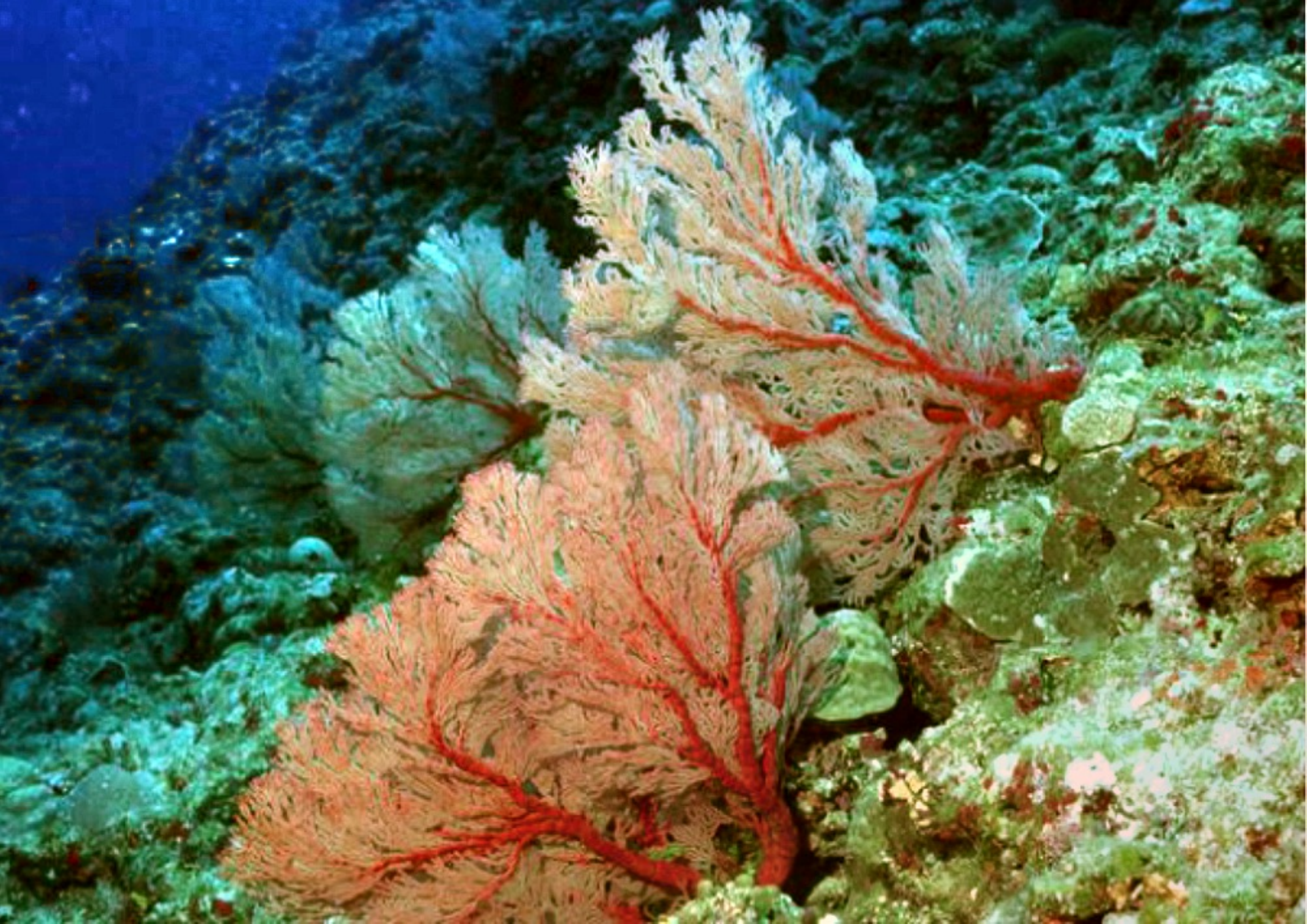}&\includegraphics[height=0.059\textheight]{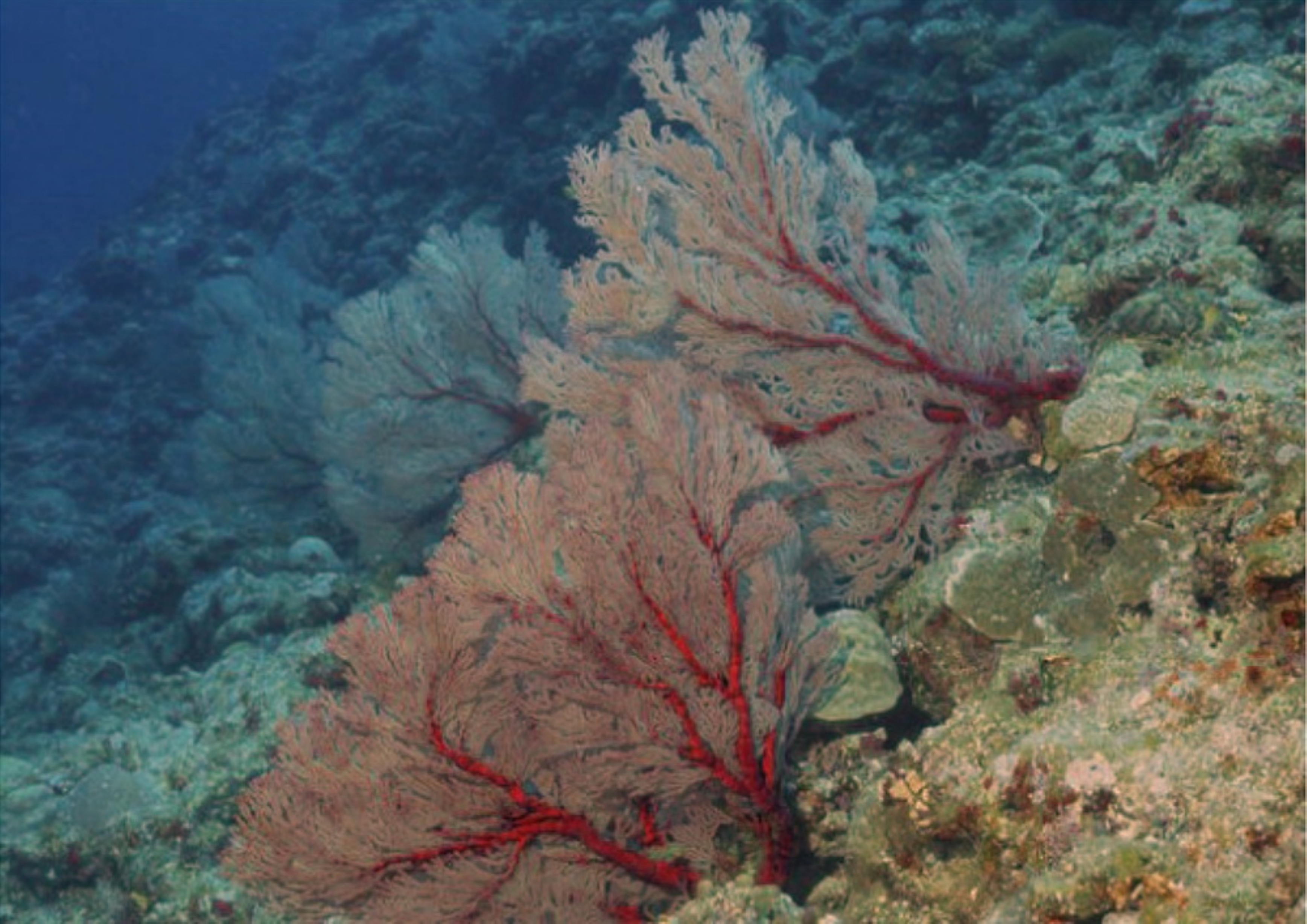}&\includegraphics[height=0.059\textheight]{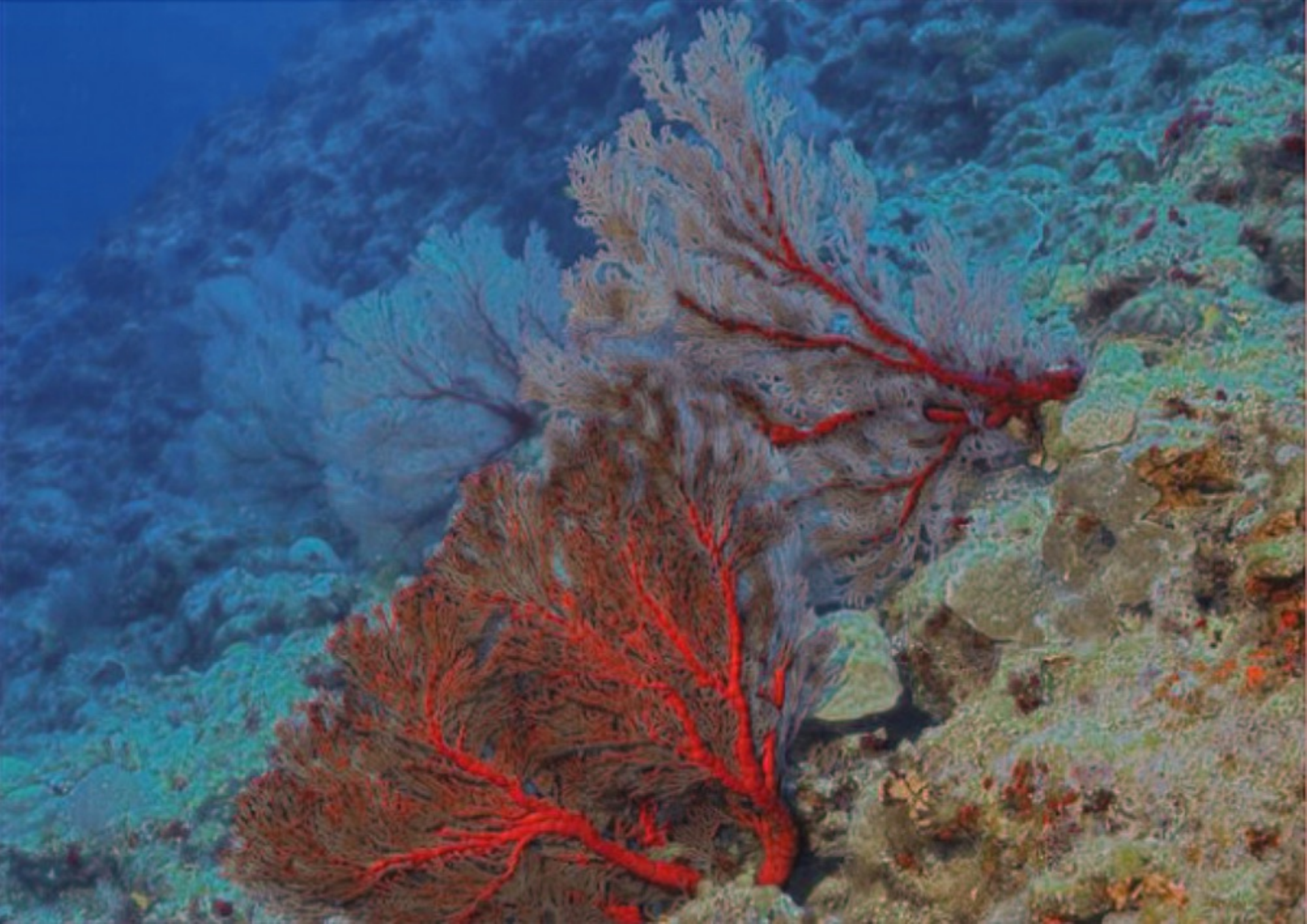}&\includegraphics[height=0.059\textheight]{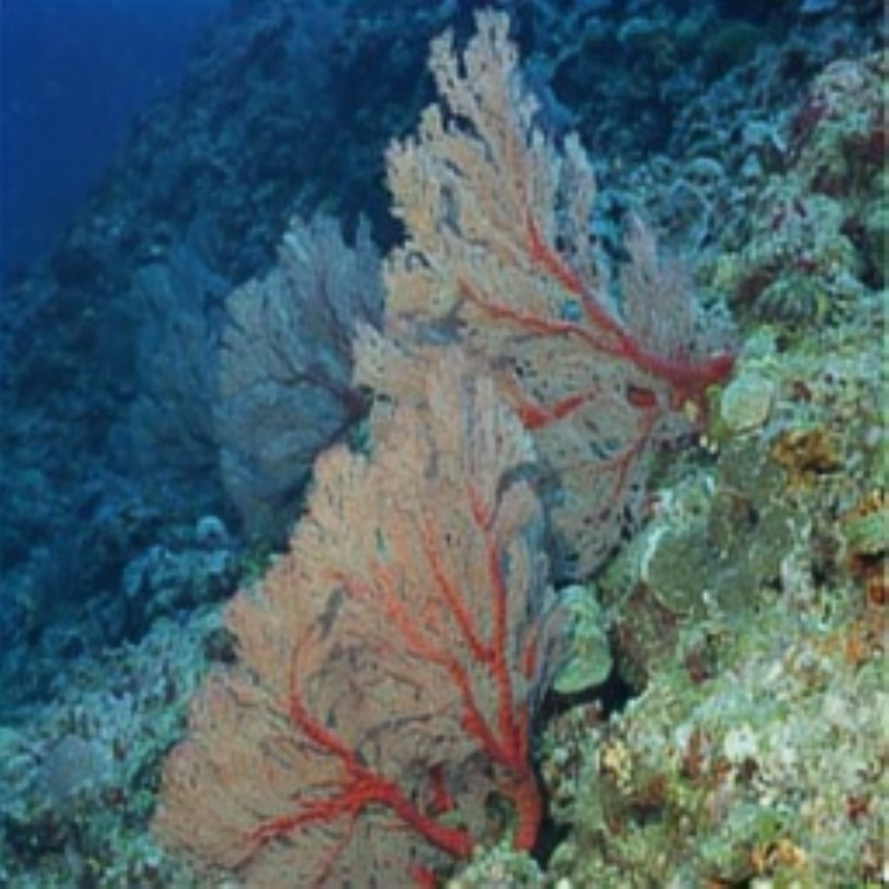}&\includegraphics[height=0.059\textheight]{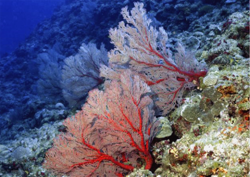}&\includegraphics[height=0.059\textheight]{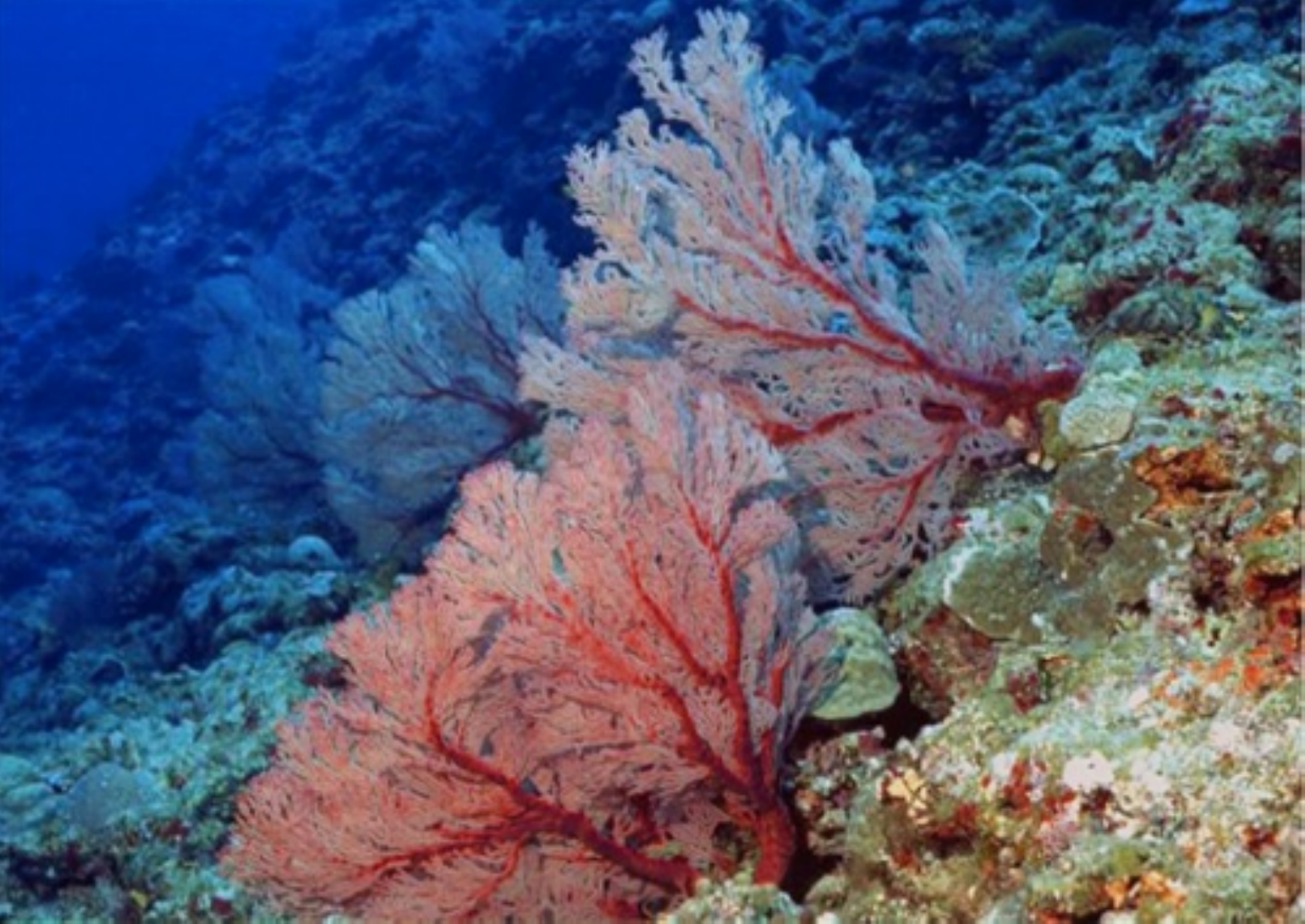}\\
\includegraphics[height=0.060\textheight]{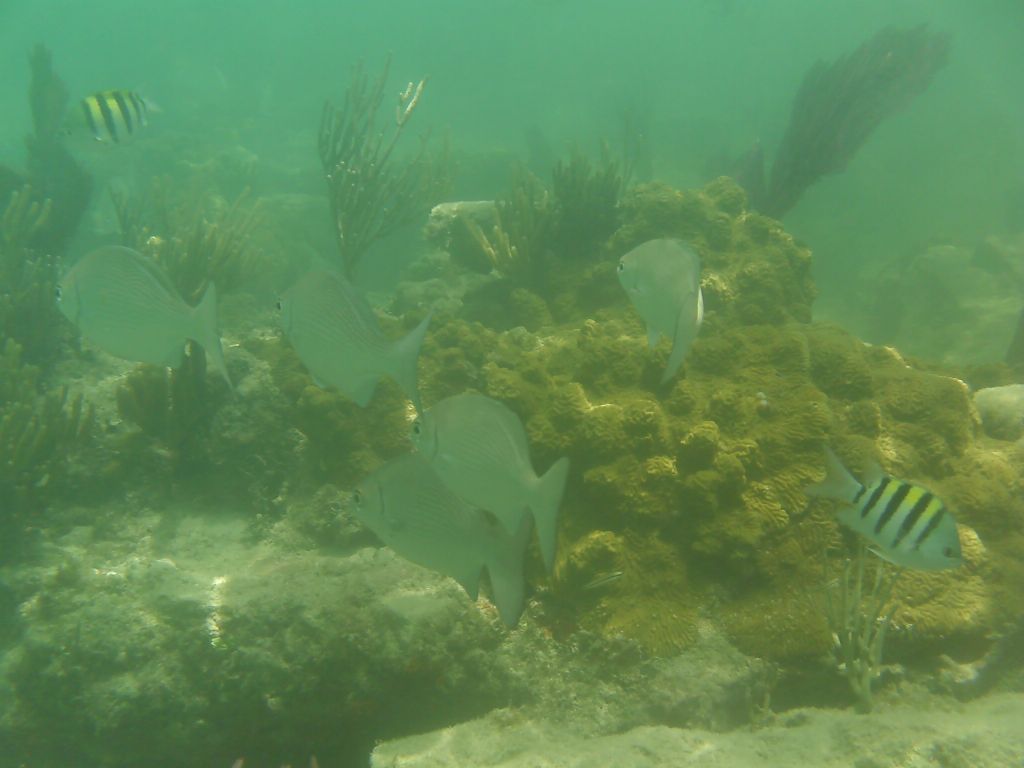}&\includegraphics[height=0.063\textheight]{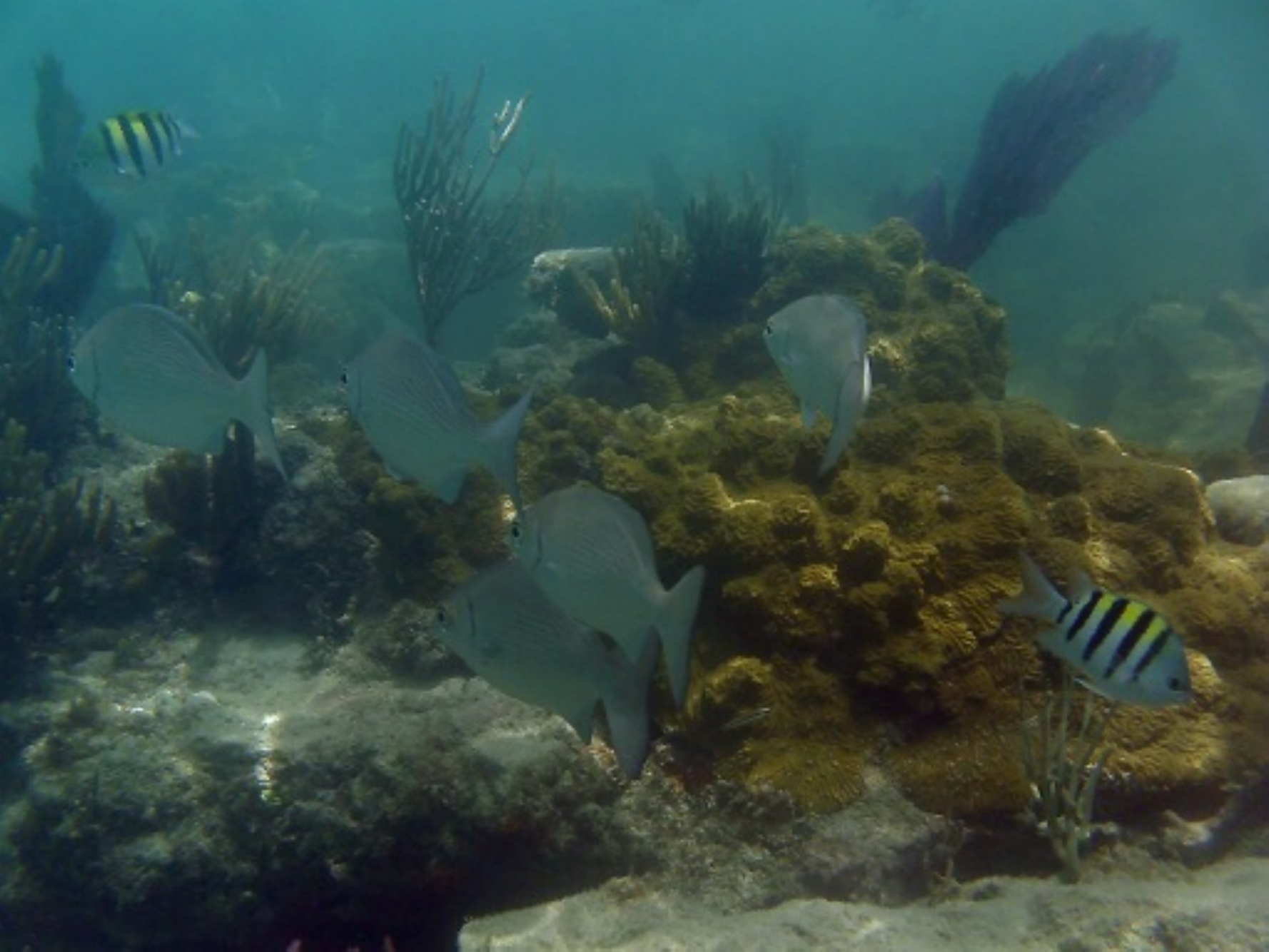}&\includegraphics[height=0.063\textheight]{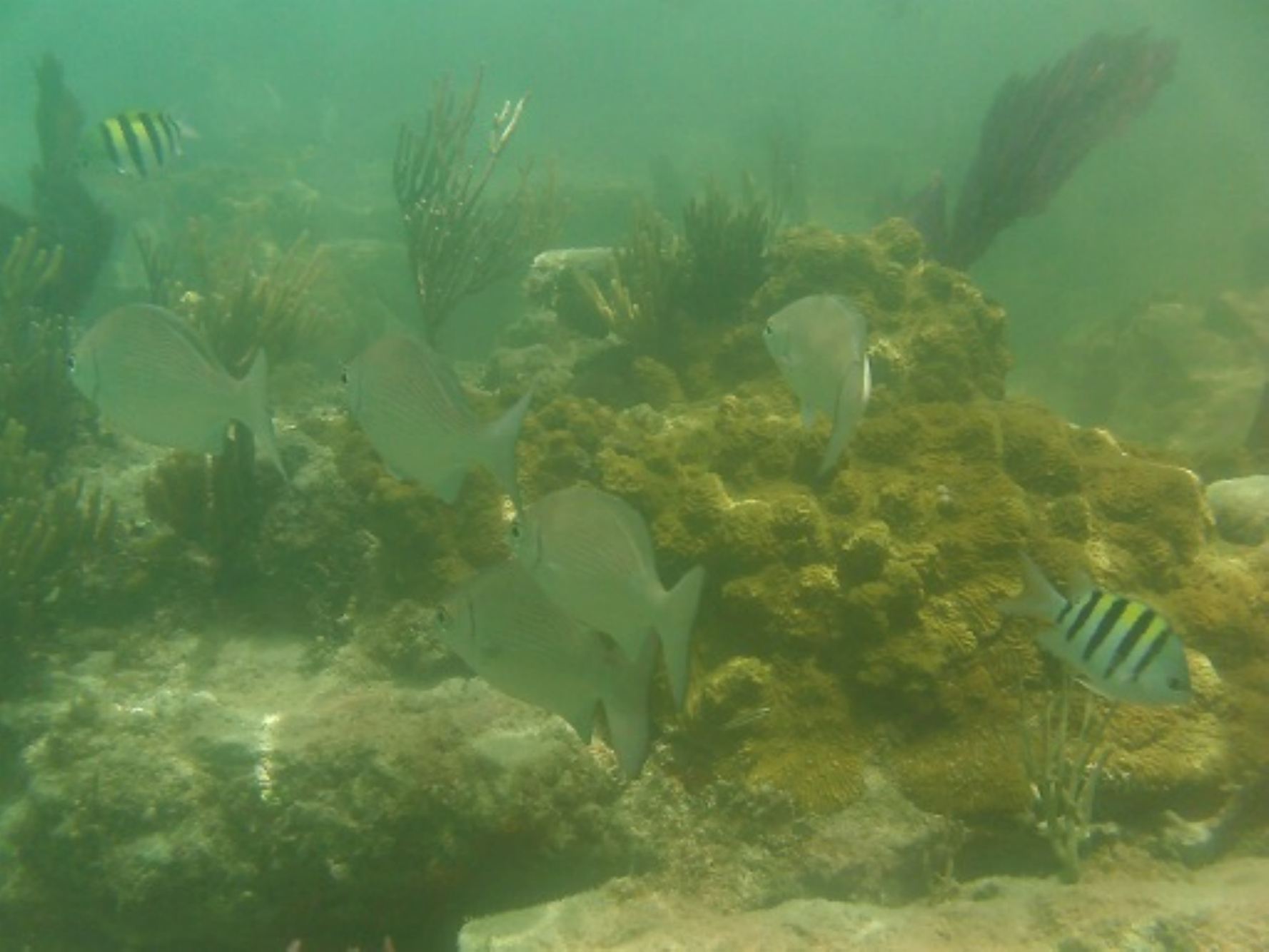}&\includegraphics[height=0.063\textheight]{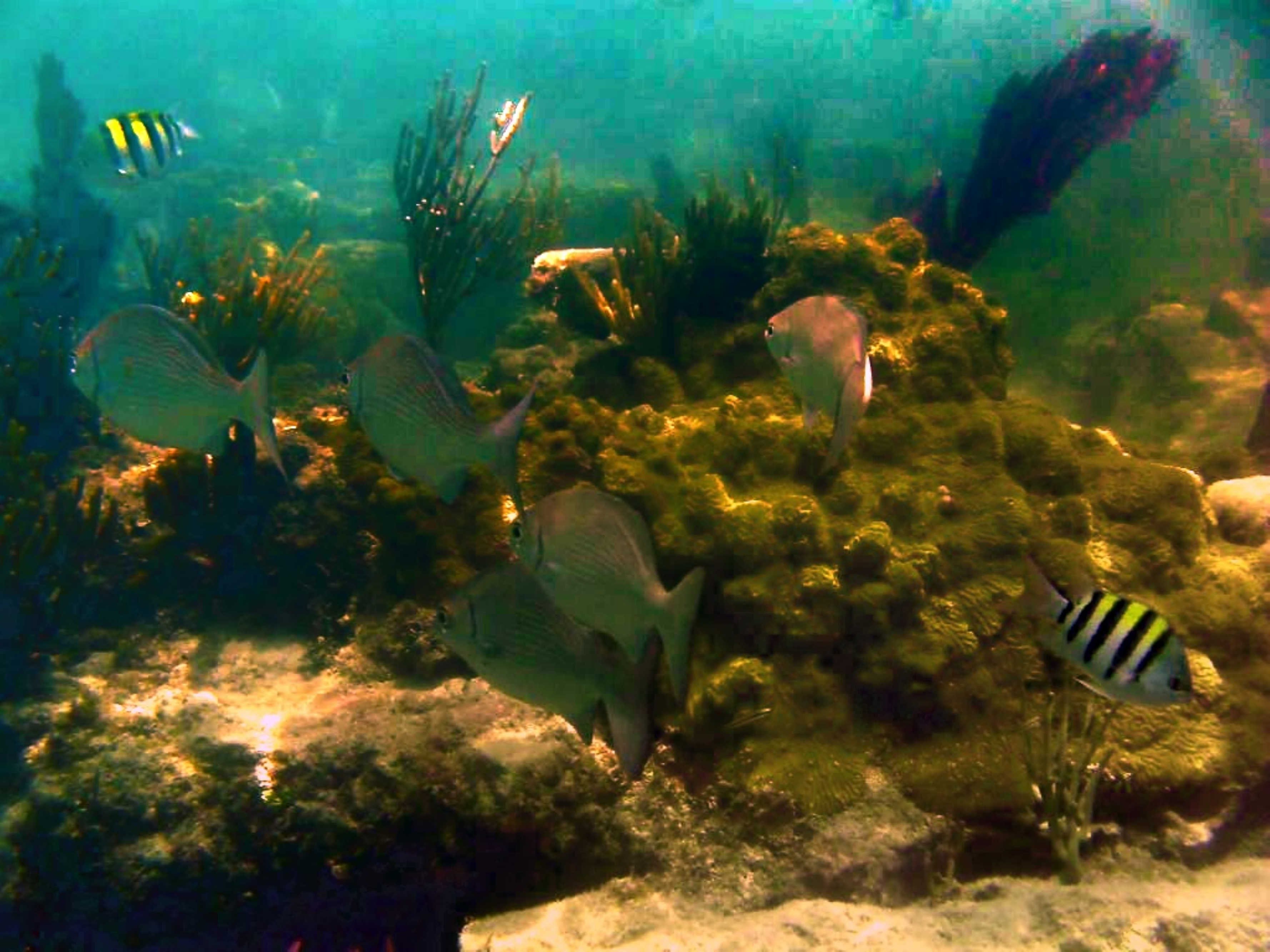}&\includegraphics[height=0.063\textheight]{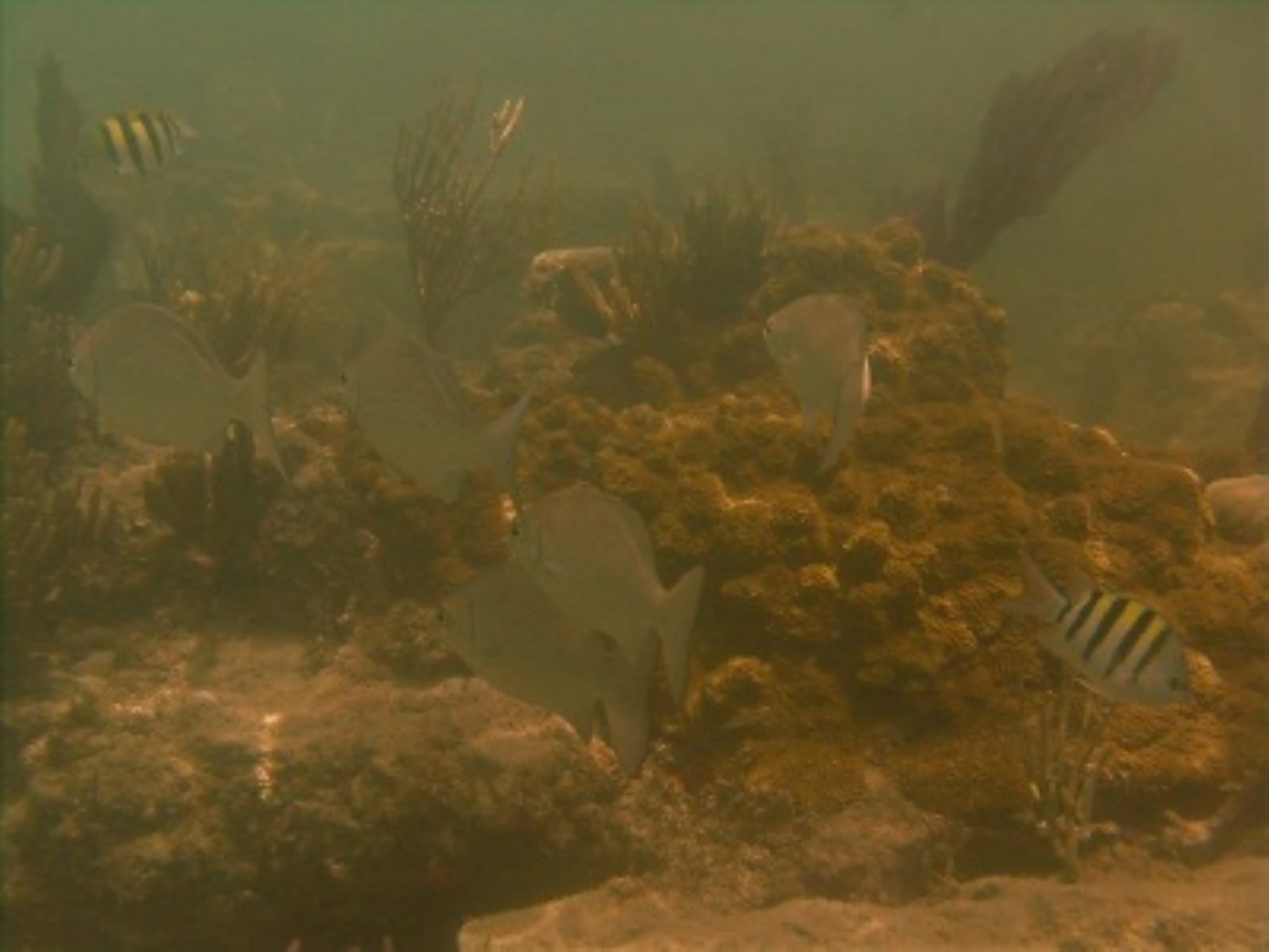}&\includegraphics[height=0.063\textheight]{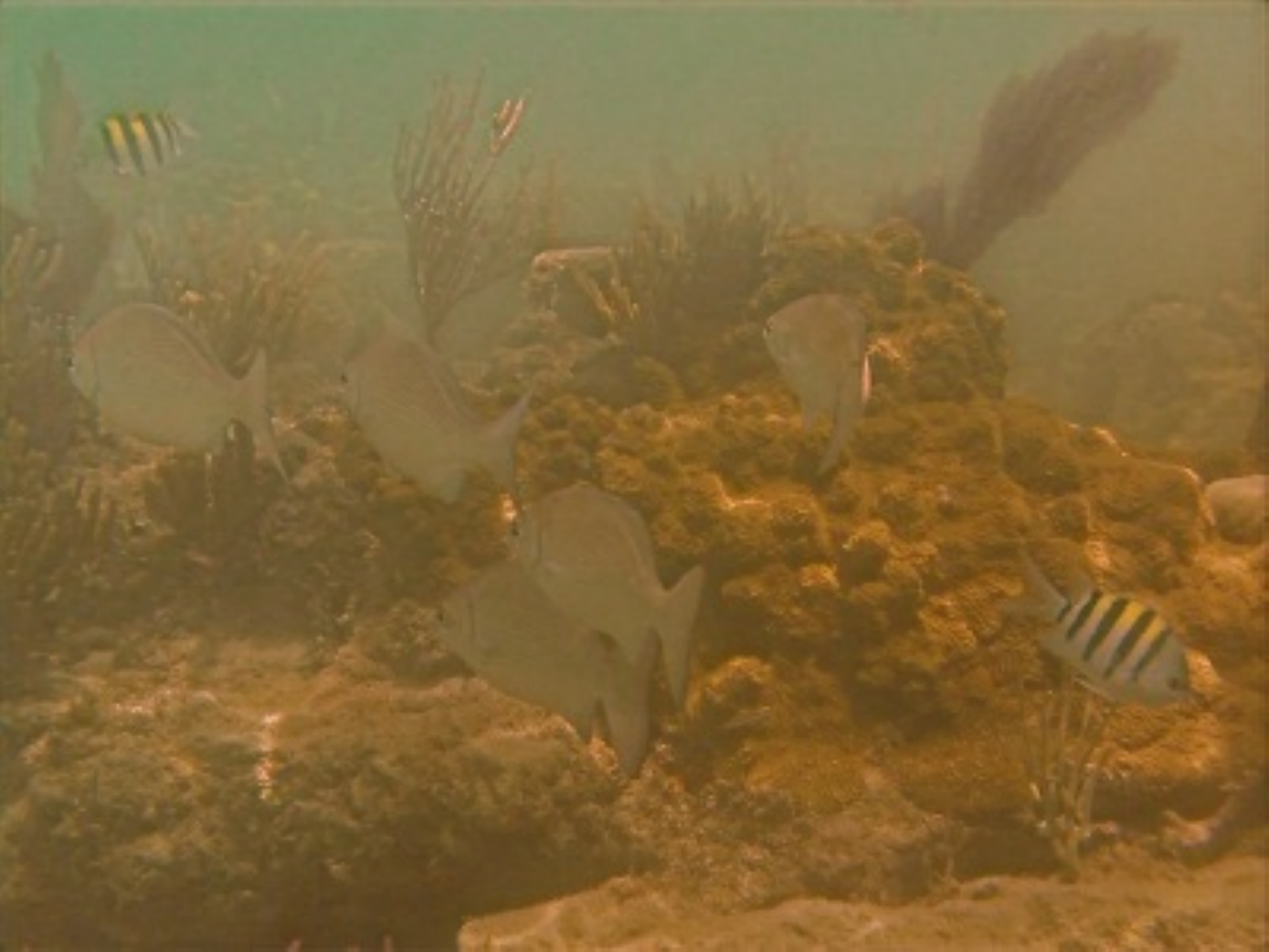}&\includegraphics[height=0.063\textheight]{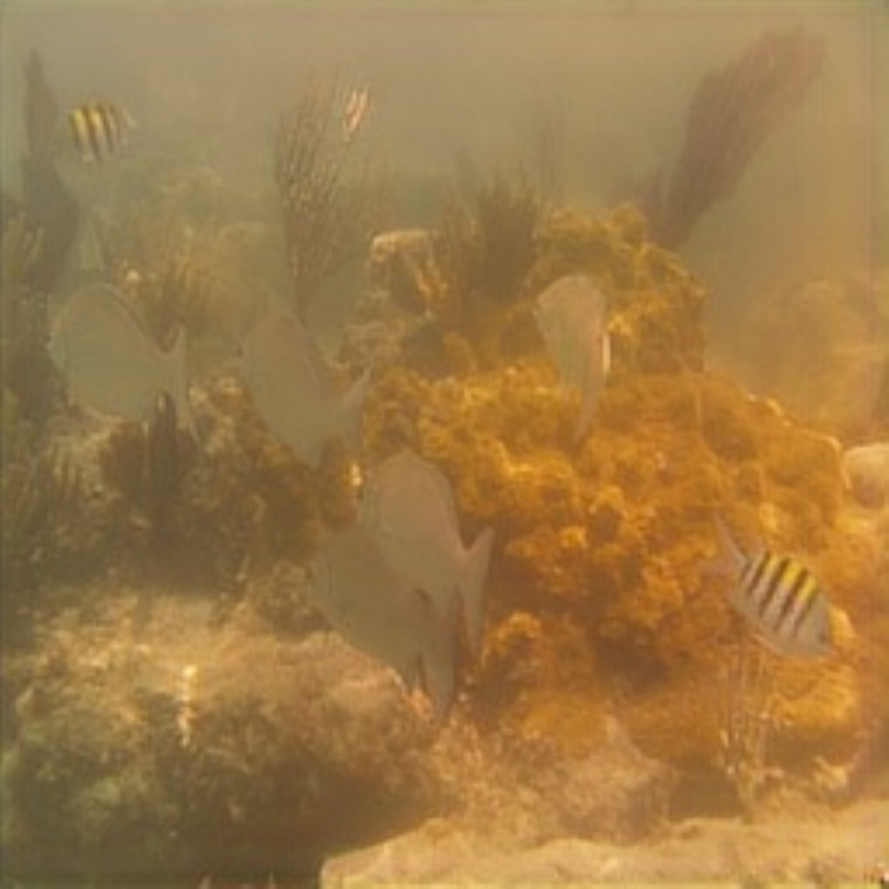}&\includegraphics[height=0.063\textheight]{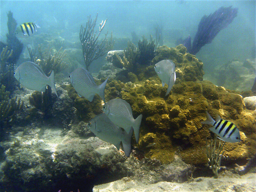} &\includegraphics[height=0.063\textheight]{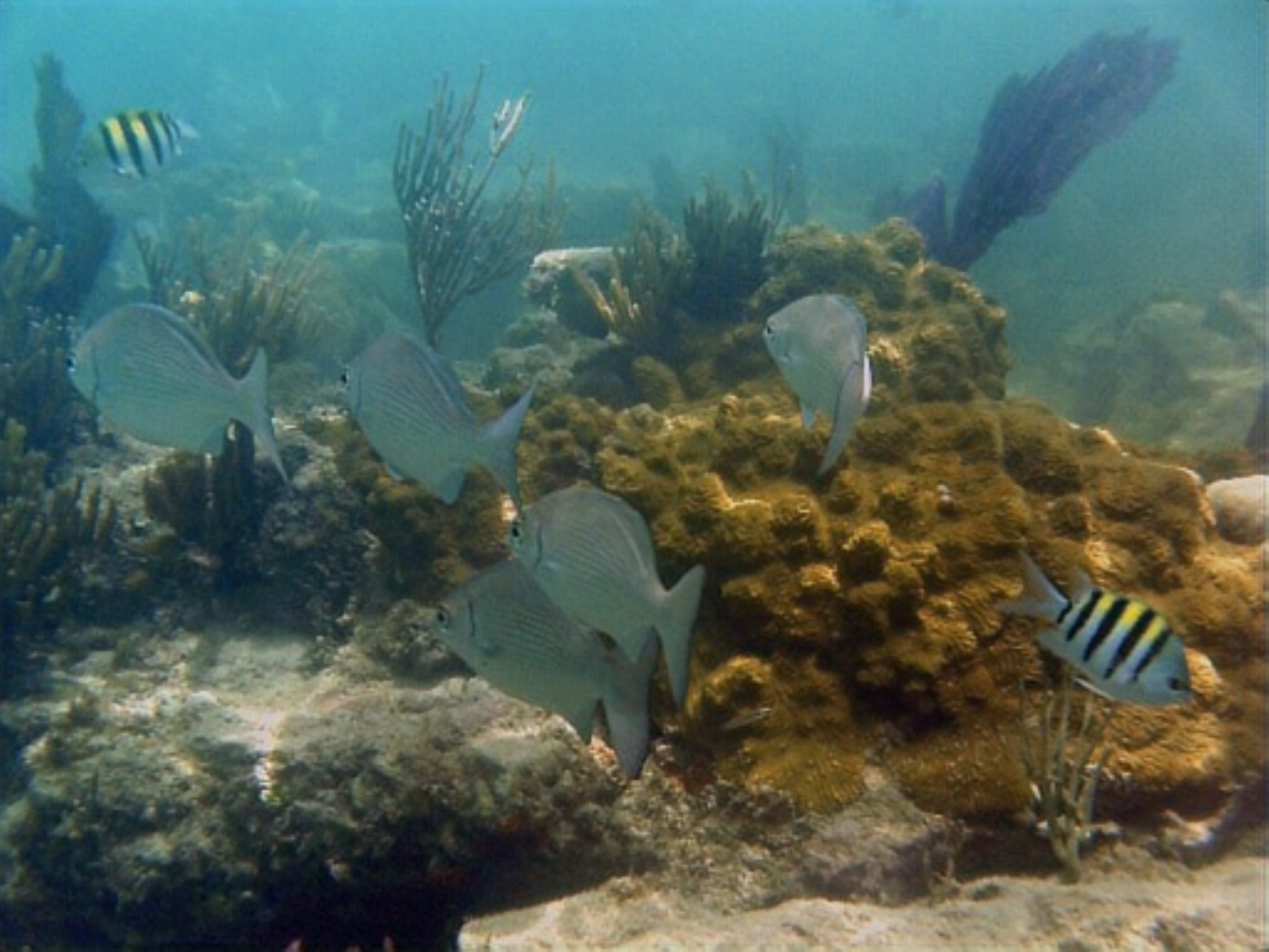}\\
\includegraphics[height=0.060\textheight]{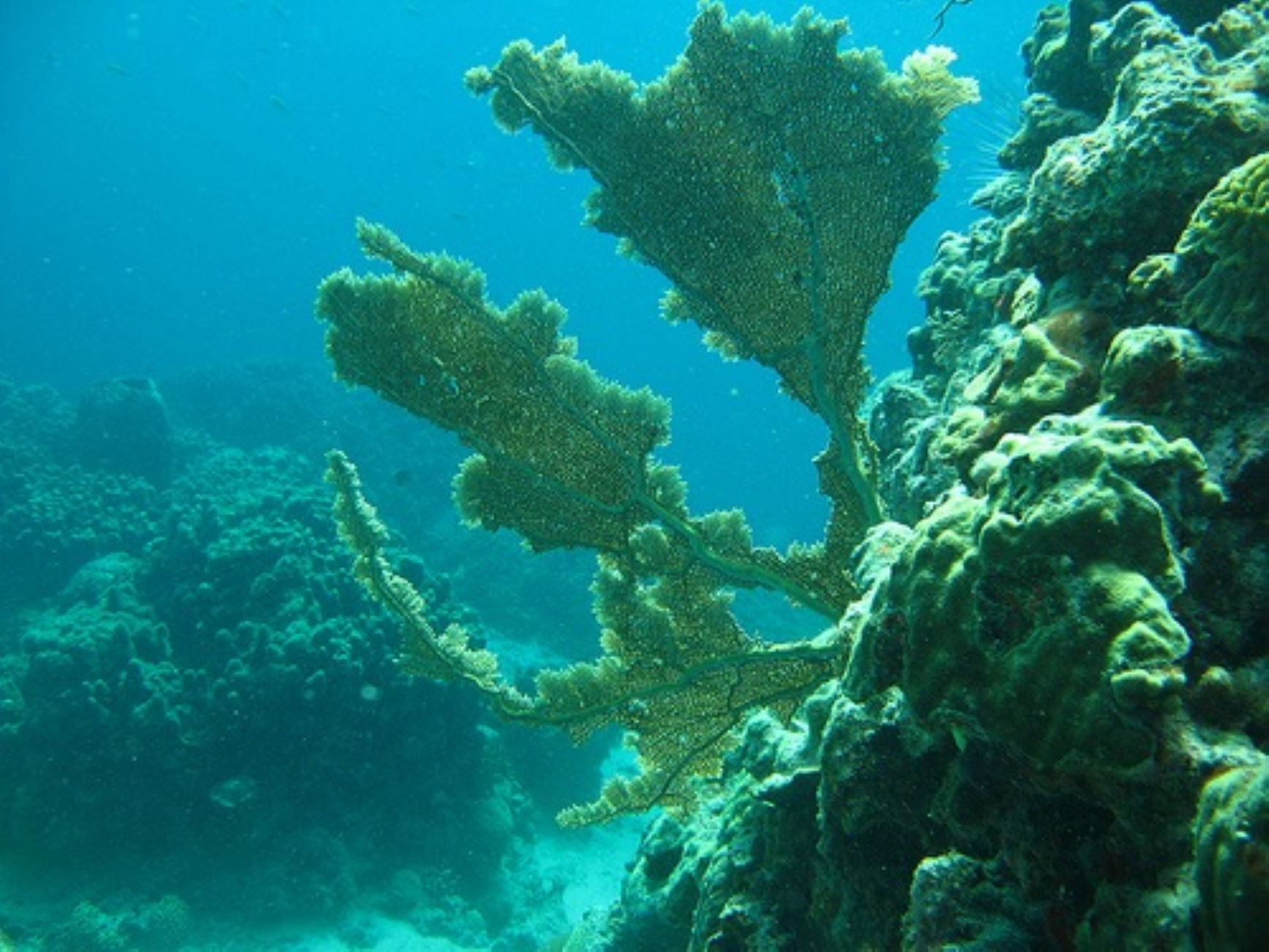}&\includegraphics[height=0.063\textheight]{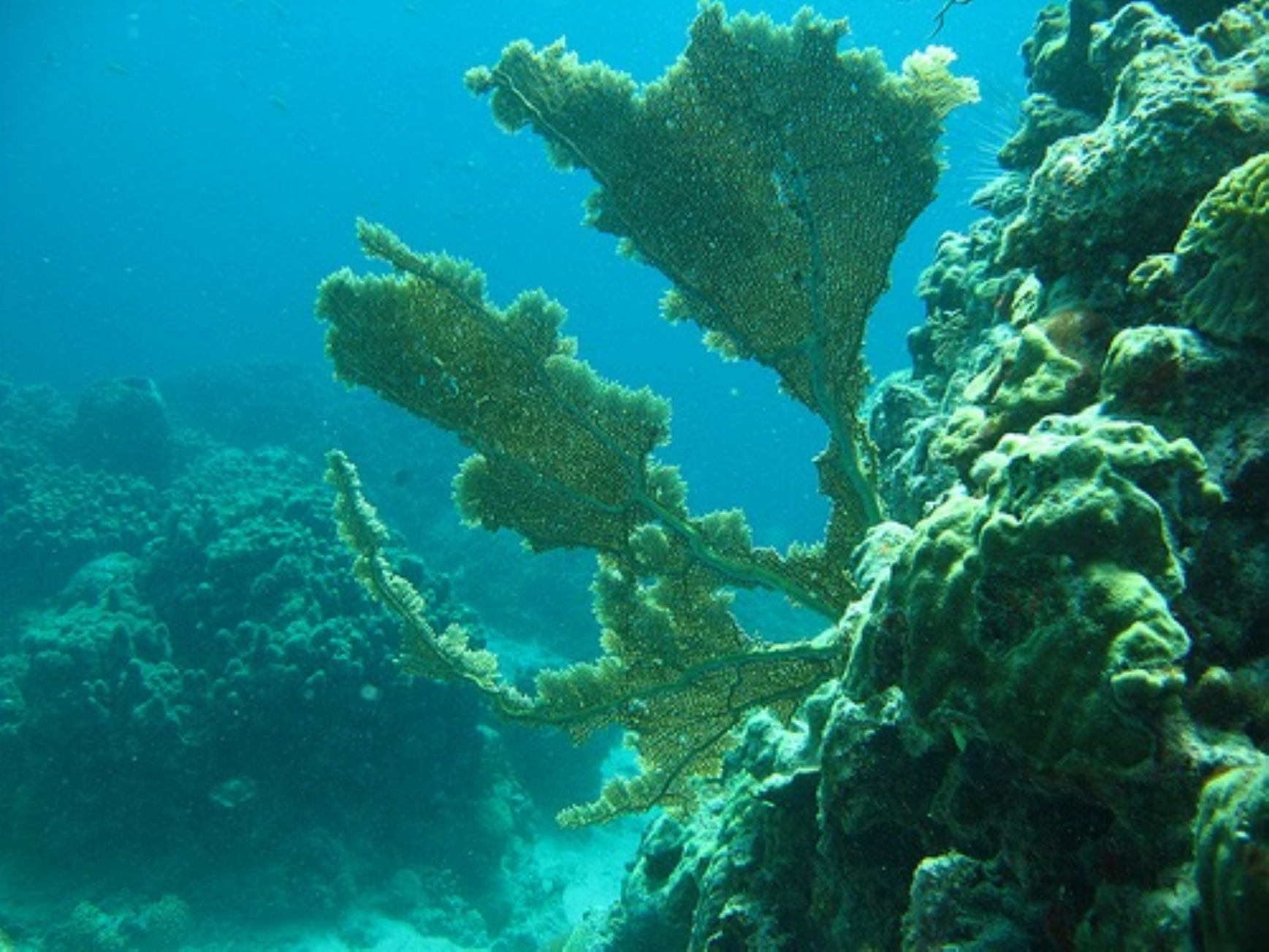}&\includegraphics[height=0.063\textheight]{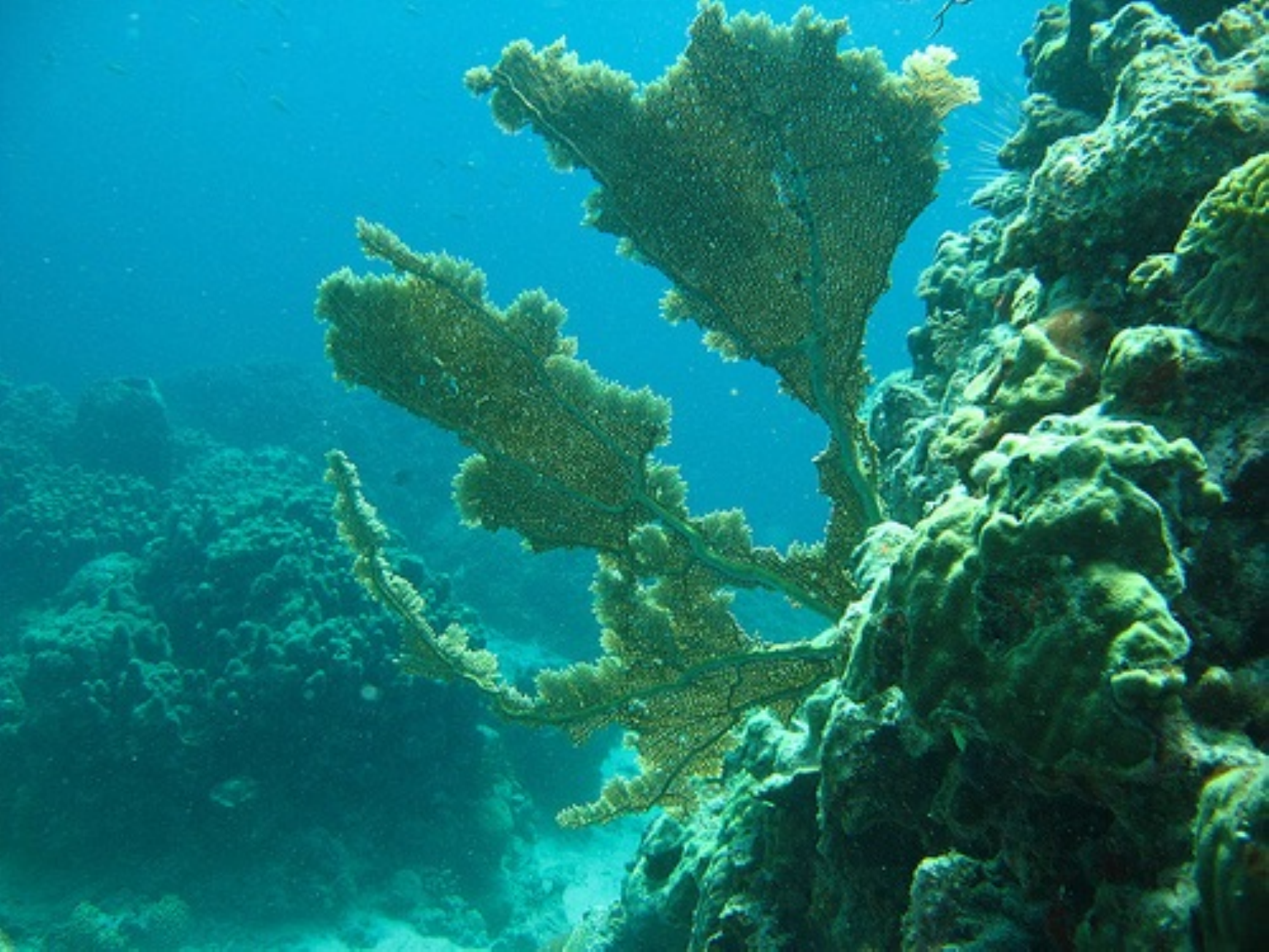}&\includegraphics[height=0.063\textheight]{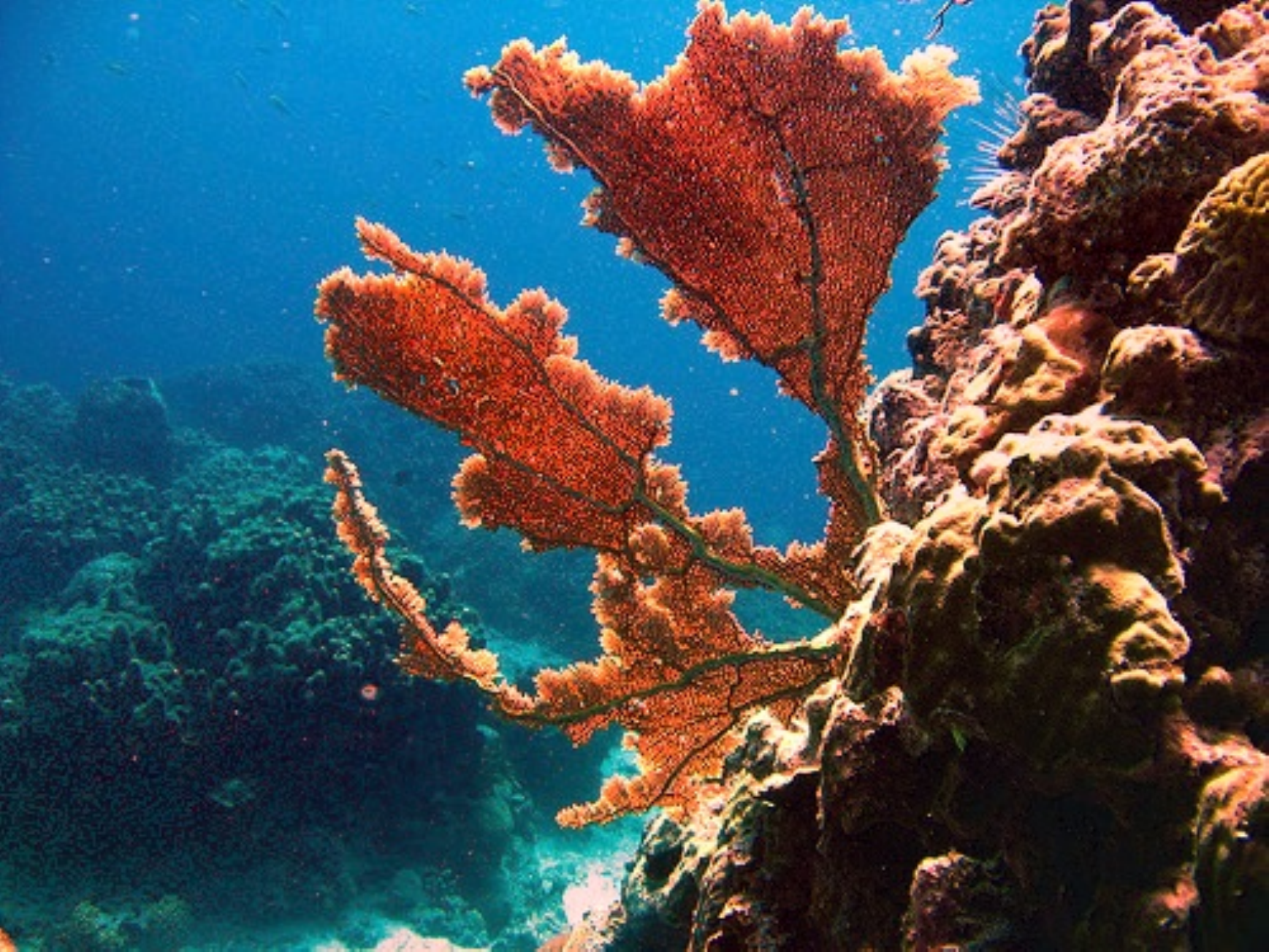}&\includegraphics[height=0.063\textheight]{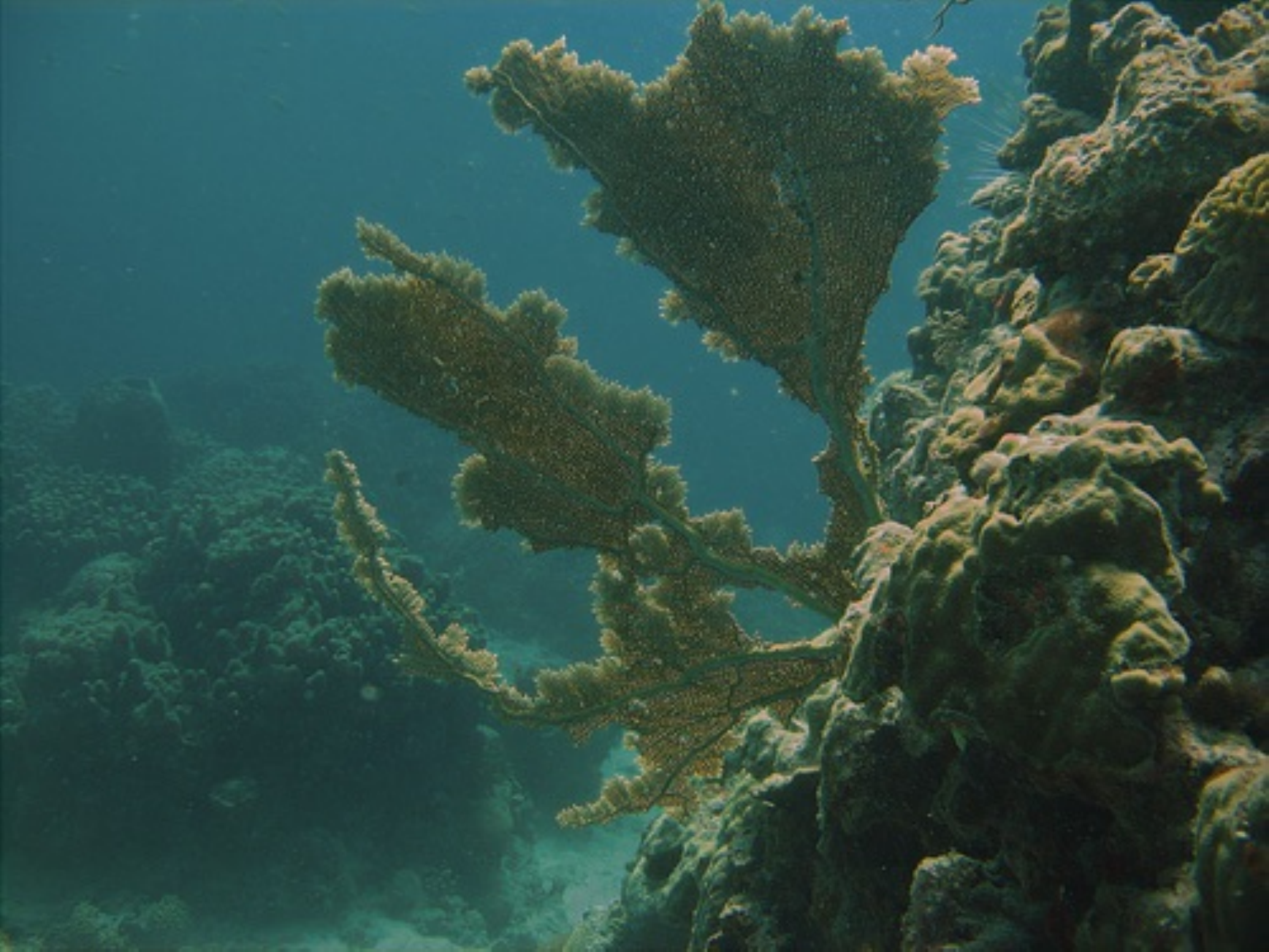}&\includegraphics[height=0.063\textheight]{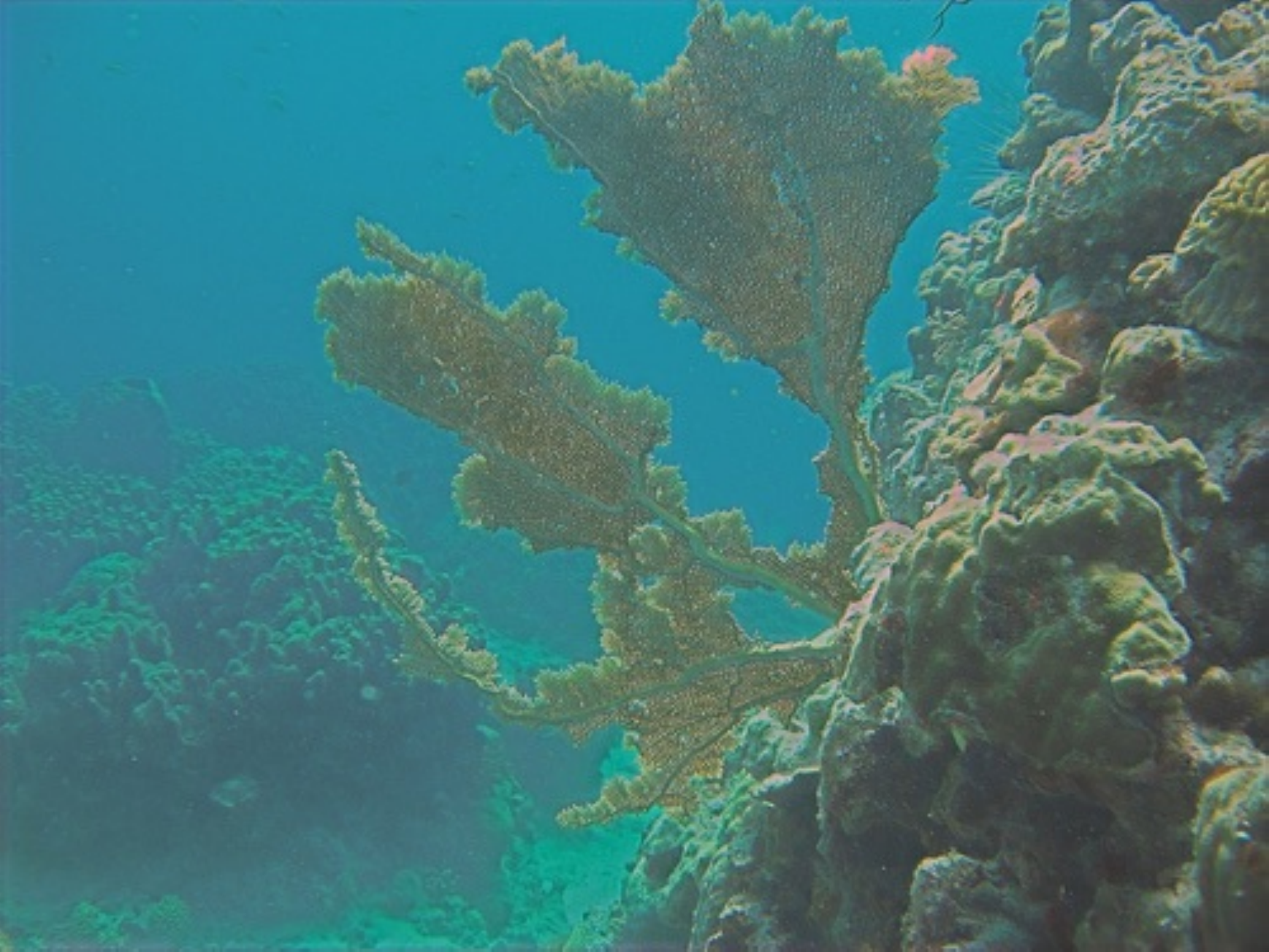}&\includegraphics[height=0.063\textheight]{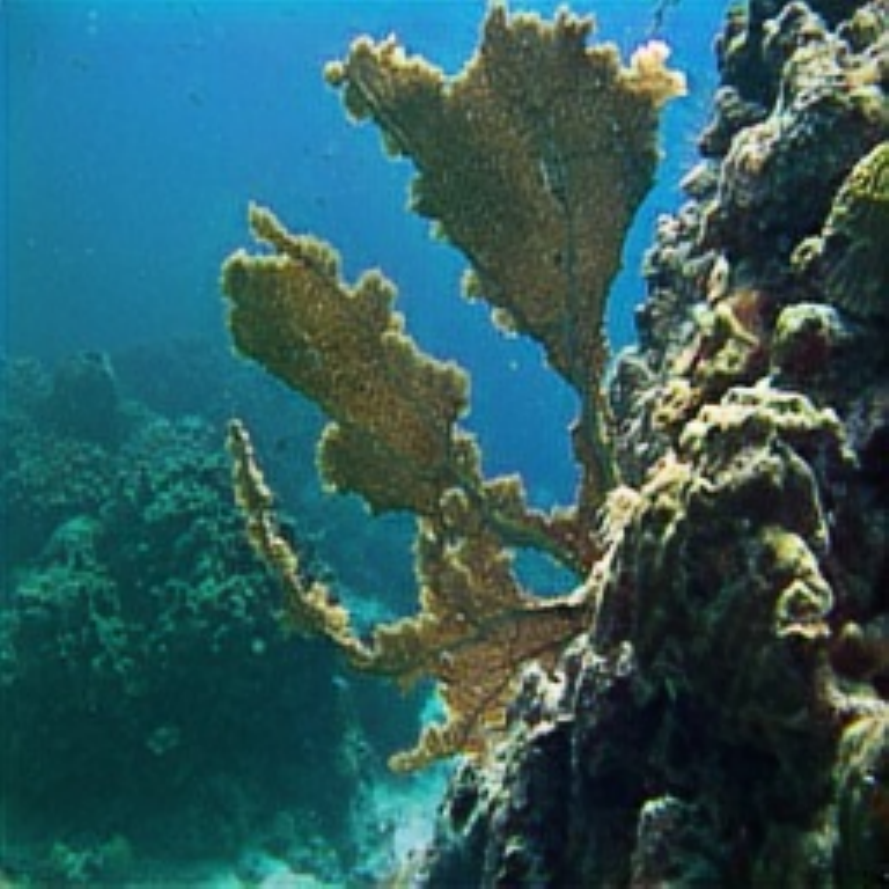}&\includegraphics[height=0.063\textheight]{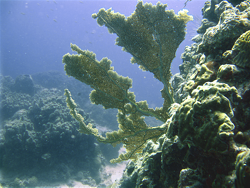}&\includegraphics[height=0.063\textheight]{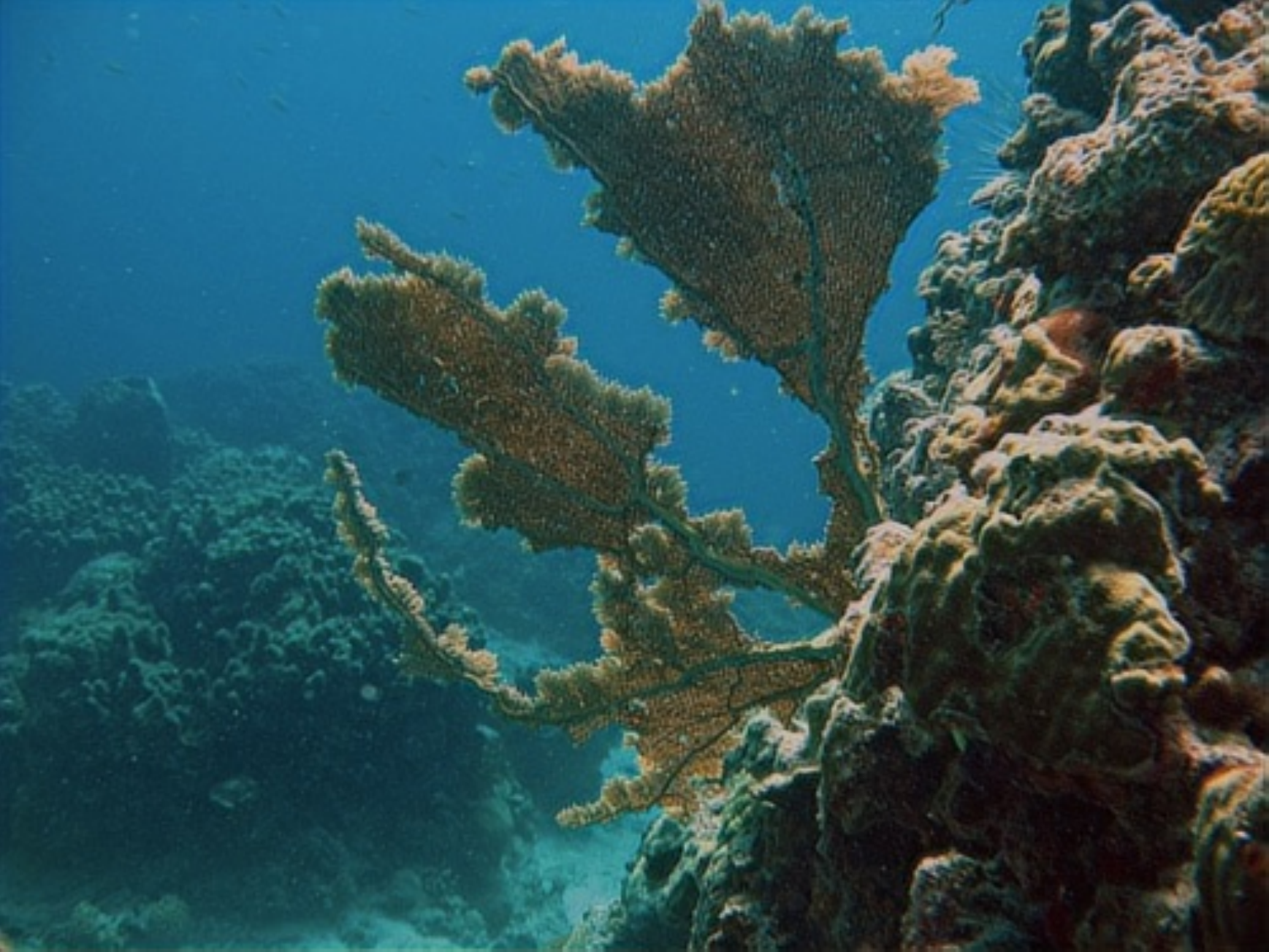} \\
\includegraphics[height=0.060\textheight]{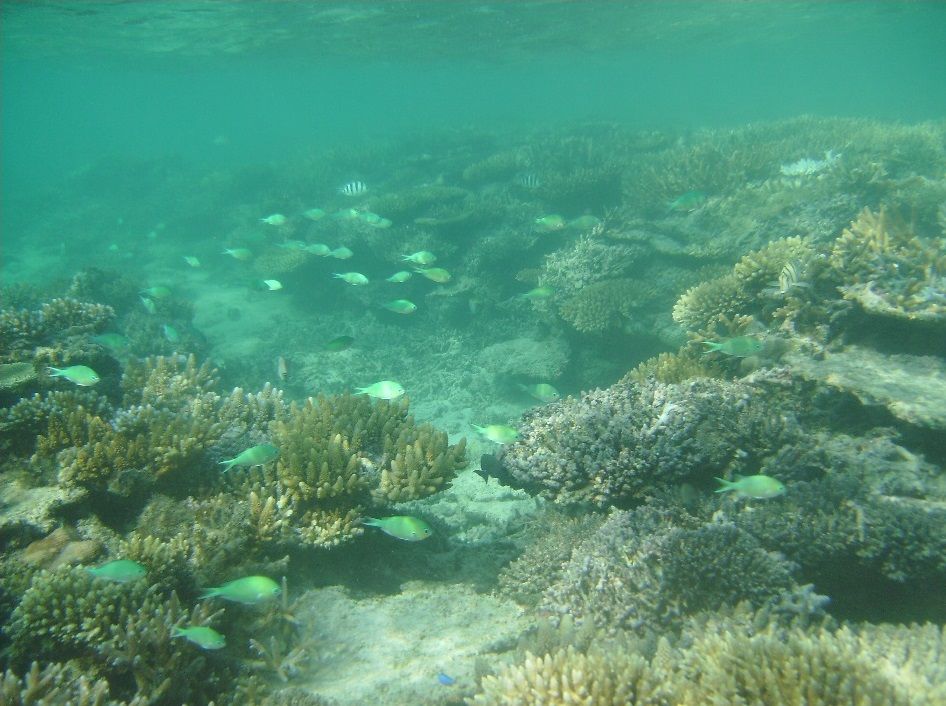}&\includegraphics[height=0.063\textheight]{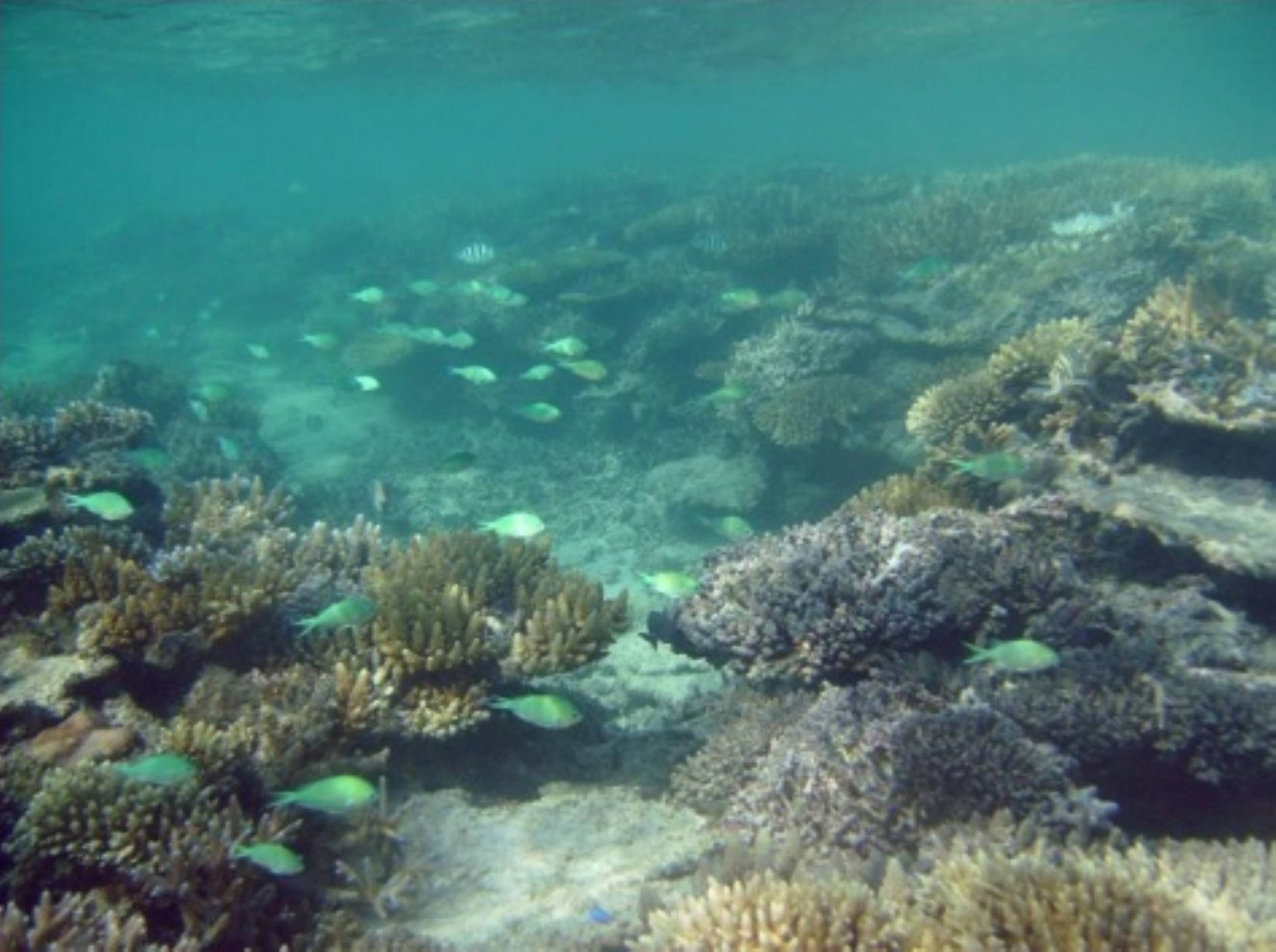}&\includegraphics[height=0.063\textheight]{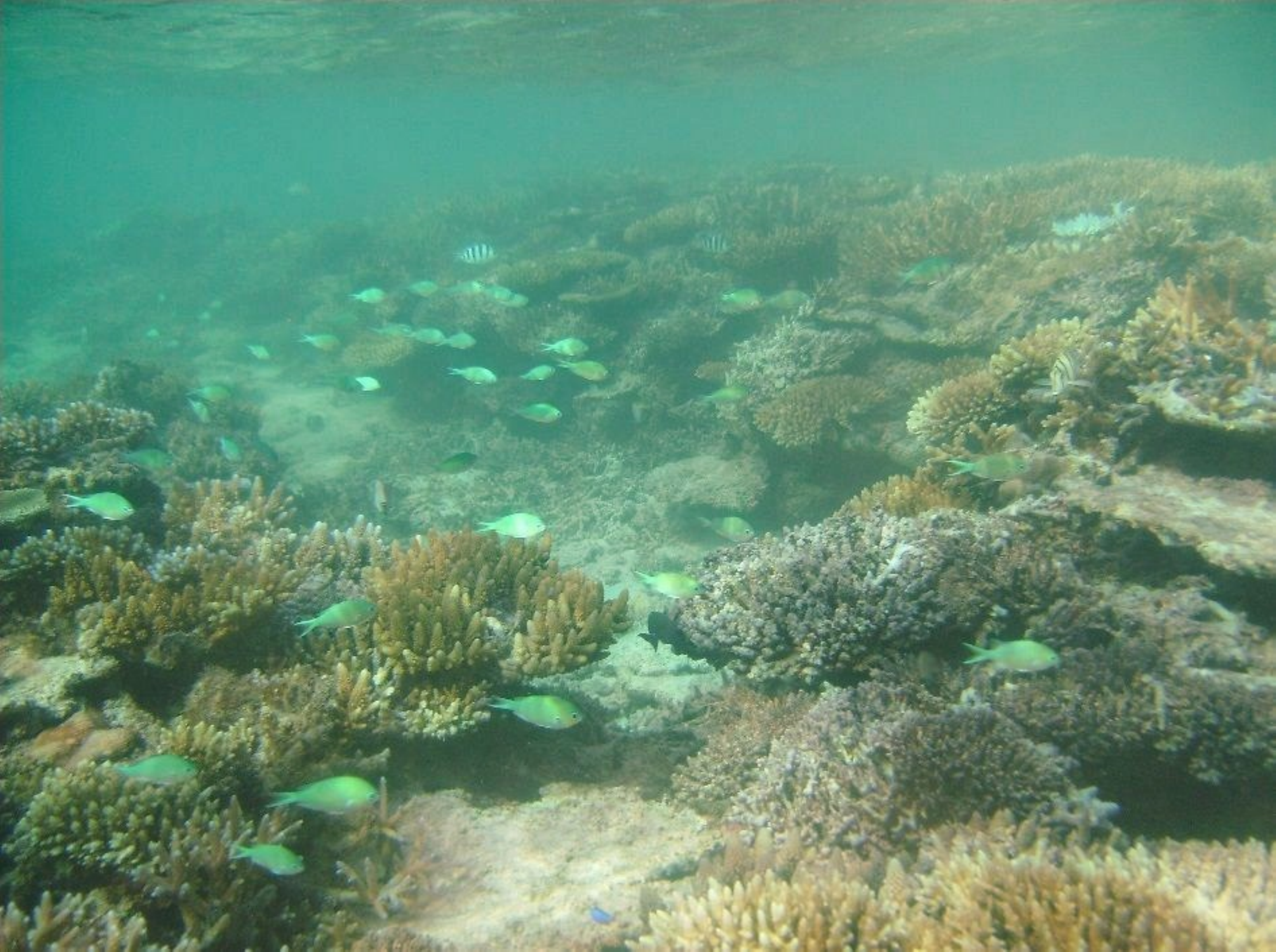}&\includegraphics[height=0.063\textheight]{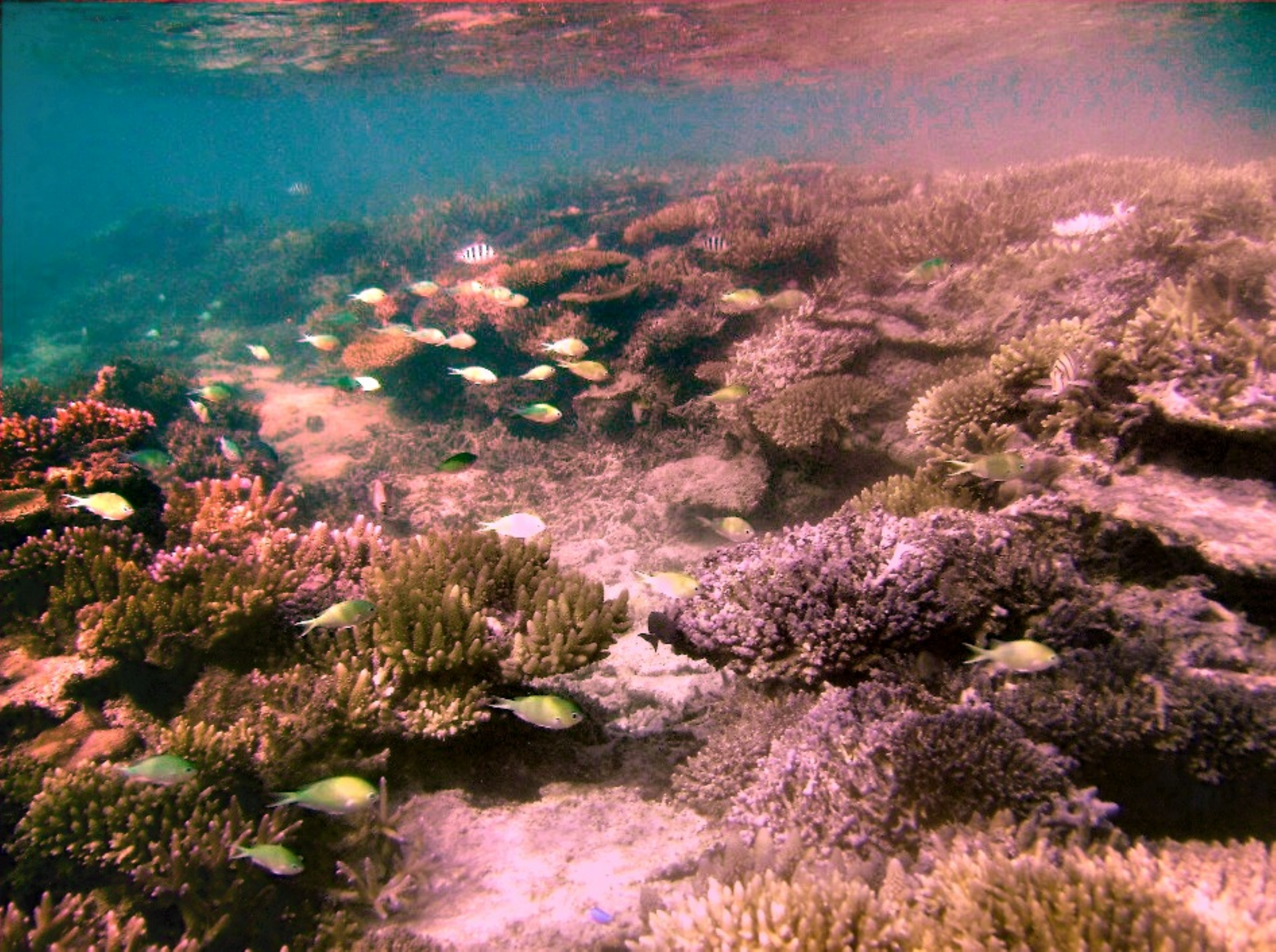}&\includegraphics[height=0.063\textheight]{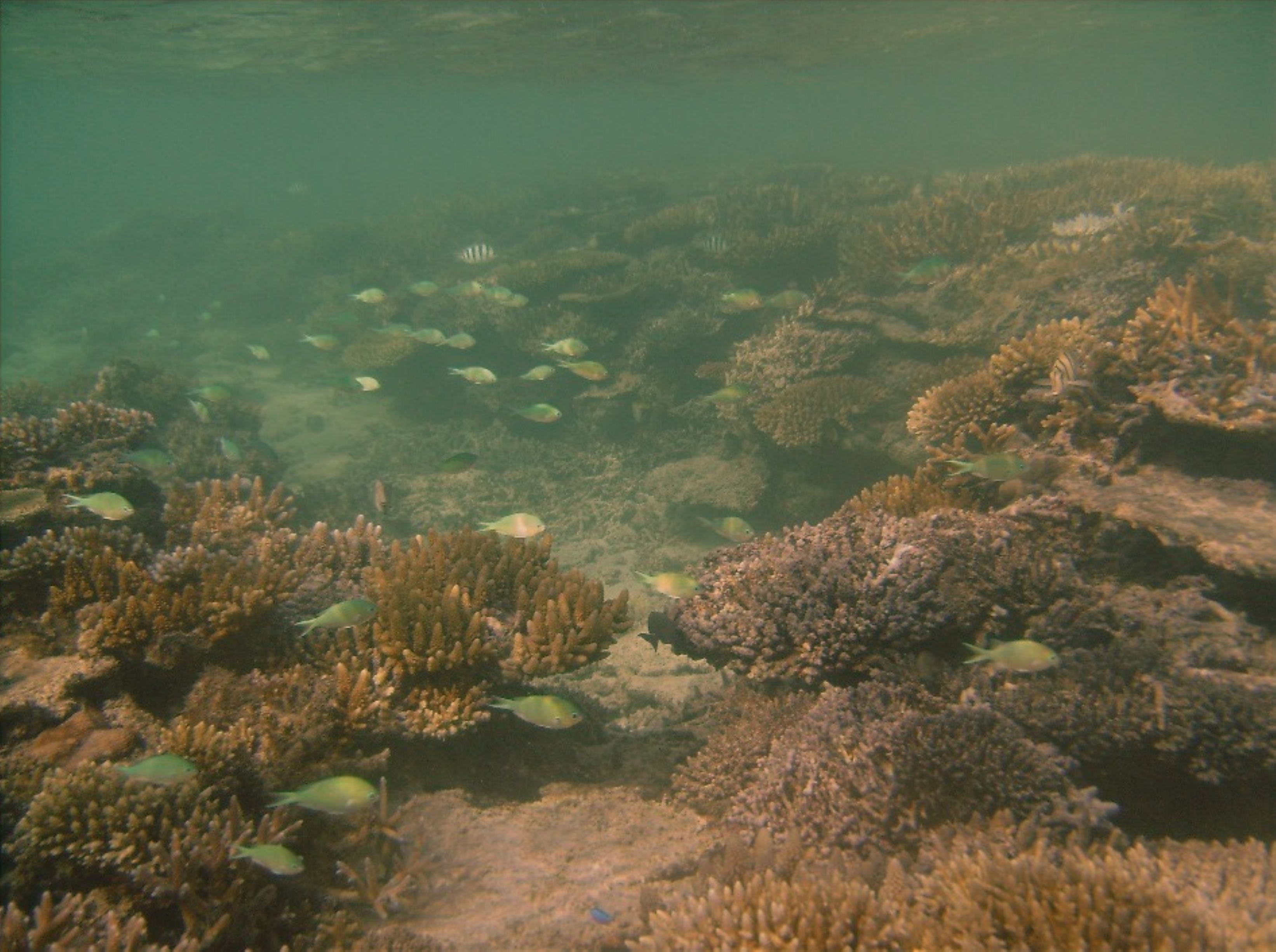}&\includegraphics[height=0.063\textheight]{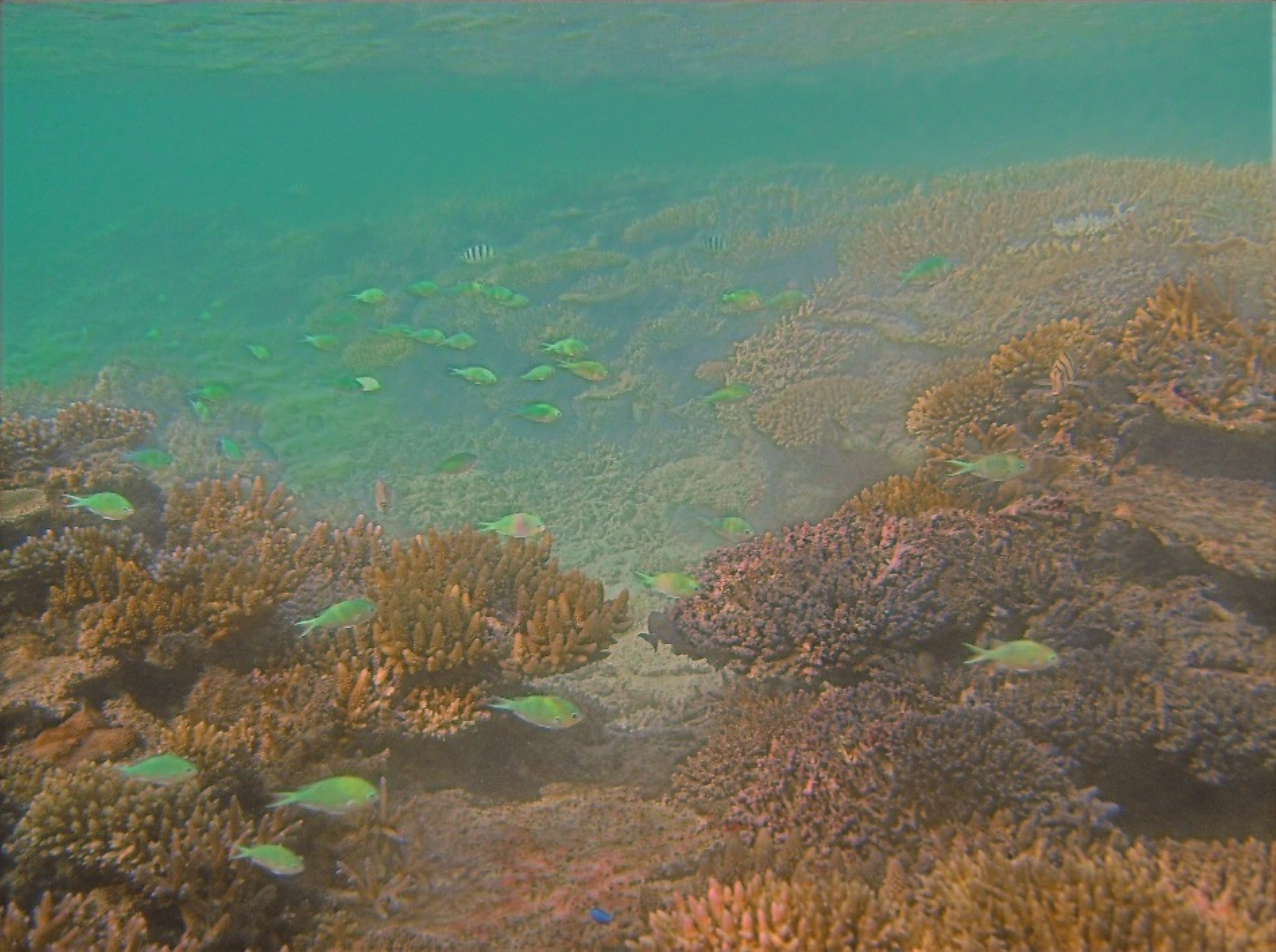}&\includegraphics[height=0.063\textheight]{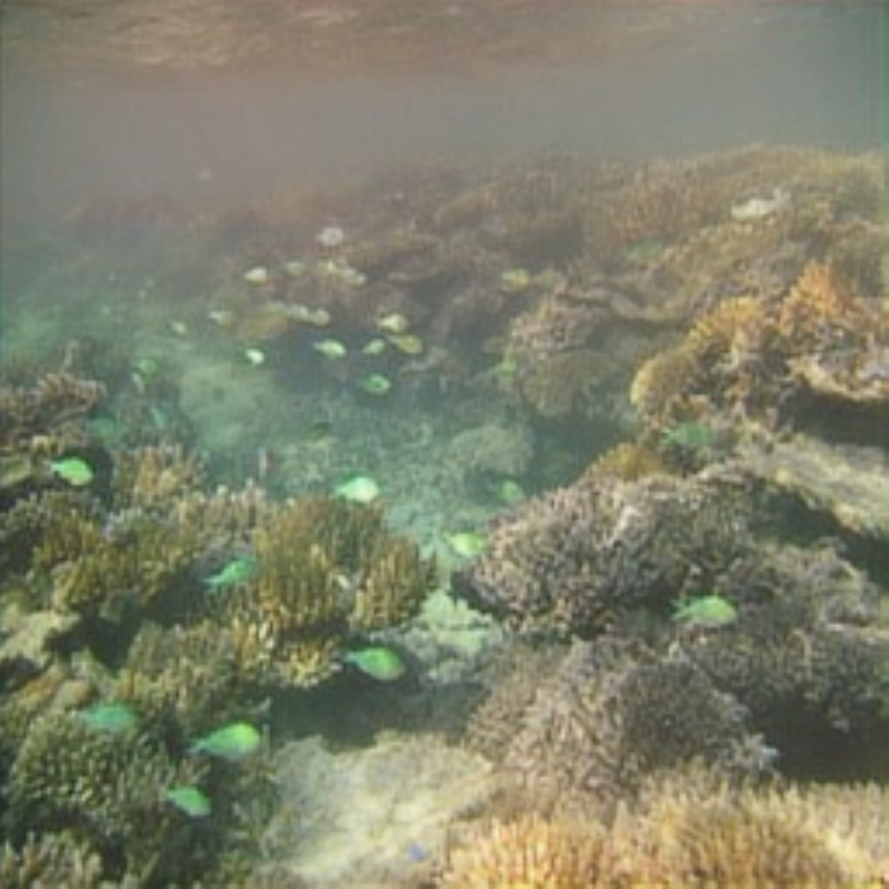}&\includegraphics[height=0.063\textheight]{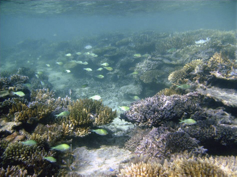}&\includegraphics[height=0.063\textheight]{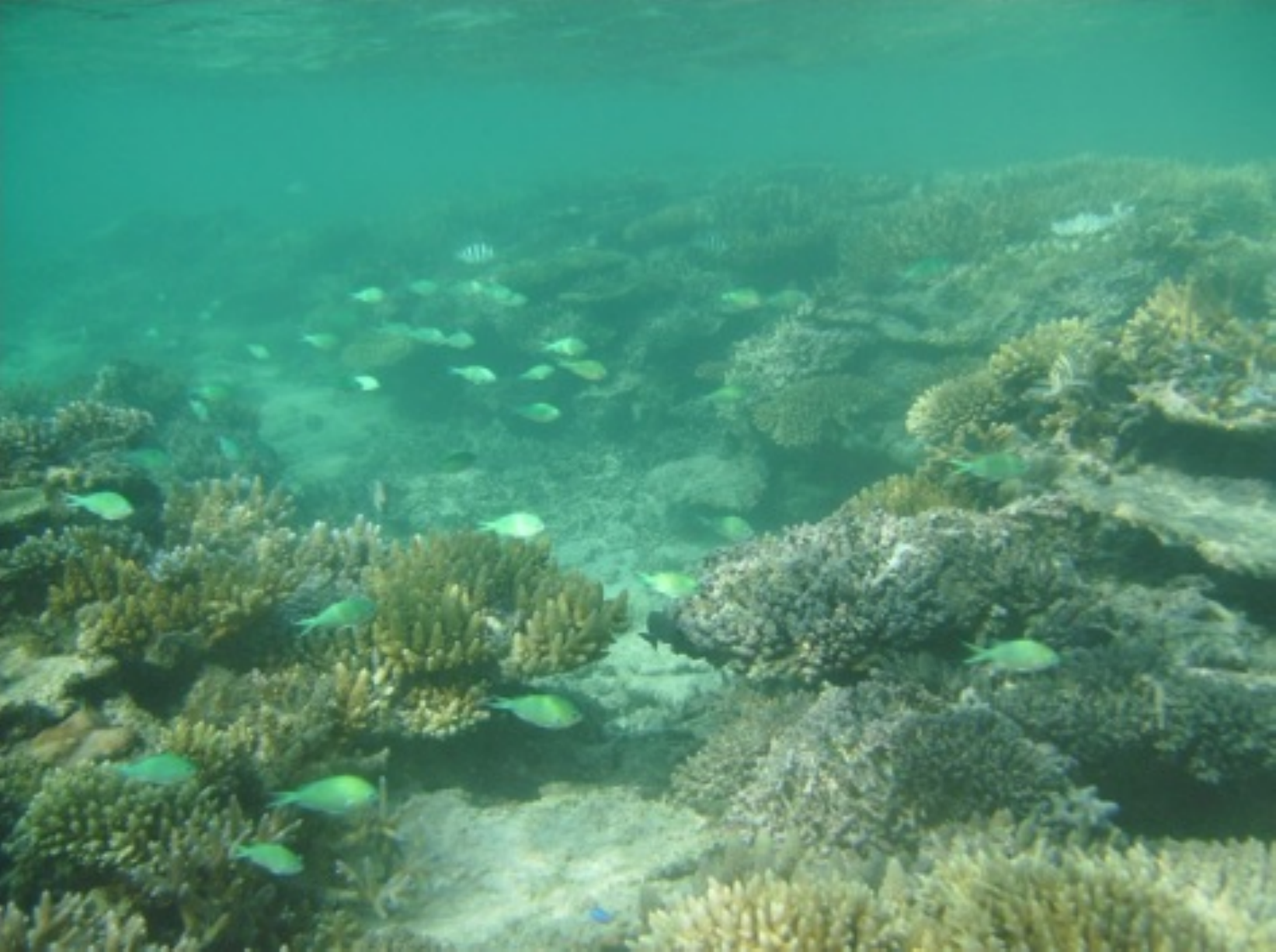} \\
\footnotesize{Original} & \footnotesize{ARC~\cite{ARC}}&\footnotesize{DBL~\cite{BglightEstimationICIP2018_Submitted}}&\footnotesize{UWHL~\cite{BMVC_HL_2017}}&\footnotesize{\mbox{UWCNN-I}~\cite{Anwar2018DeepUI_UWCNN}}&\footnotesize{\mbox{UWCNN-3}~\cite{Anwar2018DeepUI_UWCNN}}&\footnotesize{FUnIE~\cite{Islam2019FastUI4ImprovedPerception}}&\footnotesize{Fusion~\cite{Ancuti2018}}&\footnotesize{WaterNET~\cite{WaterNET_TIP2019}}
\end{tabular}
\caption{Comparison of underwater image filtering results for images that capture the two main Jerlov categories: oceanic waters (rows \mbox{1-4}) and coastal waters (rows \mbox{5-7}). \mbox{UWCNN-I} and \mbox{UWCNN-3} are UWCNN trained for type I oceanic water and for type 3 coastal water, respectively. Note that FUnIE can only process images of fixed size \mbox{(256~$\times$ 256)} and, for visualisation, we resized the image to align with the height of the original image.}
\label{fig:: experiment_natural_images}
\end{figure*}

{\em Generic} IQAs used in the underwater literature measure the 'reconstruction' error or desirable properties of an image. The reconstruction error can be measured by FR IQA measures such as mean square error, PSNR and Structural Similarity~\cite{wang2004ssim}. As reference images are unavailable  (Sec.~\ref{sec:natDatasets}), authors use as  reference the degraded (original)  image~\cite{JTF,Prabhakar2010_wavelet_threshold,Dai2019_uwlowlight,Khan2016_wavelet,Zhang2017_multisclae_retinex,Protasiuk2019_localcolourmapping,  Yu2019_UnderwaterGAN_PRIF, Hashisho_UNet_ISISPA_2019, P2P_Sun_IET_2019} or the filtered version of the degraded image preferred in a subjective test~\cite{WaterNET_TIP2019}.  
Alternatively, desirable properties such as  sharpness, contrast and naturalness can be considered. Sharpness can be measured with a visible-edge count or entropy~\cite{ICIP2017UWCNN, MIL, Khan2016_wavelet, VASAMSETTI2017_wavelet_variational}, but  noise introduced in the filtering process may also increase the sharpness scores, thus causing  incorrect quality predictions.  
Contrast can be measured by Global Contrast Factor~\cite{Matkovic2005GlobalCF}, Patch-based Contrast Quality Index~\cite{PCQI_Wang2015}, or Visibility Metric, originally proposed for hazy images~\cite{visibility_measure}. The (un)naturalness of an image can be measured with  the Blind/Referenceless Image Spatial Quality Evaluator (BRISQUE)~\cite{BRISQUE}, which quantifies distortions, including  noise and blurring~\cite{Blurriness_and_Light}. These IQAs have not yet been comprehensively validated for underwater images and are thus unsuitable as a standalone quality indicators.

{\em Underwater-specific IQAs} are designed to measure degradations caused by absorption and scattering in water (Table~\ref{Table: evaluation_measures_obj}).  We discuss two FR-IQAs, namely~$Q_u$~\cite{Lu2016_descattering_quality_ICIP} and  PKU-EAQA~\cite{PKU}, and three NR-IQAs, namely CCF~\cite{WANG2018_NR_CFF}, UCIQE~\cite{UCIQE} and UIQM~\cite{UIQM_Panetta2016}. All IQAs except~$Q_u$ derived the combination of their attributes by subjective tests. 

\newcommand\RedCell[1]{%
  {\cellcolor{red!28}{#1}}
}

\newcommand\GreenCell[1]{%
  {\cellcolor{green!28}{#1}}
}

An ideal reference image is required for~$Q_u$, which combines the colour difference in the CIELab colour space~\cite{sharma2005ciede2000} and structural  similarity~\cite{wang2004ssim} with equal importance. However, this measure was not validated by a subjective test. PKU-EAQA~\cite{PKU} is a ranking measure that assesses the relative quality between two images. The ranking is based on colour degradation features (colour motion~\cite{Stricker1995}) and edge details (GIST~\cite{Oliva2001}) using a Support Vector Machine (SVM). The SVM is trained on a dataset consisting of relative ranks between 400 images (from 100 images, each processed by 4 methods) obtained from a subjective test with 30 participants. 

Three underwater NR-IQAs are UCIQE~\cite{UCIQE}, UIQM~\cite{UIQM_Panetta2016}, and CCF~\cite{WANG2018_NR_CFF}. UCIQE~\cite{UCIQE} measures only colour degradations, using statistics in the CIELab colour space~\cite{CIELab}, namely the standard deviation in chroma, average in saturation and difference between extreme values in luminance (defined in~\cite{MeasureColorfulness}). The chroma and saturation statistics only focus on individual pixel intensity values and not on the spatial distribution of intensities. The difference in luminance only concerns extreme pixels that could be easily biased towards distortion. In the subjective test, the participants were asked to score the image from 'Very annoying',  'Annoying',  'Slightly annoying',  'Perceptible but not annoying',  and 'Imperceptible'. UCIQE is a linear combination with weights obtained from a subjective test with 12 participants and 44 underwater images.
UIQM~\cite{UIQM_Panetta2016} accounts for the colour degradation  with measurements of colourfulness using the opponent colour theory~\cite{OpponentColour_Hering}, and blurring degradation with measurements of edge sharpness~\cite{Panetta2011ParameterizedLF} and contrasts~\cite{Panetta2014EME}. UIQM is a linear combination with weights obtained from a subjective test with 10 participants and 126 images (14 different images processed by 8 methods and the original). However, the question asked in the subjective test is not reported in the publication.
CCF~\cite{WANG2018_NR_CFF} measures the colour degradation with a colourfulness index~\cite{MeasureColorfulness},  and blurring and lack of visibility with a contrast measure~\cite{Ferzli_Sharpness_JNBlur_TIP2009} and a foggy index~\cite{NRFogginessDensity}. In the subjective test, 20  participants evaluated images of a colour chart captured at different distances in a water tank with different turbidity caused by adding aluminium hydroxide. The participants were asked to score from 1 to 5, which corresponds to 'Bad', 'Poor', 'Fair', 'Good' and 'Excellent'. The main limitation of this measure is that evaluating images taken in chemically created water environment may not fully reflect that of natural underwater images. Appendix details the definitions of the aforementioned NR-IQAs. 

Although underwater-specific IQAs are straightforward and simple to use, they may fail in evaluating certain types of scenes~\cite{Emberton2018}. Furthermore, existing IQAs are not directly comparable with each other as different collections of images were employed in their validation. To  illustrate these limitations, Table~\ref{Table:: IQAs_natural_images} reports the IQA measures on the images in Fig.~\ref{fig:: experiment_natural_images}. Since the three NR-IQA measures have different ranges, we use as the original images' measured quality as a reference: a  measure value higher than the original image (green shade) indicates the filtering method has improved the image quality, and vice versa (red shade). Out of the 56 filtered images, in only 35 instances (62.5\%) that the three measures are in agreement (all consider the image quality has increased or decreased). Furthermore, the IQAs may not capture obvious distortions. An example is the images processed by UWHL~\cite{BMVC_HL_2017}  (Fig.~\ref{fig:: experiment_natural_images} row 3), in which an unnatural red colour is introduced to the water region. However, all three measures indicate improvements in image quality, with UIQM and UCIQE indicating the processed image has the best quality among all the processed images. The lack of comparison between measures, and hence a standard quantitative evaluation scheme, is still an obstacle for using these measures to help improve (and compare) filtering methods.

\begin{table}[t!]
\caption{Comparison of UIQM~\cite{UIQM_Panetta2016}, UCIQE~\cite{UCIQE} and CCF~\cite{WANG2018_NR_CFF} measures for the images in Fig.~\ref{fig:: experiment_natural_images}. All measures state that a higher value indicates a better image quality. Green shade indicates a higher value (better quality) than the original image (Orig.), whereas red shade indicates worse quality. KEYS: r$k$ -- image in the $k^{\text{th}}$ row. }
\centering
\scriptsize
\setlength\tabcolsep{0.1pt}
\begin{tabular}{c}
\begin{tabular}{L{1.4cm}|C{0.78cm}|R{0.7cm}R{0.05cm}|R{0.6cm}R{0.7cm}R{0.7cm}R{0.7cm}R{0.7cm}R{0.7cm}R{0.7cm}R{0.7cm}}
\Xhline{3\arrayrulewidth}
&  Image &  Orig. & & {\cite{ARC} }& {\cite{BglightEstimationICIP2018_Submitted} }& {\cite{BMVC_HL_2017}}& {\cite{Anwar2018DeepUI_UWCNN}-I }& {\cite{Anwar2018DeepUI_UWCNN}-3}& {\cite{Islam2019FastUI4ImprovedPerception} }& {\cite{Ancuti2018} }&  {\cite{WaterNET_TIP2019}} \\
\hline
\multirow{7}{*}{\rotatebox{0}{\textbf{UIQM~\cite{UIQM_Panetta2016}}}} 
  & r1 & 0.58 & & \GreenCell{0.66} & \GreenCell{0.60} & \GreenCell{0.73} & \RedCell{0.55} & \GreenCell{0.62} & \GreenCell{0.70} & \GreenCell{0.75} & \GreenCell{0.75} \\
 & r2 & 0.45 & & \RedCell{0.38} & \RedCell{0.42} & \GreenCell{0.55} & \RedCell{0.34} & \RedCell{0.42} & \GreenCell{0.50} & \RedCell{0.41} & \RedCell{0.39} \\
 & r3 & 0.40 && \GreenCell{0.54} & \GreenCell{0.43} & \GreenCell{0.71} & \GreenCell{0.42} & \GreenCell{0.41} & \GreenCell{0.61} & \GreenCell{0.57} & \GreenCell{0.53} \\
 & r4 & 0.63 & & \GreenCell{0.69} & \GreenCell{0.71} & \GreenCell{0.82} & \GreenCell{0.64} & \GreenCell{0.68} & \GreenCell{0.79} & \GreenCell{0.78} & \GreenCell{0.81} \\
 & r5 & 0.40 && \GreenCell{0.65} & \GreenCell{0.44} & \GreenCell{0.76} & \GreenCell{0.42} & \GreenCell{0.42} & \GreenCell{0.43} & \GreenCell{0.62} & \GreenCell{0.73} \\
 & r6 & 0.89 && \GreenCell{0.89} & \GreenCell{0.90} & \GreenCell{1.18} & \RedCell{0.69} & \RedCell{0.66} & \RedCell{0.84} & \RedCell{0.77} & \RedCell{0.79} \\
 & r7 & 0.56 && \GreenCell{0.65} & \GreenCell{0.58} & \GreenCell{0.98} & {\GreenCell{0.57}} & \GreenCell{0.63} & \GreenCell{0.60} & \GreenCell{0.69} & \GreenCell{0.70}
\\
\hline
\multirow{7}{*}{\rotatebox{0}{\textbf{UCIQE~\cite{UCIQE}}}}& r1 & 0.54& & \GreenCell{0.58} & \GreenCell{0.56} & \GreenCell{0.67} & \RedCell{0.50} & \GreenCell{0.54} & \RedCell{0.53} & \GreenCell{0.65} & \GreenCell{0.64} \\
 & r2 & 0.53 && \GreenCell{0.55} & \GreenCell{0.56} & \GreenCell{0.61} & \RedCell{0.46} & \RedCell{0.46} & \RedCell{0.52} & \GreenCell{0.56} & \GreenCell{0.55} \\
 & r3 & 0.50 & & \GreenCell{0.58} & \RedCell{0.50} & \GreenCell{0.68} & \RedCell{0.49} & \RedCell{0.45} & \GreenCell{0.52} & \GreenCell{0.59} & \GreenCell{0.59} \\
 & r4 & 0.54 & & \GreenCell{0.57} & \GreenCell{0.58} & \GreenCell{0.67} & \RedCell{0.54} & \GreenCell{0.55} & \GreenCell{0.56} & \GreenCell{0.64} & \GreenCell{0.63} \\
 & r5 & 0.43 & & \GreenCell{0.53} & \GreenCell{0.45} & \GreenCell{0.67} & \RedCell{0.43} & \RedCell{0.43} & \GreenCell{0.46} & \GreenCell{0.62} & \GreenCell{0.60} \\
 & r6 & 0.61 & & \GreenCell{0.61} & \GreenCell{0.61} & \GreenCell{0.74} & \RedCell{0.53} & \RedCell{0.50} & \GreenCell{0.64} & \RedCell{0.61} & \RedCell{0.60} \\
 & r7 & 0.50 & & \GreenCell{0.56} & \GreenCell{0.51} & \GreenCell{0.69} & \RedCell{0.49} & \RedCell{0.50} & \RedCell{0.48} & \GreenCell{0.59} & \GreenCell{0.58}\\
\hline 
\multirow{7}{*}{\rotatebox{0}{\textbf{CCF~\cite{WANG2018_NR_CFF}}}}  & r1 & 15.92 & & \GreenCell{17.63} & \GreenCell{17.66} & \GreenCell{36.84} & \RedCell{12.69} & \RedCell{12.12} & \GreenCell{18.78} & \GreenCell{23.64} & \GreenCell{23.95} \\
 & r2 & 11.44 && \GreenCell{12.35} & \GreenCell{12.97} & \GreenCell{15.71} & \RedCell{9.26} & \RedCell{9.72} & \GreenCell{17.09} & \GreenCell{13.95} & \GreenCell{12.79} \\
 & r3 & 13.12 & & \GreenCell{14.85} & \GreenCell{28.70} & \GreenCell{39.89} & \RedCell{11.34} & \RedCell{9.40} & \GreenCell{19.92} & \GreenCell{16.22} & \GreenCell{14.62} \\
 & r4 & 22.18 && \GreenCell{23.37} & \GreenCell{23.64} & \GreenCell{45.98} & \RedCell{14.18} & \RedCell{13.04} & \RedCell{20.54} & \RedCell{21.65} & \GreenCell{54.19} \\
 & r5 & 10.61 && \GreenCell{13.98} & \GreenCell{11.76} & \GreenCell{34.56} & \RedCell{9.22} & \RedCell{9.65} & \GreenCell{13.55} & \GreenCell{17.26} & \GreenCell{18.69} \\
 & r6 & 46.23 && \GreenCell{46.33} & \GreenCell{49.13} & \RedCell{45.98} & \RedCell{14.28} & \RedCell{11.44} & \RedCell{24.73} & \RedCell{24.77} & \RedCell{17.86} \\
 & r7 & 14.66 & & \GreenCell{16.54} & \GreenCell{49.13} & \GreenCell{54.47} & \RedCell{12.44} & \RedCell{10.72} & \GreenCell{18.25} & \GreenCell{20.34} & \GreenCell{17.71}
\\
\Xhline{3\arrayrulewidth}
\end{tabular}%
\end{tabular}
\label{Table:: IQAs_natural_images}
\end{table}

\section{Conclusion and outlook}
\label{sec: conclusion}

We presented an extensive survey on underwater restoration and enhancement methods. We categorised the key design principles and discussed which methods are suited to different environments. We also covered datasets and loss functions employed in training neural networks and discussed their limitations. Moreover, we categorised and described  underwater datasets based on their source of degradation. Finally, we  discussed evaluation strategies, including subjective tests, and presented the limitations of popular image quality assessment measures. To conclude the survey, we highlight open research issues. 

The success of deep neural networks in many computer vision tasks has not yet been replicated for underwater image filtering. 
Accurately modelled degraded images should be synthesised for training, accounting for oceanology models (see Section~\ref{sec:Preliminaries}). While existing methods mainly use the Schechner-Karpel model~\cite{DBLP:conf/cvpr/SchechnerK04}, a more sophisticated model, such as the one proposed by Akkaynak-Treibitz~\cite{RevisedModel_CVPR2018}, could be used to accurately account for the attenuation along the range and incorporate camera responses. Moreover, the attenuated illuminant and scattering should be incorporated, as done in~\cite{Li_2020_Cast-GAN}.  

Existing restoration methods focus mainly on compensating the colour degradation along the distance between an object and the camera~\cite{Schechner2005_polarised,Treibitz2009_descattering,Drews2013UDCP,Emberton2015,CB2010InitialResult,ARC,Blurriness_and_Light,BglightEstimationICIP2018_Submitted,MIL,Hou2020_DCPguided_TV, pmlr-v95-hu18a_restoration_CNN, Shin2016CNN,ICIP2017UWCNN}. A desirable research direction is removing the colour cast caused by attenuation through the depth under the water surface and considering the water colour when the water mass is visible. Other directions include compensating for the blurring and restoring  images captured with artificial light. 

Most enhancement methods apply the filtering on the whole image~\cite{Fu_retinex-based_ICIP2014,Tang_retinex_video_SOVP2019,Gao2019_Adaptive_Retinex,Sanchez-Ferreira_2019_BioInspired_ABC,Zhang2017_multisclae_retinex,Protasiuk2019_localcolourmapping,Iqbal2010UCC,JTF,Prabhakar2010_wavelet_threshold,Dai2019_uwlowlight}. However, the enhancement of captured objects and pure water mass should differ. Image fusion focuses on combining the processed versions by properties of the image (saliency or edges)~\cite{Ancuti2012,Ancuti2018,Khan2016_wavelet,VASAMSETTI2017_wavelet_variational, WaterNET_TIP2019}. Regions that are most worthy of enhancement could be more explicitly identified using  saliency~\cite{Ancuti2018} or attention mechanisms in learning-based methods~\cite{DeepVisualAttentionPrediction_Wang_2018}. 

To conclude, defining a robust and accurate evaluation is an important open issue in underwater image filtering.  
Datasets that provide colour references (e.g. Sea-thru~\cite{seathru2019CVPR} and SQUID~\cite{BermanArvix_UWHL_dataset}) should be used to validate colour accuracy in restoration methods, whereas subjective tests should be used for enhancement methods, instead of relying completely on IQAs  which have several limitations, as we have shown in  Section~\ref{sec:imgQualAssessment}. A thoroughly validated IQA measure, with a description of its applicable scenarios and limitations, would contribute to a standardised evaluation approach for underwater images and support progress in this field. To this end, to complement our survey, we provide a companion online platform, PUIQE~\cite{PUIQE_website},  which supports underwater image filtering and evaluation. 


\section*{Appendix: IQA definitions}
\label{Appendix: IQA}
This appendix summarises the definitions of no-reference underwater-specific image quality assessment measures~(IQAs) discussed in Sec.~\ref{sec:imgQualAssessment}, namely UCIQE~\cite{UCIQE}, UIQM~\cite{UIQM_Panetta2016} and  CCF~\cite{WANG2018_NR_CFF}. 

Let an image of height~$H$ and width~$W$ be~$[\textbf{R},\textbf{G},\textbf{B}]$ in RGB colour space and be~$[\textbf{L},\textbf{a},\textbf{b}]$ in CIELab colour space, where~$\textbf{L}$ is the luminance channel. Let \mbox{$N = H \cdot W$} be the total number of pixels in the image. Let~$\textbf{I}'$ represent the grayscale image of~$\textbf{I}$, computed as the average intensity between the R,G and B colour channels. Let~$\textbf{I}'$ be divided into \mbox{$L\cdot M$} blocks and let~$\textbf{W}^{\textbf{I}'}_{l,m}$ represent one of these blocks, \mbox{$l \in \{1,...,L\}, m \in \{1,...,M\}$}. 
Unless otherwise stated, all operations are element-wise. We have modified the notations from the publications to avoid possible confusion among the three measures. 

\newcounter{Cequ}
\newenvironment{CEquation}
{\stepcounter{Cequ}%
\addtocounter{equation}{-1}%
\renewcommand\theequation{A.\arabic{Cequ}}\equation}
{\endequation}

\subsection*{i. UCIQE (Yang and Sowmya~\cite{UCIQE})}

UCIQE measures the colour degradation in the CIELab colour space, assuming the more colourful an image is, the better the image quality. Colourfulness in natural images can be represented by a combination of quantities including statistics of chroma~$\textbf{C}$ and saturation~$\textbf{S}$~\cite{MeasureColorfulness}, defined as
\begin{CEquation}
\begin{alignedat}{4}
\textbf{C} &= \sqrt{\textbf{a}^2 + \textbf{b}^2},\\
\textbf{S} &= \frac{\textbf{C}}{\textbf{L}}.
\end{alignedat}
\end{CEquation}

UCIQE quantifies the colourfulness by the standard deviation in chroma,~$\sigma_\textbf{C}$, the difference between the top 1\% and bottom 1\% of values in~$\textbf{L}$,~$con_l$, and the average of saturation,~$\mu_\textbf{S}$. The weights that combine the attributes are derived from the Mean Opinion Score (MOS) of testing images from the subjective test. The IQA is defined as
\begin{CEquation}
\text{UCIQE} = 0.4680 \cdot \sigma_\textbf{C} + 0.2745 \cdot con_l + 0.2576 \cdot \mu_\textbf{S}.
\end{CEquation} 
UCIQE states that the higher the value, the better the image  quality.


\subsection*{ii. UIQM (Panetta et al.~\cite{UIQM_Panetta2016})}
UIQM is a combination of three attributes: colourfulness (UICM), sharpness (UISM) and contrast (UIConM). 

UICM uses an asymmetric alpha-trimmed statistics~\cite{alpha_trimmed} that avoids the effect of outlier intensities on the measure, using the opponent colour components~\cite{OpponentColour_Hering} 
\begin{CEquation}
\begin{alignedat}{4}
&\textbf{RG} &&= \textbf{R}-\textbf{G},\\ 
&\textbf{YB} &&=  \frac{\textbf{R}+\textbf{G}}{2} - \textbf{B}.
\end{alignedat}
\label{eq:: UIQM_colourfulness_RG}
\end{CEquation}
Let~$\{{\textbf{RG}}_1,{\textbf{RG}}_2,...,{\textbf{RG}}_{N} \}$ be the intensity values in~$\textbf{RG}$ sorted in ascending order. The alpha-trimmed average for~$\textbf{RG}$, $\hat{\mu}_{\textbf{RG}}$, is defined as the average intensity in $\textbf{RG}$ with the top $T_{R}$ and bottom $T_{L}$ values  discarded, as 
\begin{CEquation}
\mu^*_{\textbf{RG}} = \frac{1}{N - T_{L}-T_{R}} \sum_{k=T_{L}+1}^{N-T_{R}} {\textbf{RG}}_k,
\end{CEquation}
whereas the alpha-trimmed variance of~$\textbf{RG}$,~$\hat{\sigma}^2_{\textbf{RG}}$, is defined as
\begin{CEquation}
\hat{\sigma}^2_{\textbf{RG}} = \frac{1}{N}\sum_{k=1}^{N} \left({\textbf{RG}}_k - \hat{\mu}_{\textbf{RG}} \right)^2.
\end{CEquation}
The alpha-trimmed mean and variance of~$\textbf{YB}$, $\hat{\mu}_{\textbf{YB}}$ and~$\hat{\sigma}^2_{\textbf{YB}}$, are similarly defined.
UICM is defined as
\begin{CEquation}
   \text{UICM}\! =\!\! -0.0268\sqrt{\hat{\sigma}^2_{\textbf{RG}}\!+\!\hat{\sigma}^2_{\textbf{YB}}}\!+\!0.1586 \sqrt{\hat{\mu}_{\textbf{RG}}\!+\!\hat{\mu}_{\textbf{YB} \vphantom{\beta^K}}}.
\end{CEquation}

UISM accounts for the  sharpness degradation caused by forward scattering in each colour channel. Let~$\mathcal{S}(\cdot)$ be the function  that applies the Sobel operator on a grayscale image~\cite{Sobel}. For example,~$\mathcal{S}(\textbf{R})$ is the edge image obtained by applying the Sobel operator on~$\textbf{R}$. UISM is defined in terms of the enhancement measure estimation function~\cite{Panetta2014EME}, $\mathcal{E}(\cdot)$, that measures the average logarithmic contrast ratio of blocks in the edge image, as
\begin{CEquation}
\mathcal{E}(\textbf{I}') = \frac{2}{L\cdot M}\sum_{l=1}^{L}\sum_{m=1}^{M}\ln\left(\frac{\max(\textbf{W}^{\textbf{I}'}_{l,m})}{\min(\textbf{W}^{\textbf{I}'}_{l,m})}\right),
\end{CEquation}
where~$\ln(\cdot)$ is the natural logarithm function,~$\max(\cdot)$ and~$\min(\cdot)$ are the maximum and minimum values in the block, respectively. UISM is defined as
\begin{CEquation}
\text{UISM} = 0.299
\cdot\mathcal{E}(\mathcal{S}(\textbf{R}))
+0.584
\cdot\mathcal{E}(\mathcal{S}(\textbf{G}))+0.114
\cdot\mathcal{E}(\mathcal{S}(\textbf{B})),
\end{CEquation}
where the weights for each of~$\mathcal{E}(\cdot)$ is chosen to reflect human visual system's response on the colour channel~\cite{DigitalImgProccessing_Pratt_1991}.

UIConM measures the relative contrast sensitivity using the logarithmic entropy~\cite{Panetta2014EME} of the Michelson Contrast with PLIP operators~\footnote{The Michelson Contrast of~$\textbf{I}'$ is defined as~$
  \frac{\max(\textbf{I}')- \min(\textbf{I}')}{\max(\textbf{I}')+ \min(\textbf{I}')}$.}~\cite{MichelsonContrast} in a block $\textbf{W}^{\textbf{I}'}_{l,m}$. The non-linear PLIP addition, subtraction and multiplication ($\oplus$,~$\ominus$ and $\otimes$, respectively) are used instead of their corresponding arithmetic operations \mbox{($+$, $-$ and $\times$)} as they are more consistent with human visual perceptions~\cite{Panetta2011ParameterizedLF}. 
  The attribute is defined as
  \begin{CEquation}
\begin{split}
\text{UIConM} 
\!=\! \frac{1}{L\! \cdot \!M}\!\otimes\! \sum_{l=1}^{L}\sum_{m=1}^{M}\!\mathcal{M}(\textbf{W}^{\textbf{I}'}_{l,m})
\, \ln\!\Big(\mathcal{M}(\textbf{W}^{\textbf{I}'}_{l,m})\Big),
\end{split}
\end{CEquation}
where $\mathcal{M}(\cdot)$ is the PLIP Michelson Contrast, defined as
  \begin{CEquation}
 \mathcal{M}(\textbf{W}^{\textbf{I}'}_{l,m}) = \frac{\max(\textbf{W}^{\textbf{I}'}_{l,m})\ominus \min(\textbf{W}^{\textbf{I}'}_{l,m})}{\max(\textbf{W}^{\textbf{I}'}_{l,m})\oplus \min(\textbf{W}^{\textbf{I}'}_{l,m})}.
\end{CEquation}

The IQA is defined as
\begin{CEquation}
\text{UIQM} = 0.0282\cdot \text{UICM} + 0.2953 \cdot \text{UIConM} + 3.5753\cdot \text{UISM}.
\end{CEquation}
UIQM states that the higher the value, the better the image quality.

\subsection*{iii. CCF (Wang et al.~\cite{WANG2018_NR_CFF})}

CCF  is a combination of three attributes: colourfulness, contrast and fog density. 

Colourfulness in CCF uses the~$L\alpha\beta$ colour model proposed by Ruderman et al.~\cite{Ruderman:98_lab}. 

\begin{CEquation}
\begin{alignedat}{8}
&\textbf{R}_{\ln} &&= &\ln(\textbf{R}) &- \mu_{\ln(\textbf{R})},\\
&\textbf{G}_{\ln} &&= &\ln(\textbf{G}) &- \mu_{\ln(\textbf{G})},\\
&\textbf{B}_{\ln} &&= &\ln(\textbf{B}) &- \mu_{\ln(\textbf{B})},
\end{alignedat}
\end{CEquation}
where~$\mu_{\ln(\textbf{R})}$ is the average intensity of~$\ln(\textbf{R})$. The opponent colour components are then defined as (compared to Eq.~\ref{eq:: UIQM_colourfulness_RG})
\begin{CEquation}
\begin{alignedat}{4}
&\boldsymbol \alpha &&= \textbf{R}_{\ln}-\textbf{G}_{\ln},\\
&\boldsymbol \beta && =  \frac{\textbf{R}_{\ln}+\textbf{G}_{\ln}}{2} - \textbf{B}_{\ln}.
\end{alignedat}
\end{CEquation}
Colourfulness is defined as
\begin{CEquation}
\text{Colourfulness} = \frac{\sqrt{\sigma^2_{\boldsymbol \alpha}+\sigma^2_{\boldsymbol \beta}}+0.3 \sqrt{\vphantom{\beta^K}\mu_{\boldsymbol \alpha}+\mu_{\boldsymbol \beta}}}{85.59},
\end{CEquation}
where~$\sigma^2_{\boldsymbol \alpha}$ and~$\mu_{\boldsymbol \alpha}$ ($\sigma^2_{\boldsymbol \beta}$ and~$\mu_{\boldsymbol \beta}$) are the variance and average of~$\boldsymbol \alpha$ ($\boldsymbol \beta$), respectively. 

The contrast attribute is defined in terms of the change of intensity of responses to an edge operator in an 'edge block' - 
defined as blocks with more than 0.2\% of the pixels being identified as an edge by the Sobel operator~\cite{Sobel}. The size of the block is chosen to be~\mbox{$64\times 64$}, which approximates the size of the foveal region representing the highest visual acuity of human visual system~\cite{Ferzli_Sharpness_JNBlur_TIP2009}. 
Let~$\textbf{E}^{\textbf{I}'}_t$ where~\mbox{$t \in \{1,...,T\}$}, be the one of the~$T$ edge blocks of~$\textbf{I}'$. Contrast of the image is defined as the sum of the $T$ intensity variances of the edge blocks, as 
\begin{CEquation}
\text{Contrast} = \sum_{t=1}^T\sqrt{\frac{1}{64\cdot 64}\sum_{e \in \textbf{E}^{\textbf{I}'}_t} (e - \mu_{\textbf{E}^{\textbf{I}'}_t})^2},
\end{CEquation}
where~$\mu_{\textbf{E}^{\textbf{I}'}_t}$ is the average intensity in the block~$
\textbf{E}^{\textbf{I}'}_t$.

The fog density was proposed to measure the visibility in outdoor hazy images by Choi et al.~\cite{FADE_FogIndex_2015}.
It is defined as the ratio between the image's Mahalanobis distance~\cite{Mahal} to a pre-collected set of 500 foggy images,~\mbox{$D_{\!f}$}, and that to a set of 500 fog-free images,~\mbox{$D_{\!f\!f}$}. The attribute is defined as
\begin{CEquation}
\text{Fog} = \frac{D_{\!f}}{D_{\!f\!f}+1}.
\end{CEquation}
The IQA is defined as
\begin{CEquation}
\begin{alignedat}{4}
\text{CCF} = & 0.17593\cdot \text{Colourfulness}\\ & +0.61759\cdot\text{Contrast} + 0.33988\cdot\text{Fog}.
\end{alignedat}
\end{CEquation}
CFF states that the higher the value, the better the image quality.

\newpage
\bibliographystyle{IEEEtran}
\bibliography{citations.bib}
\end{document}